\documentclass{article}
\usepackage{mathpazo}

\usepackage[utf8]{inputenc}
\usepackage[a4paper]{geometry}
\usepackage{xcolor}
\usepackage{array}
\usepackage{rotating}
\usepackage{float}
\usepackage{multirow}
\usepackage{amstext}
\usepackage{graphicx}
\usepackage[authoryear]{natbib}
\usepackage{nomencl}
\providecommand{\printnomenclature}{\printglossary}
\providecommand{\makenomenclature}{\makeglossary}
\makenomenclature
\usepackage[unicode=true,
 bookmarks=true,bookmarksnumbered=false,bookmarksopen=false,
 breaklinks=false,pdfborder={0 0 1},backref=false,colorlinks=false]
 {hyperref}
\hypersetup{pdftitle={Survey and synthesis of state of the art in driver monitoring},
 pdfauthor={Halin,  Anaïs; Verly, Jacques; Van Droogenbroeck, Marc},
 pdfkeywords={Keywords:  survey, driver monitoring, driver state, sensor, indicator, drowsiness, mental workload, distraction, emotions, under the influence}}

\makeatletter

\providecommand{\tabularnewline}{\\}

\usepackage{ifthen}
\newboolean{FORTHEWEB}
\setboolean{FORTHEWEB}{false}
\newboolean{FORARXIV}
\setboolean{FORARXIV}{true}

\setcitestyle{numbers}

\usepackage{wrapfig}
\usepackage{graphicx} 
\usepackage{lscape} 
\usepackage{framed,multirow}
\usepackage{diagbox} 

\usepackage{nomencl}
\usepackage{pifont}

\usepackage{xspace}

\newcommand{\ie}{i.e.\xspace}
\newcommand{\eg}{e.g.\xspace}

\newcommand{\he}{he/she\xspace}
\newcommand{\He}{He/She\xspace}
\newcommand{\his}{his/her\xspace}
\newcommand{\him}{him/her\xspace}

\newcommand{\Constructs}{States\xspace}
\newcommand{\Construct}{State\xspace}
\newcommand{\constructs}{states\xspace}
\newcommand{\construct}{state\xspace}

\makeatother

\begin{document}
\global\long\def\cardinality{\#}%


\global\long\def\gdl{\text{g/dL}}%

\global\long\def\microliter{\text{\ensuremath{\mu}L}}%

\global\long\def\micrometer{\text{\ensuremath{\mu}m}}%

\global\long\def\millimeter{\text{mm}}%

\global\long\def\dl{\text{dL}}%

\global\long\def\minute{\text{min}}%

\global\long\def\second{\text{sec}}%

\global\long\def\cm{\text{cm}}%

\global\long\def\m{\text{m}}%

\global\long\def\kmh{\text{km/h}}%

\global\long\def\numberOfSurveys{56}%

\global\long\def\numberOfArticles{254}%

\newcommand{\eqx}[1]{equation\ (#1)\xspace}               
\newcommand{\eqxy}[2]{equations\ (#1) and\ (#2)\xspace}   
\newcommand{\eqxyz}[3]{equations\ (#1),\ (#2), and\ (#3)\xspace}   
\newcommand{\Eqx}[1]{Equ.\ (#1)\xspace}                   
\newcommand{\Eqxy}[2]{Equ.\ (#1) and\ (#2)\xspace}        
\newcommand{\figx}[1]{Figure\ #1\xspace}                  
\newcommand{\figxy}[2]{Figures\ #1 and\ #2\xspace}        
\newcommand{\Figx}[1]{Fig.\ #1\xspace}                    
\newcommand{\Figxy}[2]{Fig.\ #1 and\ #2\xspace}           
\newcommand{\tabx}[1]{Table\ #1\xspace}                   
\newcommand{\secx}[1]{Section\ #1\xspace}                 
\newcommand{\secxy}[2]{Sections\ #1 and\ #2\xspace}       
\newcommand{\secxyzt}[4]{Sections\ #1,\ #2,\ #3, and\ #4\xspace}
\newcommand{\chapx}[1]{Chapter\ #1\xspace}                
\newcommand{\chapxy}[2]{Chapters\ #1 and #2\xspace}                
\title{Survey and synthesis of state of the art in driver monitoring}
\author{Anaïs Halin, Jacques G. Verly and Marc Van Droogenbroeck}
\date{August 2021}
\maketitle
\begin{abstract}
Road-vehicle accidents are mostly due to human errors, and many such
accidents could be avoided by continuously monitoring the driver.
Driver monitoring (DM) is a topic of growing interest in the automotive
industry, and it will remain relevant for all vehicles that are not
fully autonomous, and thus for decades for the average vehicle owner.
The present paper focuses on the first step of DM, which consists
in characterizing the state of the driver. Since DM will be increasingly
linked to driving automation (DA), this paper presents a clear view
of the role of DM at each of the six SAE levels of DA. This paper
surveys the state of the art of DM, and then synthesizes it, providing
a unique, structured, polychotomous view of the many characterization
techniques of DM. Informed by the survey, the paper characterizes
the driver state along the five main dimensions---called here “(sub)states”---of
drowsiness, mental workload, distraction, emotions, and under the
influence. The polychotomous view of DM is presented through a pair
of interlocked tables that relate these states to their indicators
(\eg, the eye-blink rate) and the sensors that can access each of
these indicators (\eg, a camera). The tables factor in not only the
effects linked directly to the driver, but also those linked to the
(driven) vehicle and the (driving) environment. They show, at a glance,
to concerned researchers, equipment providers, and vehicle manufacturers
(1) most of the options they have to implement various forms of advanced
DM systems, and (2) fruitful areas for further research and innovation.

\end{abstract}

\section{Introduction\label{sec:introduction}}

A report published in 2018~\citep{Singh2018Critical} gives the results
of an analysis performed on data about the events and related factors
that led to crashes of small road vehicles from 2005 to 2007 across
the USA. It indicates that the critical reasons for these crashes
are likely attributable to the driver (in $94\%$ of the cases), the
vehicle ($2\%$), the environment ($2\%$), and unknown causes ($2\%$).
An overwhelming proportion of these crashes is thus due to human error.
It is widely recognized that most of them could be avoided by constantly
monitoring the driver~\citep{Wouters2000Traffic,Aidman2015RealTime},
and by taking proper, timely actions when necessary.

Monitoring the driver is thus critically important, and this applies
to all vehicles, with the exception of those that are fully autonomous,
\ie, where the driver does not control the vehicle under any circumstances.
Given that the average driver will not own a fully-autonomous vehicle
for decades to come, ``\nomenclature{DM}{driver monitoring}driver
monitoring (DM)\footnote{The list of all abbreviations and their definitions appears in Section~\ref{list-of-acronyms},
in the Appendix.}'' will remain critically important during all this time.

This paper focuses on the topic of DM, which is usefully viewed as
consisting of two successive steps. In the first, one characterizes
the driver, or more precisely the state of the driver, and, in the
second, one decides what safety actions to take based on this characterization.
For example, in the monitoring of drowsiness, the first step might
compute the level of drowsiness, whereas the second might check whether
this level is at, or will soon reach, a critical level. More generally,
the decision process should ideally fuse the various characterization
parameters available and predict the future state of the driver based
on them. This paper focuses almost exclusively on the characterization
of the state of the driver, \ie, on the first step in DM, which is
also the one that is almost exclusively considered in the literature.

By “state of the driver” or “driver state”, we mean, in a loose way,
the state, or situation, that the driver is in from various perspectives,
in particular physical, physiological, psychological, and behavioral.
To deal with this driver state in a manageable, modular way, we consider
a specific number of distinct facets (such as drowsiness) of this
driver state, which we call “driver (sub)states”. In the sequel, “state”
thus refers either to the global state of the driver or to one of
its facets, or substates. This paper covers the main (sub)states of
drowsiness, mental workload, distraction, emotions, and under the
influence, which emerge as being the most significant ones in the
literature.

The core of the paper focuses on the characterization of each of these
(sub)states, using indicators (of this state) and sensors (to access
the values of these indicators in real time and in real driving conditions).
In the example of the (sub)state of drowsiness, an indicator thereof
is the eye-blink rate, and it can be accessed using a camera.

DM is important, whether the vehicle is equipped with some form of
“\nomenclature{DA}{driving automation}driving automation (DA)” (except
for full automation) or not. In future vehicles, DA and DM will need
to increasingly interact, and they will need to be designed and implemented
in a synergistic way. While the paper focuses on DM (and, more precisely,
on its characterization part), it considers and describes, at a high-level,
how DM and DA interact at the various, standard levels of DA.

As suggested by its title, the paper comprises two main phases: (1)
it reports on a systematic survey of the state of the art of DM (as
of early 2021), (2) it provides a synthesis of the many characterization
techniques of DM. This synthesis leads to an innovative, structured,
polychotomous view of the recent developments in the characterization
part of DM. In a nutshell, this view is provided by two interlocked
tables that involve the main driver (sub)states, the indicators of
these states, and the sensors allowing access to the values of these
indicators. The polychotomy presented should prove useful to researchers,
equipment providers, and vehicle manufacturers for organizing their
approach concerning the characterization and monitoring of the state
of the driver.

Section~\ref{sec:DA-and-DM} describes the standard levels of DA,
and the role played by DM for each. Section~\ref{sec:bibliographic-study}
indicates the strategy for, and the results of, our survey of the
literature on DM. Section~\ref{sec:framework-DMS} describes the
rationale and strategy for expressing the characterization of the
driver state as much as possible in terms of the triad of the (sub)states,
indicators, and sensors. Section~\ref{sec:Goal-and-approach} provides
our innovative, structured, polychotomous view of the characterization
part of DM. Sections~\ref{sec:drowsiness} to \ref{sec:under-the-influence}
successively describe the five driver (sub)states that the survey
revealed as being the most important. Section~\ref{sec:conclusion}
summarizes and concludes.

\section{Driving automation and driver monitoring\label{sec:DA-and-DM}}

In autonomous vehicles---also called self-driving or fully-automated
vehicles---DM plays a critical role as long as the automation allows
the driver to have some control over the vehicle. This section describes
the interaction between DM and DA in the context of the six levels
of DA defined by the \nomenclature{SAE}{Society of Automotive Engineers}Society
of Automotive Engineers (SAE) International~\citep{Sae2021Taxonomy},
ranging from 0 (no automation) to 5 (full automation).

Table~\ref{tab-automation-levels}, inspired by the \textit{SAE J3016
Levels of Driving Automation Graphic}, describes the role of each
of the three key actors in the driving task, namely the driver, the
\nomenclature{DS}{driver support}driver-support (DS) features, and
the \nomenclature{AD}{automated driving}automated-driving (AD) features,
at each of the six SAE levels. We also integrated into this table
a fourth actor, \ie, DM, as its role is crucial at all levels except
the highest, to ensure that the state of the driver allows \him\footnote{We use the inclusive pronoun ``\he'' and adjective ``\his''
to refer to the driver.} to perform the driving task safely, when applicable.

\begin{table}
\begin{centering}
\caption{This table shows the role played by each of the four key actors, \ie,
driver, driver-support~(DS) features, automated-driving~(AD) features,
and driver monitoring~(DM), at each of the six SAE levels of driving
automation (from 0 to 5).\label{tab-automation-levels}}
\par\end{centering}
\centering{}\resizebox{\textwidth}{!}{%
\begin{tabular}{|>{\raggedright}m{0.1428\textwidth}||>{\centering}m{0.1428\textwidth}|>{\centering}m{0.1428\textwidth}|>{\centering}m{0.1428\textwidth}||>{\centering}m{0.1428\textwidth}|>{\centering}m{0.1428\textwidth}|>{\centering}m{0.1428\textwidth}|}
\hline 
\textbf{SAE levels} & 0 & 1 & 2 & 3 & 4 & 5\tabularnewline
\textbf{Actors} $\downarrow$ & No

driving

automation & Driver

assistance & Partial

driving

automation & Conditional

driving

automation & High

driving

automation & Full

driving

automation\tabularnewline
\hline 
\hline 
Driver & \multicolumn{3}{>{\centering}m{0.4284\textwidth}||}{Driving

and

supervising DS features} & Driving

when AD~features request it & Driving

(if desired) when AD~features reach their limits & /\tabularnewline
\hline 
Driver-support (DS)

features & Warning and

temporary

support & Lateral \textit{or}

longitudinal

support & Lateral \textit{and}

longitudinal

support & \multicolumn{3}{c|}{/}\tabularnewline
\hline 
Automated-

driving (AD) features & \multicolumn{3}{c||}{/} & \multicolumn{2}{>{\centering}m{0.2856\textwidth}|}{Driving

when AD features

permit it} & Driving\tabularnewline
\hline 
Driver

monitoring

(DM) & Monitoring & \multicolumn{2}{>{\centering}m{0.2856\textwidth}||}{Monitoring

with relevant indicators} & Monitoring fallback-

ready driver & Monitoring

when driver

in control & /\tabularnewline
\hline 
\end{tabular}}
\end{table}

We now discuss some terminology. In Section~\ref{sec:introduction},
we introduced the term “driving automation (DA)” (as a convenient,
companion term for DM) and, in the previous paragraph, the SAE-suggested
term “automated driving (AD)”. While these two terms seem to further
add to a jumble of terms and abbreviations, they both appear in the
literature through their corresponding systems, \ie, the “\nomenclature{DAS}{driving-automation system}driving-automation
system (DAS)” and “\nomenclature{ADS}{automated-driving system}automated-driving
system (ADS)”. An ADS is a system consisting of the AD features, and
a DAS is a system that includes, among other things, both DS features
and AD features. One could also view the DS features as constituting
a system, but this is not needed here.

In future vehicles with progressively increasing degrees of automation,
the development of DASs and, in particular, ADSsshould go hand-in-hand
with the development of \nomenclature{DMS}{driver-monitoring system}driver-monitoring
systems (DMSs). The next four paragraphs complement the information
in Table~\ref{tab-automation-levels}.

At Levels 0 to 2, the driver is responsible for the driving task,
and \he may be aided by a variable number of DS features such as
automatic emergency braking, adaptive cruise control, and lane centering.
At Level 1, the DS features execute the subtask of controlling either
the lateral motion or the longitudinal motion of the vehicle (but
not both), expecting the driver to perform the rest of the driving
task. At Level 2, the DS features execute the subtasks of controlling
both the lateral motion and the longitudinal motion, expecting the
driver to complete the \nomenclature{OEDR}{object and event detection and response}object-and-event-detection-and-response
(OEDR) subtask and to supervise these features. At Levels 0 to 2,
a DMS should thus be used continuously. At Levels 1 and 2, for monitoring
the state of the driver, a vehicle-related indicator of driving performance
should be either avoided or used only when compatible with the DS
features that are engaged. The speed cannot, for instance, be used
as an indicator of the driver state when an adaptive cruise control
is regulating this speed. As more and more DS features are introduced
in vehicles, vehicle-related indicators of driving performance become
less and less relevant for monitoring the state of the driver, whereas,
driver-related parameters (both physiological and behavioral) remain
reliable indicators.

At any of Levels 3 to 5, and when the corresponding AD features are
engaged, the driver is no longer in charge of the driving task and
does not need to supervise them. Additionally, at Level 3, and at
any time, the driver must, however, be fallback-ready, namely, ready
to take over the control of \his vehicle when the AD features request
it (\ie, ask for it). A DMS should, therefore, be capable of (1)
assessing whether the current state of the driver allows \him to
take over the control of \his vehicle if requested now or in the
near future, and of (2) monitoring \his state as long as \he is
in control. \citet{ElKhatib2020Driver} discuss the potential need
for a DMS even when the vehicle is in control and does not require
the driver to supervise the driving or to monitor the driving environment.
Whenever the driver has the option of, \eg, engaging in some entertainment
activity, \he must be prepared to regain control in due course. Therefore,
at Level 3, despite that the driver is allowed to perform a secondary
task, a DMS is still necessary to ensure that the driver is ready
to take control at any time. Although the findings of various studies
are sometimes contradictory, \citet{Johns2014Effect} suggest that
it may be beneficial for the driver to maintain a certain level of
mental workload while \his vehicle is operated by a DAS, as this
could lead to better performance during a transfer of control from
automated to manual.

At Level 4, the AD features can only drive the vehicle under limited
conditions, but they will not require the driver to respond within
some specified time delay to a take-over request. The \nomenclature{ODD}{operational design domain}operational
design domain (ODD) specifies the conditions under which the DAS is
specifically designed to operate, including, but not limited to, (1)
environmental, geographical, and time-of-day restrictions, and/or
(2) the requisite presence or absence of certain traffic or roadway
characteristics. Still at Level 4, the AD features are capable of
automatically (1) performing a fallback of the driving task and (2)
reaching a minimal-risk condition (\eg, parking the car) if the driver
neither intervenes nor takes over the driving task within the delay.
If the driver decides to respond to the take-over request, one can
assume that the DMS would check that \his state allows for this,
even though the \textit{SAE J3016} does not say so explicitly.

At Level 5, the driving is fully automated under all possible conditions,
and no DMS is required as the driver is never in control, and becomes,
in effect, a passenger of the vehicle.

\section{Survey of literature on driver monitoring\label{sec:bibliographic-study}}

This section describes our survey of the literature on DM and DMSs.
The subsections below successively describe (1) our strategy for building
an initial set of references, (2) some conclusions drawn from these
references, (3) the design of a table for organizing them, (4) comments
about the content of this table, and (5,6) trends observable in it
or in some references. The analysis performed here guides the developments
in subsequent sections.

\subsection{Strategy for building an initial set of references, and number of
these}

\begin{figure}
\begin{centering}
\ifthenelse{\boolean{FORTHEWEB}}{\includegraphics[width=0.55\columnwidth]{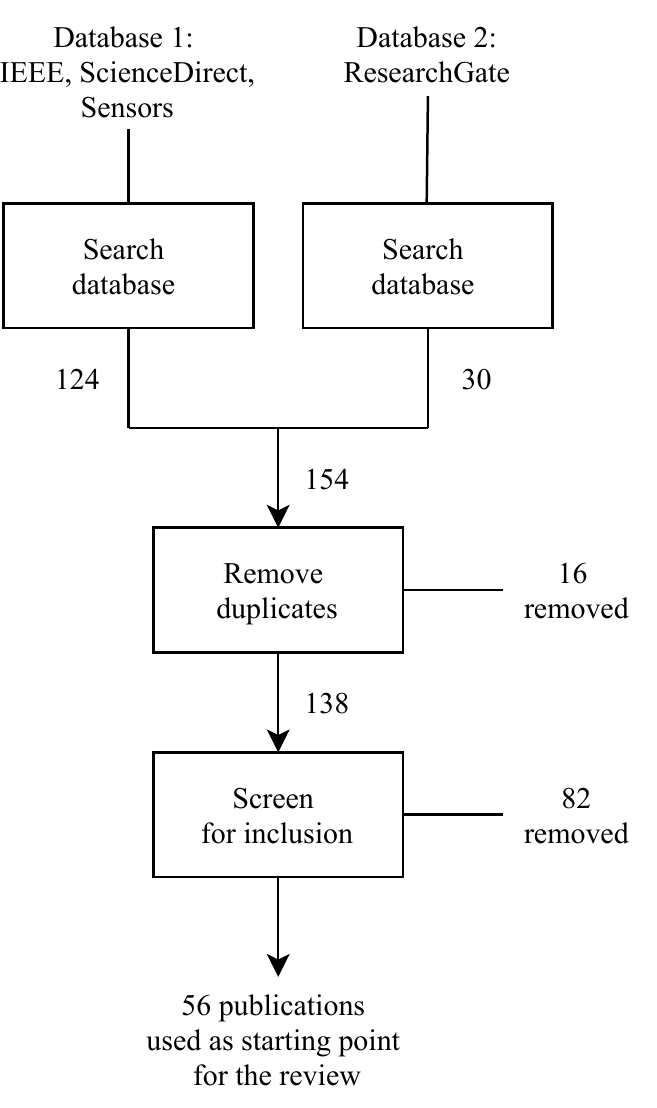}}{\includegraphics[width=0.55\columnwidth]{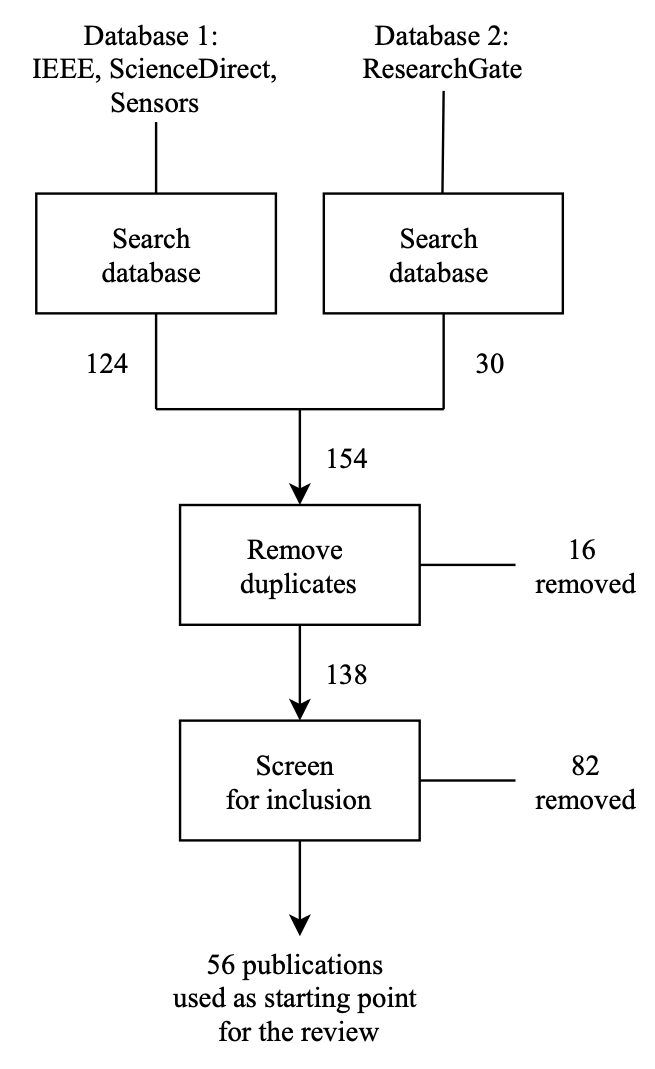}}
\par\end{centering}
\centering{}\caption{The flow diagram (1) illustrates the strategy used for our survey
of the literature on driver monitoring (DM) and driver-monitoring
systems (DMSs), and (2) shows the number of publications at each stage
of the process.\label{fig-flow-diagram}}
\end{figure}

To build an initial set of relevant references, we used an approach
inspired from~\citet{Gutierrez2021Comprehensive}. The block (or
flow) diagram of Figure~\ref{fig-flow-diagram} describes it.

Our search focused on surveys, reviews, and similar studies about
DM and DMSs. We independently performed two searches during February~2021.
The first focused on publications from IEEE, ScienceDirect, and Sensors,
and the second on publications from ResearchGate; these four databases
appeared well-suited for providing a useful set of initial references.
We used the search engine specific to each database and a boolean
query equivalent to \textit{(``survey'' OR ``review'') AND (``driver''
OR ``driving'') AND (``detection'' OR ``detecting'' OR ``behavior''
OR ``state'' OR ``monitoring'')}. We limited the search to publications
in English, and did not place any constraint on the dates of publication.
The two searches yielded $124$ and $30$ items, respectively. After
removing $16$ duplicates, we obtained a set of $138$ references.
We manually screened these, and only kept the ones satisfying the
two criteria of (1)~being in scientific journals or conference proceedings,
and (2)~providing a survey, review, or similar study of one or more
aspects of the domain of interest. This screening led to $\numberOfSurveys$
references, which appear in the first column of Table~\ref{tab-reviews-summary-sumvers}\footnote{A version of Table~\ref{tab-reviews-summary-sumvers} suitable for
printing appears in Appendix~\ref{appendix:table-for-printing}.}, and in the References section, which contains additional references
quoted later.

\subsection{Conclusions from preliminary analysis of $\protect\numberOfSurveys$
initial references}

The preliminary analysis of the $\numberOfSurveys$ initial references
led to the following high-level conclusions:
\begin{enumerate}
\item To characterize the (global) state of a driver, one should consider
the five main substates of drowsiness, mental workload, distraction,
emotions, and under the influence.
\item A wide variety of parameters, which we call “indicators”, are used
to characterize each of these substates, and some indicators are applicable
to more than one substate.
\item Ideally, a DMS should monitor not only the driver, but also the (driven)
vehicle and the (driving) environment.
\item A value for each indicator is obtained by processing data (mainly
signals and images) obtained from sensors “observing” the driver,
the vehicle, and the environment.
\item A DMS generally involves one or more types and/or instances of each
of the following: substate, indicator, and sensor.
\end{enumerate}
These conclusions guided the structuring and writing of the bulk of
the paper.

When the context is clear, we use “state” for the global state and
each of the five substates. The phrase ``state i'' and the plural
“states” imply that one is talking about one substate and several
substates, respectively.

\subsection{Design of structure of table organizing the initial references}

We used the above conclusions to design the structure of a table---namely
Table~\ref{tab-reviews-summary-sumvers}---for organizing the $\numberOfSurveys$
initial references in a useful way, in particular for the later synthesis
in this paper.


\global\long\def\void{}%

\global\long\def\typography#1{\text{#1}}%

\global\long\def\check{\typography V}%


\global\long\def\driverStateConstruct#1{{\color{blue}#1}}%

\global\long\def\manualDriverState{\typography{man}}%
 
\global\long\def\visualDriverState{\typography{vis}}%
 
\global\long\def\auditoryDriverState{\typography{aud}}%
 
\global\long\def\cognitiveDriverState{\typography{cog}}%

\global\long\def\stressEmotion{\typography{stress}}%
 
\global\long\def\angerEmotion{\typography{ang}}%

\global\long\def\alcoholUnderInfluence{\typography{alc}}%


\global\long\def\indicators#1{{\color{red}{\color{red}#1}}}%

\global\long\def\heartRatePhysiological{\typography{HR}}%
 
\global\long\def\electrodermalActivityPhysiological{\typography{EDA}}%
 
\global\long\def\breathingActivityPhysiological{\typography{breath}}%
 
\global\long\def\brainActivityPhysiological{\typography{brain}}%
 
\global\long\def\pupilDiameterPhysiological{\typography{pupil}}%

\global\long\def\gazeParametersBehavioral{\typography{gaze}}%
 
\global\long\def\blinksBehavioral{\typography{blink}}%
 
\global\long\def\percentageClosureBehavioral{\typography{PERCLOS}}%
 
\global\long\def\facialExpressionsBehavioral{\typography{facial}}%
 
\global\long\def\bodyPostureBehavioral{\typography{body}}%

\global\long\def\handsParametersBehavioral{\typography{hands}}%
 
\global\long\def\speechBehavioral{\typography{speech}}%

\global\long\def\wheelSteeringVehicle{\typography{wheel}}%
 
\global\long\def\laneDisciplineVehicle{\typography{lane}}%
 
\global\long\def\brakingBehaviorVehicle{\typography{brake}}%
 
\global\long\def\speedVehicle{\typography{speed}}%

\global\long\def\roadGeometryEnvironment{\typography{road}}%
 
\global\long\def\trafficDensityEnvironment{\typography{traf}}%
 
\global\long\def\weatherEnvironment{\typography{wea}}%


\global\long\def\indirectVehicleSensor{\typography{\check*}}%

\global\long\def\sensors#1{{\color{brown}#1}}%

\global\long\def\seatDriver{\typography{seat}}%
 
\global\long\def\steeringWheelDriver{\typography{ste\,w}}%
 
\global\long\def\safetyBeltDriver{\typography{saf\,b}}%
 
\global\long\def\cameraDriver{\typography{cam}}%
 
\global\long\def\cameraMobileDriver{\typography{cam*}}%
 
\global\long\def\microphoneDriver{\typography{mic}}%
 
\global\long\def\microphoneMobileDriver{\typography{mic*}}%
 
\global\long\def\wearableDriver{\typography{wea\,d}}%
\global\long\def\electrodesDriver{\typography{elec}}%
 
\global\long\def\eyeTrackerDriver{\typography{eye\,t}}%

\global\long\def\radarEnvironment{\typography{radar}}%
\global\long\def\cameraEnvironment{\typography{ext\,cam}}%


\global\long\def\testConditions#1{{\color{teal}#1}}%

\global\long\def\realConditions{\typography{real}}%
 
\global\long\def\simulatorConditions{\typography{sim}}%

\begin{table}
\caption{The first column of the table lists, by alphabetical order of first
author, the $\protect\numberOfSurveys$ references that resulted from
our survey on driver monitoring (DM) and related systems (DMSs). The
next three megacolumns and the last column briefly describe, for each
reference, the \constructs, indicators, sensors, and test conditions
considered therein. \label{tab-reviews-summary-sumvers}}

\resizebox{\textwidth}{!}{%
\begin{tabular}{|>{\raggedright}p{4cm}|c|>{\centering}m{1.5cm}|c|c|>{\centering}p{1.5cm}|c|c|c|c|c|c|c|c|>{\centering}p{1.5cm}|}
\hline 
\multirow{3}{4cm}{\textbf{References}} & \multicolumn{5}{c|}{\Constructs} & \multicolumn{5}{c|}{\textbf{Indicators}} & \multicolumn{3}{c|}{\textbf{Sensors}} & \multirow{3}{1.5cm}{\centering{}\textbf{\small{}Tests}}\tabularnewline
\cline{2-14} \cline{3-14} \cline{4-14} \cline{5-14} \cline{6-14} \cline{7-14} \cline{8-14} \cline{9-14} \cline{10-14} \cline{11-14} \cline{12-14} \cline{13-14} \cline{14-14} 
 & \multirow{2}{*}{\textbf{\small{}Drowsiness}} & \multirow{2}{1.5cm}{\centering{}\textbf{\small{}Mental workload}} & \multirow{2}{*}{\textbf{Distraction}} & \multirow{2}{*}{\textbf{Emotions}} & \multirow{2}{1.5cm}{\centering{}\textbf{\small{}Under the influence}} & \multicolumn{3}{c|}{\textbf{Driver}} & \multirow{2}{*}{\textbf{Vehicle}} & \multirow{2}{*}{\textbf{Environment}} & \multirow{2}{*}{\textbf{Driver}} & \multirow{2}{*}{\textbf{\small{}Vehicle}} & \multirow{2}{*}{\textbf{\small{}Environment}} & \tabularnewline
\cline{7-9} \cline{8-9} \cline{9-9} 
 &  &  &  &  &  & \textbf{Physiological} & \textbf{Behavioral} & \textbf{\small{}Subjective} &  &  &  &  &  & \tabularnewline
\hline 
\citet{Ahir2019Driver} & $\check$ &  &  &  &  & $\indicators{\heartRatePhysiological,\brainActivityPhysiological}$ & $\indicators{\gazeParametersBehavioral,\blinksBehavioral,\percentageClosureBehavioral,\facialExpressionsBehavioral,\bodyPostureBehavioral}$ &  & $\indicators{\wheelSteeringVehicle,\laneDisciplineVehicle,\speedVehicle}$ &  & $\sensors{\cameraDriver,\electrodesDriver}$ &  & $\sensors{\cameraEnvironment}$ & $\testConditions{\realConditions,\simulatorConditions}$\tabularnewline
\hline 
\citet{Alluhaibi2018Driver} & $\check$ &  & $\check$ & $\driverStateConstruct{\angerEmotion}$ &  &  & $\indicators{\speechBehavioral}$ &  & $\indicators{\wheelSteeringVehicle,\laneDisciplineVehicle,\brakingBehaviorVehicle,\speedVehicle}$ &  & $\sensors{\cameraMobileDriver,\microphoneMobileDriver}$ & $\indirectVehicleSensor$ &  & \tabularnewline
\hline 
\citet{Arun2012Driver} &  &  & $\driverStateConstruct{\visualDriverState,\cognitiveDriverState}$ &  &  & $\indicators{\heartRatePhysiological,\brainActivityPhysiological,\electrodermalActivityPhysiological,\pupilDiameterPhysiological}$ & $\indicators{\gazeParametersBehavioral,\blinksBehavioral,\bodyPostureBehavioral}$ & $\check$ & $\indicators{\wheelSteeringVehicle,\laneDisciplineVehicle,\brakingBehaviorVehicle,\speedVehicle}$ &  & $\sensors{\cameraDriver,\wearableDriver,\eyeTrackerDriver}$ & $\check$ &  & $\testConditions{\simulatorConditions}$\tabularnewline
\hline 
\citet{Balandong2018AReview} & $\check$ &  &  &  &  & $\indicators{\heartRatePhysiological,\brainActivityPhysiological}$ & $\indicators{\gazeParametersBehavioral,\blinksBehavioral,\percentageClosureBehavioral,\bodyPostureBehavioral}$ & $\check$ & $\indicators{\wheelSteeringVehicle,\laneDisciplineVehicle,\brakingBehaviorVehicle,\speedVehicle}$ &  & $\sensors{\electrodesDriver}$ &  &  & $\testConditions{\simulatorConditions}$\tabularnewline
\hline 
\multirow{1}{4cm}{\citet{Begum2013Intelligent}} & $\check$ &  & $\check$ & $\driverStateConstruct{\stressEmotion}$ &  & $\indicators{\heartRatePhysiological,\brainActivityPhysiological}$ &  &  &  &  & $\sensors{\seatDriver,\steeringWheelDriver,\safetyBeltDriver,\wearableDriver}$ &  &  & $\testConditions{\realConditions,\simulatorConditions}$\tabularnewline
\hline 
\citet{Chacon-murguia2015Detecting} & $\check$ &  &  &  &  & $\indicators{\heartRatePhysiological,\brainActivityPhysiological,\electrodermalActivityPhysiological}$ & $\indicators{\gazeParametersBehavioral,\blinksBehavioral,\bodyPostureBehavioral}$ &  & $\indicators{\wheelSteeringVehicle,\laneDisciplineVehicle,\brakingBehaviorVehicle,\speedVehicle}$ &  & $\sensors{\steeringWheelDriver,\cameraDriver}$ &  & $\sensors{\radarEnvironment}$ & $\testConditions{\realConditions}$\tabularnewline
\hline 
\multirow{1}{4cm}{\citet{Chan2019AComprehensive}} & $\check$ &  &  &  &  & $\indicators{\heartRatePhysiological,\brainActivityPhysiological}$ & $\indicators{\blinksBehavioral,\percentageClosureBehavioral,\facialExpressionsBehavioral,\bodyPostureBehavioral}$ &  & $\indicators{\wheelSteeringVehicle,\brakingBehaviorVehicle,\speedVehicle}$ &  & $\sensors{\cameraMobileDriver,\microphoneMobileDriver}$ &  &  & $\testConditions{\realConditions}$\tabularnewline
\hline 
\multirow{1}{4cm}{\citet{Chhabra2017ASurvey}} & $\check$ &  & $\check$ &  & $\driverStateConstruct{\alcoholUnderInfluence}$ & $\indicators{\breathingActivityPhysiological}$ & $\indicators{\gazeParametersBehavioral,\percentageClosureBehavioral,\facialExpressionsBehavioral,\bodyPostureBehavioral}$ &  & $\indicators{\wheelSteeringVehicle}$ & $\indicators{\roadGeometryEnvironment}$ & $\sensors{\seatDriver,\cameraMobileDriver,\microphoneMobileDriver}$ & $\indirectVehicleSensor$ &  & $\testConditions{\realConditions,\simulatorConditions}$\tabularnewline
\hline 
\citet{Chowdhury2018Sensor} & $\check$ &  &  &  &  & $\indicators{\heartRatePhysiological,\brainActivityPhysiological,\electrodermalActivityPhysiological}$ & $\indicators{\blinksBehavioral,\percentageClosureBehavioral}$ &  &  &  &  &  &  & $\testConditions{\simulatorConditions}$\tabularnewline
\hline 
\citet{Chung2019Methods} &  &  &  & $\driverStateConstruct{\stressEmotion}$ &  & $\indicators{\heartRatePhysiological,\breathingActivityPhysiological,\brainActivityPhysiological,\electrodermalActivityPhysiological,\pupilDiameterPhysiological}$ & $\indicators{\speechBehavioral}$ & $\check$ & $\indicators{\wheelSteeringVehicle,\laneDisciplineVehicle,\brakingBehaviorVehicle,\speedVehicle}$ &  & $\sensors{\cameraDriver,\wearableDriver}$ & $\check$ &  & $\testConditions{\realConditions,\simulatorConditions}$\tabularnewline
\hline 
\citet{Coetzer2009Driver} & $\check$ &  &  &  &  & $\indicators{\brainActivityPhysiological}$ & $\indicators{\gazeParametersBehavioral,\percentageClosureBehavioral,\facialExpressionsBehavioral,\bodyPostureBehavioral}$ &  & $\indicators{\wheelSteeringVehicle,\laneDisciplineVehicle,\speedVehicle}$ &  & $\sensors{\cameraDriver}$ & $\check$ &  & $\testConditions{\realConditions,\simulatorConditions}$\tabularnewline
\hline 
\citet{Dababneh2015Driver} & $\check$ &  &  &  &  & $\indicators{\brainActivityPhysiological,\electrodermalActivityPhysiological,\pupilDiameterPhysiological}$ & $\indicators{\blinksBehavioral,\percentageClosureBehavioral,\bodyPostureBehavioral}$ &  & $\indicators{\wheelSteeringVehicle,\laneDisciplineVehicle,\speedVehicle}$ & $\indicators{\roadGeometryEnvironment}$ & $\sensors{\cameraDriver,\wearableDriver}$ &  & $\sensors{\radarEnvironment}$ & $\testConditions{\realConditions,\simulatorConditions}$\tabularnewline
\hline 
\citet{Dahiphale2015AReview} & $\check$ &  & $\check$ &  &  &  & $\indicators{\gazeParametersBehavioral,\blinksBehavioral,\facialExpressionsBehavioral,\bodyPostureBehavioral}$ &  & $\indicators{\wheelSteeringVehicle}$ &  & $\sensors{\cameraDriver}$ &  &  & $\testConditions{\realConditions}$\tabularnewline
\hline 
\multirow{1}{4cm}{\citet{Dong2011Driver}} & $\check$ &  & $\check$ &  &  & $\indicators{\heartRatePhysiological,\brainActivityPhysiological,\pupilDiameterPhysiological}$ & $\indicators{\gazeParametersBehavioral,\blinksBehavioral,\percentageClosureBehavioral,\facialExpressionsBehavioral,\bodyPostureBehavioral}$ & $\check$ & $\indicators{\wheelSteeringVehicle,\laneDisciplineVehicle,\speedVehicle}$ & $\indicators{\roadGeometryEnvironment,\weatherEnvironment}$ & $\sensors{\cameraDriver}$ & $\check$ &  & $\testConditions{\realConditions}$\tabularnewline
\hline 
\multirow{1}{4cm}{\citet{ElKhatib2020Driver}} & $\check$ &  & $\driverStateConstruct{\manualDriverState,\visualDriverState,\cognitiveDriverState}$ &  &  & $\indicators{\heartRatePhysiological,\breathingActivityPhysiological,\brainActivityPhysiological,\electrodermalActivityPhysiological,\pupilDiameterPhysiological}$ & $\indicators{\gazeParametersBehavioral,\blinksBehavioral,\percentageClosureBehavioral,\facialExpressionsBehavioral,\bodyPostureBehavioral,\handsParametersBehavioral}$ &  & $\indicators{\wheelSteeringVehicle,\laneDisciplineVehicle,\speedVehicle}$ &  & $\sensors{\cameraDriver}$ & $\indirectVehicleSensor$ & $\sensors{\cameraEnvironment,\radarEnvironment}$ & $\testConditions{\realConditions,\simulatorConditions}$\tabularnewline
\hline 
\multirow{1}{4cm}{\citet{Ghandour2020Driver}} &  &  & $\driverStateConstruct{\manualDriverState,\visualDriverState,\auditoryDriverState,\cognitiveDriverState}$ & $\driverStateConstruct{\stressEmotion}$ &  & $\indicators{\heartRatePhysiological,\breathingActivityPhysiological,\brainActivityPhysiological,\electrodermalActivityPhysiological}$ & $\indicators{\gazeParametersBehavioral,\facialExpressionsBehavioral,\bodyPostureBehavioral,\speechBehavioral}$ & $\check$ & $\indicators{\wheelSteeringVehicle,\brakingBehaviorVehicle,\speedVehicle}$ &  & $\sensors{\cameraDriver,\wearableDriver}$ &  &  & $\testConditions{\realConditions,\simulatorConditions}$\tabularnewline
\hline 
\multirow{1}{4cm}{\citet{Hecht2018AReview}} & $\check$ & $\check$ & $\check$ &  &  & $\indicators{\heartRatePhysiological,\brainActivityPhysiological,\electrodermalActivityPhysiological,\pupilDiameterPhysiological}$ & $\indicators{\gazeParametersBehavioral,\blinksBehavioral,\percentageClosureBehavioral,\facialExpressionsBehavioral,\bodyPostureBehavioral}$ & $\check$ &  &  & $\sensors{\electrodesDriver,\eyeTrackerDriver}$ &  &  & $\testConditions{\realConditions,\simulatorConditions}$\tabularnewline
\hline 
\multirow{1}{4cm}{\citet{Kang2013Various}} & $\check$ &  & $\check$ &  &  & $\indicators{\heartRatePhysiological,\breathingActivityPhysiological,\brainActivityPhysiological,\electrodermalActivityPhysiological}$ & $\indicators{\gazeParametersBehavioral,\blinksBehavioral,\facialExpressionsBehavioral,\bodyPostureBehavioral}$ &  & $\indicators{\wheelSteeringVehicle,\laneDisciplineVehicle,\brakingBehaviorVehicle,\speedVehicle}$ &  & $\sensors{\seatDriver,\steeringWheelDriver,\cameraDriver}$ & $\check$ &  & $\testConditions{\realConditions,\simulatorConditions}$\tabularnewline
\hline 
\multirow{1}{4cm}{\citet{Kaplan2015Driver}} & $\check$ &  & $\check$ &  &  & $\indicators{\heartRatePhysiological,\brainActivityPhysiological}$ & $\indicators{\gazeParametersBehavioral,\blinksBehavioral,\percentageClosureBehavioral,\facialExpressionsBehavioral,\bodyPostureBehavioral,\speechBehavioral}$ &  & $\indicators{\wheelSteeringVehicle,\laneDisciplineVehicle,\brakingBehaviorVehicle,\speedVehicle}$ &  & $\sensors{\steeringWheelDriver,\cameraMobileDriver,\microphoneMobileDriver,\wearableDriver}$ & $\check$ &  & $\testConditions{\realConditions,\simulatorConditions}$\tabularnewline
\hline 
\citet{Kaye2018Comparison} & $\check$ &  &  & $\driverStateConstruct{\stressEmotion}$ &  & $\indicators{\heartRatePhysiological,\breathingActivityPhysiological,\brainActivityPhysiological,\electrodermalActivityPhysiological}$ &  & $\check$ &  &  &  &  &  & $\testConditions{\realConditions,\simulatorConditions}$\tabularnewline
\hline 
\citet{Khan2019AComprehensive} & $\check$ &  & $\driverStateConstruct{\manualDriverState,\visualDriverState,\auditoryDriverState,\cognitiveDriverState}$ &  &  & $\indicators{\heartRatePhysiological,\brainActivityPhysiological,\electrodermalActivityPhysiological}$ & $\indicators{\gazeParametersBehavioral,\percentageClosureBehavioral,\bodyPostureBehavioral}$ &  & $\indicators{\wheelSteeringVehicle,\laneDisciplineVehicle,\brakingBehaviorVehicle,\speedVehicle}$ &  & $\sensors{\wearableDriver}$ &  &  & $\testConditions{\realConditions}$\tabularnewline
\hline 
\citet{Kumari2017ASurvey} & $\check$ &  &  &  &  & $\indicators{\heartRatePhysiological,\brainActivityPhysiological}$ & $\indicators{\gazeParametersBehavioral,\blinksBehavioral,\percentageClosureBehavioral,\bodyPostureBehavioral}$ & $\check$ & $\indicators{\wheelSteeringVehicle,\laneDisciplineVehicle}$ &  & $\sensors{\cameraDriver}$ &  &  & \tabularnewline
\hline 
\citet{Lal2001ACritical} & $\check$ &  &  &  &  & $\indicators{\heartRatePhysiological,\brainActivityPhysiological,\electrodermalActivityPhysiological}$ & $\indicators{\percentageClosureBehavioral,\facialExpressionsBehavioral}$ &  &  &  & $\sensors{\cameraDriver}$ &  &  & $\testConditions{\simulatorConditions}$\tabularnewline
\hline 
\citet{Laouz2020Literature} & $\check$ &  &  &  &  & $\indicators{\heartRatePhysiological,\brainActivityPhysiological,\electrodermalActivityPhysiological}$ & $\indicators{\blinksBehavioral,\percentageClosureBehavioral,\facialExpressionsBehavioral,\bodyPostureBehavioral}$ & $\check$ & $\indicators{\wheelSteeringVehicle,\speedVehicle}$ &  & $\sensors{\seatDriver,\cameraDriver,\wearableDriver}$ &  & $\sensors{\cameraEnvironment}$ & $\testConditions{\realConditions}$\tabularnewline
\hline 
\citet{Leonhardt2018Unobtrusive} &  &  &  &  &  & $\indicators{\heartRatePhysiological,\breathingActivityPhysiological}$ &  &  &  &  & $\sensors{\seatDriver,\steeringWheelDriver,\safetyBeltDriver,\cameraDriver}$ &  &  & $\testConditions{\realConditions}$\tabularnewline
\hline 
\citet{Liu2019AReview} & $\check$ &  &  &  &  & $\indicators{\heartRatePhysiological,\brainActivityPhysiological,\pupilDiameterPhysiological}$ & $\indicators{\gazeParametersBehavioral,\blinksBehavioral,\percentageClosureBehavioral,\bodyPostureBehavioral}$ &  & $\indicators{\wheelSteeringVehicle,\laneDisciplineVehicle,\speedVehicle}$ &  & $\sensors{\cameraDriver}$ & $\check$ &  & $\testConditions{\realConditions}$\tabularnewline
\hline 
\citet{Marquat2015Review} &  & $\check$ &  &  &  & $\indicators{\pupilDiameterPhysiological}$ & $\indicators{\gazeParametersBehavioral,\blinksBehavioral,\percentageClosureBehavioral}$ & $\check$ &  &  & $\sensors{\eyeTrackerDriver}$ &  &  & $\testConditions{\realConditions,\simulatorConditions}$\tabularnewline
\hline 
\citet{Martinez2017Driving} &  &  &  & $\driverStateConstruct{\angerEmotion}$ &  &  &  &  & $\indicators{\brakingBehaviorVehicle,\speedVehicle}$ &  &  & $\indirectVehicleSensor$ &  & \tabularnewline
\hline 
\citet{Mashko2015Review} & $\check$ &  &  &  &  & $\indicators{\heartRatePhysiological,\brainActivityPhysiological,\electrodermalActivityPhysiological}$ & $\indicators{\gazeParametersBehavioral,\blinksBehavioral,\bodyPostureBehavioral}$ &  & $\indicators{\wheelSteeringVehicle,\laneDisciplineVehicle,\brakingBehaviorVehicle,\speedVehicle}$ &  & $\sensors{\cameraDriver,\wearableDriver}$ & $\check$ & $\sensors{\cameraEnvironment,\radarEnvironment}$ & $\testConditions{\realConditions,\simulatorConditions}$\tabularnewline
\hline 
\citet{Mashru2018Detection} & $\check$ &  &  &  &  & $\indicators{\heartRatePhysiological,\breathingActivityPhysiological}$ & $\indicators{\blinksBehavioral,\percentageClosureBehavioral,\facialExpressionsBehavioral,\bodyPostureBehavioral}$ & $\check$ & $\indicators{\wheelSteeringVehicle,\laneDisciplineVehicle}$ &  & $\sensors{\seatDriver,\steeringWheelDriver,\cameraDriver,\wearableDriver}$ &  &  & $\testConditions{\simulatorConditions}$\tabularnewline
\hline 
\citet{Melnicuk2016Towards} & $\check$ & $\check$ & $\driverStateConstruct{\cognitiveDriverState}$ & $\driverStateConstruct{\stressEmotion,\angerEmotion}$ &  & $\indicators{\heartRatePhysiological,\brainActivityPhysiological}$ & $\indicators{\blinksBehavioral,\percentageClosureBehavioral,\facialExpressionsBehavioral}$ &  & $\indicators{\wheelSteeringVehicle,\brakingBehaviorVehicle,\speedVehicle}$ & $\indicators{\roadGeometryEnvironment,\trafficDensityEnvironment,\weatherEnvironment}$ & $\sensors{\seatDriver,\steeringWheelDriver,\safetyBeltDriver,\cameraMobileDriver,\wearableDriver}$ & $\indirectVehicleSensor$ &  & $\testConditions{\realConditions}$\tabularnewline
\hline 
\citet{Mittal2016Head} & $\check$ &  &  &  &  & $\indicators{\heartRatePhysiological,\brainActivityPhysiological,\pupilDiameterPhysiological}$ & $\indicators{\blinksBehavioral,\percentageClosureBehavioral,\bodyPostureBehavioral}$ & $\check$ & $\indicators{\wheelSteeringVehicle,\laneDisciplineVehicle,\brakingBehaviorVehicle,\speedVehicle}$ &  & $\sensors{\cameraDriver,\electrodesDriver}$ & $\check$ & $\sensors{\cameraEnvironment}$ & $\testConditions{\realConditions}$\tabularnewline
\hline 
\citet{Murugan2019Analysis} & $\check$ &  &  &  &  & $\indicators{\heartRatePhysiological,\breathingActivityPhysiological,\brainActivityPhysiological,\electrodermalActivityPhysiological,\pupilDiameterPhysiological}$ & $\indicators{\blinksBehavioral,\percentageClosureBehavioral,\bodyPostureBehavioral}$ & $\check$ & $\indicators{\wheelSteeringVehicle,\laneDisciplineVehicle,\speedVehicle}$ &  & $\sensors{\cameraDriver,\electrodesDriver}$ & $\check$ &  & $\testConditions{\simulatorConditions}$\tabularnewline
\hline 
\citet{Nair2016ASurvey} & $\check$ &  & $\check$ &  & $\driverStateConstruct{\alcoholUnderInfluence}$ &  & $\indicators{\gazeParametersBehavioral,\percentageClosureBehavioral,\facialExpressionsBehavioral,\bodyPostureBehavioral}$ &  & $\indicators{\laneDisciplineVehicle}$ &  & $\sensors{\seatDriver,\cameraMobileDriver}$ & $\check$ & $\sensors{\radarEnvironment}$ & \tabularnewline
\hline 
\citet{Nemcova2021Multimodal} & $\check$ &  &  & $\driverStateConstruct{\stressEmotion}$ &  & $\indicators{\heartRatePhysiological,\breathingActivityPhysiological,\brainActivityPhysiological,\electrodermalActivityPhysiological}$ & $\indicators{\gazeParametersBehavioral,\blinksBehavioral,\percentageClosureBehavioral,\facialExpressionsBehavioral,\bodyPostureBehavioral}$ &  & $\indicators{\wheelSteeringVehicle,\laneDisciplineVehicle,\brakingBehaviorVehicle,\speedVehicle}$ &  & $\sensors{\seatDriver,\steeringWheelDriver,\cameraDriver,\wearableDriver,\eyeTrackerDriver}$ & $\check$ &  & $\testConditions{\realConditions,\simulatorConditions}$\tabularnewline
\hline 
\citet{Ngxande2017Driver} & $\check$ &  &  &  &  &  & $\indicators{\blinksBehavioral,\percentageClosureBehavioral,\facialExpressionsBehavioral,\bodyPostureBehavioral}$ &  &  &  & $\sensors{\cameraDriver}$ &  &  & \tabularnewline
\hline 
\citet{Oviedo-Trespalacios2016Understanding} &  & $\check$ & $\check$ &  &  &  & $\indicators{\gazeParametersBehavioral}$ &  & $\indicators{\wheelSteeringVehicle,\laneDisciplineVehicle,\brakingBehaviorVehicle,\speedVehicle}$ &  &  &  &  & $\testConditions{\realConditions,\simulatorConditions}$\tabularnewline
\hline 
\citet{Papantoniou2017Review} &  & $\check$ & $\check$ &  &  & $\indicators{\heartRatePhysiological,\breathingActivityPhysiological,\brainActivityPhysiological}$ & $\indicators{\gazeParametersBehavioral,\blinksBehavioral,\speechBehavioral}$ & $\check$ & $\indicators{\wheelSteeringVehicle,\laneDisciplineVehicle,\speedVehicle}$ &  & $\sensors{\cameraDriver}$ &  & $\sensors{\cameraEnvironment,\radarEnvironment}$ & $\testConditions{\realConditions,\simulatorConditions}$\tabularnewline
\hline 
\citet{Pratama2017AReview} & $\check$ &  &  &  &  & $\indicators{\heartRatePhysiological,\brainActivityPhysiological,\electrodermalActivityPhysiological}$ & $\indicators{\gazeParametersBehavioral,\blinksBehavioral,\percentageClosureBehavioral,\facialExpressionsBehavioral,\bodyPostureBehavioral,\handsParametersBehavioral}$ & $\check$ & $\indicators{\wheelSteeringVehicle,\laneDisciplineVehicle}$ &  & $\sensors{\cameraDriver,\wearableDriver,\electrodesDriver}$ &  & $\sensors{\cameraEnvironment}$ & $\testConditions{\realConditions,\simulatorConditions}$\tabularnewline
\hline 
\citet{Ramzan2019ASurvey} & $\check$ &  &  &  &  & $\indicators{\heartRatePhysiological,\breathingActivityPhysiological,\brainActivityPhysiological}$ & $\indicators{\blinksBehavioral,\percentageClosureBehavioral,\facialExpressionsBehavioral,\bodyPostureBehavioral}$ &  & $\indicators{\wheelSteeringVehicle,\laneDisciplineVehicle,\speedVehicle}$ &  & $\sensors{\cameraDriver,\wearableDriver,\electrodesDriver}$ & $\check$ &  & $\testConditions{\realConditions,\simulatorConditions}$\tabularnewline
\hline 
\citet{Sahayadhas2012Detecting} & $\check$ &  &  &  &  & $\indicators{\heartRatePhysiological,\brainActivityPhysiological,\pupilDiameterPhysiological}$ & $\indicators{\gazeParametersBehavioral,\blinksBehavioral,\percentageClosureBehavioral,\bodyPostureBehavioral}$ & $\check$ & $\indicators{\wheelSteeringVehicle,\laneDisciplineVehicle}$ &  & $\sensors{\seatDriver,\steeringWheelDriver,\cameraDriver,\wearableDriver}$ & $\check$ &  & $\testConditions{\realConditions,\simulatorConditions}$\tabularnewline
\hline 
\citet{Scott-Parker2017Emotions} &  &  &  & $\driverStateConstruct{\stressEmotion,\angerEmotion}$ &  & $\indicators{\heartRatePhysiological,\brainActivityPhysiological,\electrodermalActivityPhysiological}$ & $\indicators{\gazeParametersBehavioral,\facialExpressionsBehavioral}$ & $\check$ & $\indicators{\wheelSteeringVehicle,\laneDisciplineVehicle,\brakingBehaviorVehicle,\speedVehicle}$ & $\indicators{\trafficDensityEnvironment}$ & $\sensors{\eyeTrackerDriver}$ &  & $\sensors{\cameraEnvironment}$ & $\testConditions{\realConditions,\simulatorConditions}$\tabularnewline
\hline 
\citet{Seth2020ASurvey} & $\check$ &  &  &  &  &  &  &  &  &  & $\sensors{\cameraDriver}$ & $\check$ &  & $\testConditions{\realConditions}$\tabularnewline
\hline 
\citet{Shameen2018Electroencephalography} & $\check$ &  &  &  &  & $\indicators{\brainActivityPhysiological}$ & $\indicators{\gazeParametersBehavioral,\blinksBehavioral}$ &  &  &  & $\sensors{\electrodesDriver}$ &  &  & $\testConditions{\simulatorConditions}$\tabularnewline
\hline 
\citet{Sigari2014AReview} & $\check$ &  &  &  &  &  & $\indicators{\gazeParametersBehavioral,\blinksBehavioral,\percentageClosureBehavioral,\facialExpressionsBehavioral,\bodyPostureBehavioral}$ &  &  &  & $\sensors{\cameraDriver}$ &  &  & $\testConditions{\realConditions}$\tabularnewline
\hline 
\citet{Sikander2019Driver} & $\check$ &  &  &  &  & $\indicators{\heartRatePhysiological,\brainActivityPhysiological,\pupilDiameterPhysiological}$ & $\indicators{\gazeParametersBehavioral,\blinksBehavioral,\percentageClosureBehavioral,\bodyPostureBehavioral}$ & $\check$ & $\indicators{\wheelSteeringVehicle,\laneDisciplineVehicle}$ &  & $\sensors{\seatDriver,\steeringWheelDriver,\safetyBeltDriver,\cameraDriver,\wearableDriver,\electrodesDriver}$ &  &  & $\testConditions{\realConditions}$\tabularnewline
\hline 
\citet{Singh2021Analyzing} & $\check$ & $\check$ & $\check$ & $\check$ &  & $\indicators{\pupilDiameterPhysiological}$ & $\indicators{\gazeParametersBehavioral,\blinksBehavioral,\percentageClosureBehavioral,\facialExpressionsBehavioral}$ &  & $\indicators{\wheelSteeringVehicle,\brakingBehaviorVehicle,\speedVehicle}$ & $\indicators{\roadGeometryEnvironment,\trafficDensityEnvironment}$ & $\sensors{\cameraDriver,\wearableDriver}$ & $\check$ & $\sensors{\cameraEnvironment,\radarEnvironment}$ & $\testConditions{\realConditions}$\tabularnewline
\hline 
\citet{Subbaiah2019Driver} & $\check$ &  &  &  &  & $\indicators{\heartRatePhysiological,\brainActivityPhysiological,\pupilDiameterPhysiological}$ & $\indicators{\blinksBehavioral,\percentageClosureBehavioral,\facialExpressionsBehavioral,\bodyPostureBehavioral}$ &  &  &  & $\sensors{\cameraDriver}$ &  &  & $\testConditions{\realConditions,\simulatorConditions}$\tabularnewline
\hline 
\citet{Tu2016ASurvey} & $\check$ &  &  &  &  & $\indicators{\heartRatePhysiological,\brainActivityPhysiological}$ & $\indicators{\blinksBehavioral,\percentageClosureBehavioral,\facialExpressionsBehavioral,\bodyPostureBehavioral}$ &  & $\indicators{\wheelSteeringVehicle,\laneDisciplineVehicle,\speedVehicle}$ &  & $\sensors{\cameraMobileDriver,\wearableDriver,\electrodesDriver}$ & $\check$ &  & $\testConditions{\realConditions,\simulatorConditions}$\tabularnewline
\hline 
\citet{Ukwuoma2019Deep} & $\check$ &  &  &  &  & $\indicators{\heartRatePhysiological,\breathingActivityPhysiological,\brainActivityPhysiological}$ & $\indicators{\blinksBehavioral,\percentageClosureBehavioral,\facialExpressionsBehavioral,\bodyPostureBehavioral}$ &  & $\indicators{\wheelSteeringVehicle,\laneDisciplineVehicle,\brakingBehaviorVehicle}$ &  & $\sensors{\cameraDriver,\wearableDriver,\electrodesDriver}$ &  &  & $\testConditions{\realConditions}$\tabularnewline
\hline 
\citet{Vilaca2017Systematic} & $\check$ &  & $\check$ &  &  & $\indicators{\brainActivityPhysiological}$ & $\indicators{\gazeParametersBehavioral,\bodyPostureBehavioral}$ &  & $\indicators{\wheelSteeringVehicle,\laneDisciplineVehicle,\brakingBehaviorVehicle,\speedVehicle}$ &  & $\sensors{\cameraDriver,\microphoneDriver}$ & $\check$ & $\sensors{\cameraEnvironment}$ & \tabularnewline
\hline 
\citet{Vismaya2020AReview} &  &  & $\check$ &  &  &  & $\indicators{\gazeParametersBehavioral,\blinksBehavioral,\percentageClosureBehavioral,\bodyPostureBehavioral}$ &  &  &  & $\sensors{\cameraDriver,\eyeTrackerDriver}$ &  &  & $\testConditions{\realConditions,\simulatorConditions}$\tabularnewline
\hline 
\citet{Wang2006Driver} & $\check$ &  &  &  &  & $\indicators{\brainActivityPhysiological,\pupilDiameterPhysiological}$ & $\indicators{\gazeParametersBehavioral,\blinksBehavioral,\percentageClosureBehavioral,\bodyPostureBehavioral}$ &  & $\indicators{\laneDisciplineVehicle}$ &  & $\sensors{\cameraDriver,\wearableDriver}$ &  &  & $\testConditions{\realConditions,\simulatorConditions}$\tabularnewline
\hline 
\citet{Welch2019AReview} &  &  &  & $\driverStateConstruct{\stressEmotion,\angerEmotion}$ &  & $\indicators{\heartRatePhysiological,\breathingActivityPhysiological,\brainActivityPhysiological,\electrodermalActivityPhysiological}$ & $\indicators{\blinksBehavioral,\facialExpressionsBehavioral,\speechBehavioral}$ &  & $\indicators{\wheelSteeringVehicle,\brakingBehaviorVehicle,\speedVehicle}$ &  & $\sensors{\seatDriver,\steeringWheelDriver,\cameraDriver,\wearableDriver}$ & $\check$ &  & $\testConditions{\realConditions,\simulatorConditions}$\tabularnewline
\hline 
\citet{Yusoff2017Selection} &  &  & $\driverStateConstruct{\visualDriverState,\cognitiveDriverState}$ &  &  & $\indicators{\heartRatePhysiological,\brainActivityPhysiological,\electrodermalActivityPhysiological,\pupilDiameterPhysiological}$ & $\indicators{\gazeParametersBehavioral,\bodyPostureBehavioral}$ & $\check$ & $\indicators{\laneDisciplineVehicle,\speedVehicle}$ &  & $\sensors{\eyeTrackerDriver}$ &  &  & \tabularnewline
\hline 
\citet{Zhang2013Review} & $\check$ &  &  &  &  & $\indicators{\heartRatePhysiological,\brainActivityPhysiological}$ & $\indicators{\gazeParametersBehavioral,\blinksBehavioral,\percentageClosureBehavioral,\bodyPostureBehavioral}$ &  & $\indicators{\laneDisciplineVehicle,\speedVehicle}$ &  & $\sensors{\cameraDriver}$ &  & $\sensors{\cameraEnvironment}$ & $\testConditions{\realConditions,\simulatorConditions}$\tabularnewline
\hline 
\end{tabular}}
\end{table}

The $\numberOfSurveys$ references are listed in the first column,
labelled “References”, by alphabetical order of first author. The
three megacolumns following the first column successively correspond
to the three key items above, and are accordingly labelled “\Constructs'',
“Indicators”, and “Sensors”. The last column, labelled “Tests”, indicates
whether the technique or system described in a reference was tested
in the laboratory, or in real conditions (“in the wild”), or both.

The “\Constructs'' megacolumn is divided into $5$ columns corresponding
to the $5$ (sub)\constructs of interest. Each of the “Indicators”
and “Sensors” megacolumns is divided into $3$ columns corresponding
to the $3$ previously-listed items that a DMS should ideally monitor,
\ie, the driver, vehicle, and environment. The column corresponding
to the indicators for the driver is further divided into $3$ subcolumns
corresponding to the qualifiers ``physiological'', ``behavioral'',
and ``subjective''. Some other columns could be further subdivided,
such as for “Distraction”, but the table deals with such additional
subdivisions in a different way.

\subsection{Description of content of table of references}

We successively describe the three megacolumns of Table~\ref{tab-reviews-summary-sumvers}.

\subsubsection{\Constructs}

For each of the $\numberOfSurveys$ papers, we indicate which particular
(sub)state(s) it addresses. If a paper addresses drowsiness, we place
the checkmark “$\check$” in the corresponding column, and similarly
for mental workload. For the three other \constructs, we either use
a general “$\check$” or give more specific information, often via
an abbreviation. There are four types of distraction, \ie, manual,
visual, auditory, and cognitive, respectively abbreviated via $\manualDriverState$,
$\visualDriverState$, $\auditoryDriverState$, and $\cognitiveDriverState$.
These types are self-explanatory, but they are addressed later. For
emotions, we indicate the type, \ie, stress or anger ($\angerEmotion$).
For under the influence, we also indicate the type; in all cases,
it turns out to be alcohol ($\alcoholUnderInfluence$).

As an example, the second paper, by \citet{Alluhaibi2018Driver},
addresses drowsiness, distraction, and the emotion of anger.

All abbreviations used in Table~\ref{tab-reviews-summary-sumvers},
for this and other (mega)columns, are defined in Table~\ref{tab-list-abbreviations}.

\subsubsection{Indicators}

The indicator(s) used by a paper is (are) indicated in the same way
as above.

\subsubsection{Sensors}

The sensor(s) used by a paper is (are) indicated in a similar, but
not identical, way. If a sensor is embedded in a mobile device (typically
a smartphone), rather than in the vehicle, we add a “{*}”, leading
to \textit{$\cameraMobileDriver/\microphoneMobileDriver$} for a camera/microphone
of a mobile device. In the vehicle column, “$\check$” indicates that
the sensor is integrated in the vehicle, whereas “$\indirectVehicleSensor$”
indicates that it is part of a mobile device. For example, the vehicle
speed can be obtained via the \nomenclature{CAN}{controller area network }controller-area-network
(CAN) system/bus or a mobile device.

\begin{table}
\begin{centering}
\caption{The table defines the abbreviations used in Table~\ref{tab-reviews-summary-sumvers}.
They are organized according to the megacolumns and columns of Table~\ref{tab-reviews-summary-sumvers},
and are listed in alphabetical order.\label{tab-list-abbreviations}}
\par\end{centering}
\centering{}\ifthenelse{\boolean{FORTHEWEB}}{\includegraphics[width=0.8\textwidth]{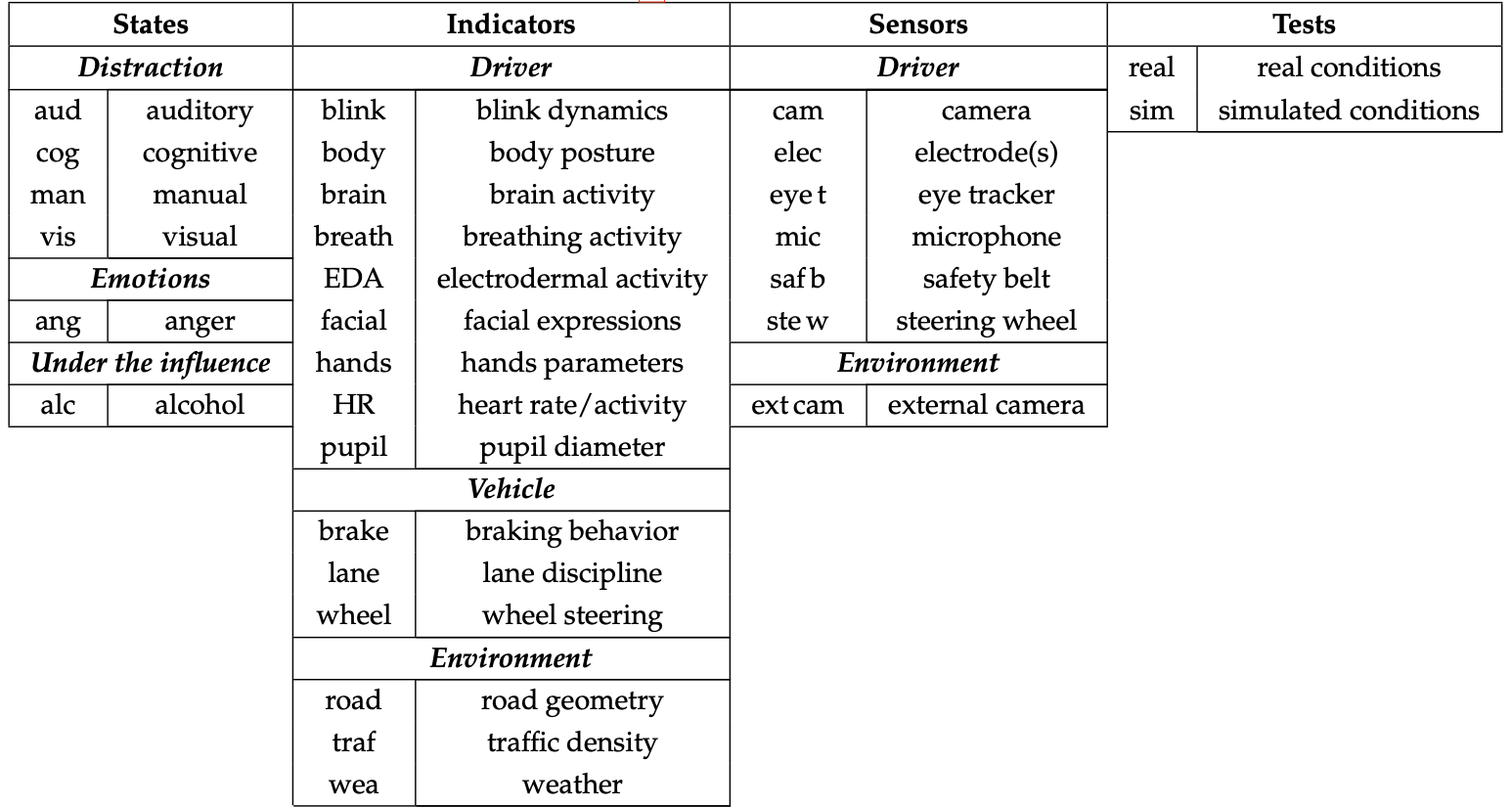}}{\resizebox{\textwidth}{!}{%
\begin{tabular}{|cc|c|c|cc|cc}
\hline 
\multicolumn{2}{|c|}{\textbf{\footnotesize{}\Constructs}} & \multicolumn{2}{c|}{\textbf{\footnotesize{}Indicators}} & \multicolumn{2}{c|}{\textbf{\footnotesize{}Sensors}} & \multicolumn{2}{c|}{\textbf{\footnotesize{}Tests}}\tabularnewline
\hline 
\multicolumn{2}{|c|}{\textbf{\textit{\footnotesize{}Distraction}}} & \multicolumn{2}{c|}{\textbf{\textit{\footnotesize{}Driver}}} & \multicolumn{2}{c|}{\textbf{\textit{\footnotesize{}Driver}}} & \multicolumn{1}{c|}{{\footnotesize{}$\realConditions$}} & \multicolumn{1}{c|}{{\footnotesize{}real conditions}}\tabularnewline
\cline{1-6} \cline{2-6} \cline{3-6} \cline{4-6} \cline{5-6} \cline{6-6} 
\multicolumn{1}{|c|}{{\footnotesize{}$\auditoryDriverState$}} & {\footnotesize{}auditory} & {\footnotesize{}$\blinksBehavioral$} & {\footnotesize{}blink dynamics} & \multicolumn{1}{c|}{{\footnotesize{}$\cameraDriver$}} & {\footnotesize{}camera} & \multicolumn{1}{c|}{{\footnotesize{}$\simulatorConditions$}} & \multicolumn{1}{c|}{{\footnotesize{}simulated conditions}}\tabularnewline
\cline{7-8} \cline{8-8} 
\multicolumn{1}{|c|}{{\footnotesize{}$\cognitiveDriverState$}} & {\footnotesize{}cognitive} & {\footnotesize{}$\bodyPostureBehavioral$} & {\footnotesize{}body posture} & \multicolumn{1}{c|}{{\footnotesize{}$\electrodesDriver$}} & {\footnotesize{}electrode(s)} &  & \tabularnewline
\multicolumn{1}{|c|}{{\footnotesize{}$\manualDriverState$}} & {\footnotesize{}manual} & {\footnotesize{}$\brainActivityPhysiological$} & {\footnotesize{}brain activity} & \multicolumn{1}{c|}{{\footnotesize{}$\eyeTrackerDriver$}} & {\footnotesize{}eye tracker} &  & \tabularnewline
\multicolumn{1}{|c|}{{\footnotesize{}$\visualDriverState$}} & {\footnotesize{}visual} & {\footnotesize{}$\breathingActivityPhysiological$} & {\footnotesize{}breathing activity} & \multicolumn{1}{c|}{{\footnotesize{}$\microphoneDriver$}} & {\footnotesize{}microphone} &  & \tabularnewline
\cline{1-2} \cline{2-2} 
\multicolumn{2}{|c|}{\textbf{\textit{\footnotesize{}Emotions}}} & {\footnotesize{}$\electrodermalActivityPhysiological$} & {\footnotesize{}electrodermal activity} & \multicolumn{1}{c|}{{\footnotesize{}$\safetyBeltDriver$}} & {\footnotesize{}safety belt} &  & \tabularnewline
\cline{1-2} \cline{2-2} 
\multicolumn{1}{|c|}{{\footnotesize{}$\angerEmotion$}} & {\footnotesize{}anger} & {\footnotesize{}$\facialExpressionsBehavioral$} & {\footnotesize{}facial expressions} & \multicolumn{1}{c|}{{\footnotesize{}$\steeringWheelDriver$}} & {\footnotesize{}steering wheel} &  & \tabularnewline
\cline{1-2} \cline{2-2} \cline{5-6} \cline{6-6} 
\multicolumn{2}{|c|}{\textbf{\textit{\footnotesize{}Under the influence}}} & {\footnotesize{}$\handsParametersBehavioral$} & {\footnotesize{}hands parameters} & \multicolumn{2}{c|}{\textbf{\textit{\footnotesize{}Environment}}} &  & \tabularnewline
\cline{1-2} \cline{2-2} \cline{5-6} \cline{6-6} 
\multicolumn{1}{|c|}{{\footnotesize{}$\alcoholUnderInfluence$}} & {\footnotesize{}alcohol} & {\footnotesize{}$\heartRatePhysiological$} & {\footnotesize{}heart rate/activity} & \multicolumn{1}{c|}{{\footnotesize{}$\cameraEnvironment$}} & {\footnotesize{}external camera} &  & \tabularnewline
\cline{1-2} \cline{2-2} \cline{5-6} \cline{6-6} 
\multicolumn{1}{c}{} &  & {\footnotesize{}$\pupilDiameterPhysiological$} & {\footnotesize{}pupil diameter} &  & \multicolumn{1}{c}{} &  & \tabularnewline
\cline{3-4} \cline{4-4} 
\multicolumn{1}{c}{} &  & \multicolumn{2}{c|}{\textbf{\textit{\footnotesize{}Vehicle}}} &  & \multicolumn{1}{c}{} &  & \tabularnewline
\cline{3-4} \cline{4-4} 
\multicolumn{1}{c}{} &  & {\footnotesize{}$\brakingBehaviorVehicle$} & {\footnotesize{}braking behavior} &  & \multicolumn{1}{c}{} &  & \tabularnewline
\multicolumn{1}{c}{} &  & {\footnotesize{}$\laneDisciplineVehicle$} & {\footnotesize{}lane discipline} &  & \multicolumn{1}{c}{} &  & \tabularnewline
\multicolumn{1}{c}{} &  & {\footnotesize{}$\wheelSteeringVehicle$} & {\footnotesize{}wheel steering} &  & \multicolumn{1}{c}{} &  & \tabularnewline
\cline{3-4} \cline{4-4} 
\multicolumn{1}{c}{} &  & \multicolumn{2}{c|}{\textbf{\textit{\footnotesize{}Environment}}} &  & \multicolumn{1}{c}{} &  & \tabularnewline
\cline{3-4} \cline{4-4} 
\multicolumn{1}{c}{} &  & {\footnotesize{}$\roadGeometryEnvironment$} & {\footnotesize{}road geometry} &  & \multicolumn{1}{c}{} &  & \tabularnewline
\multicolumn{1}{c}{} &  & {\footnotesize{}$\trafficDensityEnvironment$} & {\footnotesize{}traffic density} &  & \multicolumn{1}{c}{} &  & \tabularnewline
\multicolumn{1}{c}{} &  & {\footnotesize{}$\weatherEnvironment$} & {\footnotesize{}weather} &  & \multicolumn{1}{c}{} &  & \tabularnewline
\cline{3-4} \cline{4-4} 
\end{tabular}}}
\end{table}

\subsection{Trends observable in table}

Table~\ref{tab-reviews-summary-sumvers} reveals the following trends.

Drowsiness is the most covered \construct (with $44$ references
among the total of $\numberOfSurveys$), distraction is the second
most covered (with $20$ references), and more than one (sub)\construct
is considered in only $19$ references.

Indicators are widely used in most references, in various numbers
and combinations. Subjective indicators are not frequent (which is
to be expected given the constraints of real-time operation). While
several authors, such as~\citet{Dong2011Driver} and \citet{Sahayadhas2012Detecting},
emphasize the importance of the environment and of its various characteristics
(\eg, road type, weather conditions, and traffic density), few references
(and, specifically, only $6$) take it into account.

While the three “Sensors” columns seem well filled, several references
either neglect to talk about the sensor(s) they use, or cover them
in an incomplete way. Some references give a list of indicators, but
do not say which sensor(s) to use to get access to them. References
simply saying that, \eg, drowsiness can be measured via a camera
or an eye tracker do not help the reader. Indeed, these devices can
be head- or dashboard-mounted, and they can provide access to a variety
of indicators such as blink dynamics, PERCLOS, and gaze parameters.

Many systems are tested in real conditions, perhaps after initial
development and validation in a simulator. Many papers do not, however,
document systematically the test conditions for each method that they
describe.

\subsection{Other trends observable in references}

Other trends are not directly observable in Table~\ref{tab-reviews-summary-sumvers},
but can be identified in some individual references.

Experts agree that there does not exist any globally-accepted definition
for each of the first four \constructs that we decided to consider.
For example, even though many authors try to give a proper definition
for drowsiness, there remains a lot of confusion and inconsistencies
about the concepts of drowsiness and fatigue, and the difference between
them. There is thus a need to define, as precisely as possible, what
the first four \constructs are, and this is done in the sequel.

In the more recent references, one sees a trend, growing with time,
in the use of mobile devices, and in particular of smartphones~\citep{Alluhaibi2018Driver,Chan2019AComprehensive,Chhabra2017ASurvey,ElKhatib2020Driver,Kang2013Various,Kaplan2015Driver,Martinez2017Driving,Melnicuk2016Towards,Nair2016ASurvey,Tu2016ASurvey}.
A smartphone is relatively low-cost, and one can easily link it to
a DMS. This DMS can then use the data provided by the smartphone’s
many sensors, such as its inertial devices, microphones, cameras,
and navigation system(s). A smartphone can also receive data from
wearable sensors (\eg, from a smartwatch), which can provide information
such as heart rate (HR), skin temperature, and electrodermal activity
(EDA). A smartphone can also be used for its processing unit.

\section{Driver-state characterization via triad of states, indicators, and
sensors\label{sec:framework-DMS}}

Our survey of the field of DM and DMSs led us to the idea of synthesizing
this field in terms of the three key components of states, indicators,
and sensors. The next two subsections discuss the first two components,
and the third subsection brings all three components into a system
\nomenclature{BD}{block diagram}block diagram (BD).

\subsection{\Constructs}

Our survey convinced us that the (global) state of a driver should
be characterized along at least the five dimensions---called here
states---of drowsiness, mental workload, distraction, emotions, and
under the influence.

One goal of a DMS is to determine the levels of one or more of these
\constructs in real time, nearly continuously, and, preferably, in
a non-invasive way. We use “level” in a very general sense. The level
can take several forms, such as a numerical value or a label. The
numerical value can be on a continuous scale or on a discrete scale.
A label can be the most likely (output) class of a classifier together
with its probability, likelihood, or equivalent. A level can be binary,
e.g., 0 and 1, or “alert” and “drowsy”. The levels of one or more
of the five \constructs can then be used to issue alerts or take
safety actions; this is, however, not the object of this paper.

The first four \constructs present a formidable challenge in that
they are not defined in a precise way and cannot be measured directly,
by contrast with, say, physical quantities such as voltage and power.
The fifth \construct can be defined precisely, at least in the case
of alcohol, but the measurement of its level requires asking the driver
to blow in a breathalyzer and/or to submit to a blood test, both of
which can be performed neither in real time nor non-invasively. In
short, for all practical purposes, one cannot directly measure or
obtain the level of any of the five \constructs in any simple way.
This is the reason for having recourse to “indicators” of each of
these \constructs.

\subsection{Indicators}

While one may have an intuitive idea of what an indicator is, it is
useful to define, as precisely as possible, what it is. In a nutshell,
an indicator must be well defined, and there must be a clear procedure
for computing its values (at a succession of time instants) based
on input data provided by one or more sensors.

For the purpose of this paper, a “quantity” or “item” is called an
indicator for a given (sub)\construct if it satisfies all of the
following conditions:
\begin{itemize}
\item it has a precise definition based on science (\eg, physics, mechanics,
chemistry, biology, physiology);
\item it can be measured, or characterized in some way, with real-time constraint
when necessary, based upon data obtained from relevant sensors available
in the application of interest;
\item it must take values (such as numbers or labels) within a pre-specified
domain, and these values must preferably correspond to physical units
(such as seconds or Hertz);
\item it is not a unique and full descriptor of the \construct;
\item it is recognized, in the literature, as being linked, in some meaningful
way, to the state or trend thereof;
\item it is possibly useful with respect to one or more related, or unrelated,
\constructs;
\item it is reproducible, meaning that its value is always the same for
fixed data.
\end{itemize}
For example, the eye-blink rate (\ie, the blink rate of the left
or right pair of eyelids) is scientifically recognized as being indicative
of drowsiness. This parameter obeys all conditions above, and is thus
an indicator of drowsiness.

Similarly to the level of a \construct, we talk about the value of
an indicator. We use both ``value'' and ``level'' simply as a
way to implicitly communicate wether one is talking about an indicator
or a state. Ultimately, a set of values of the indicators of a \construct
must be converted into a level of this \construct. The conversion
may require the use of an advanced, validated algorithm.

Indicators are generally imperfect. In most cases, an indicator cannot
be guaranteed to be fully correlated with a related \construct. Due
to the presence of complex interrelationships between each (sub)\construct
and its indicators, it is important to use as many indicators as possible
to promote a valid and reliable interpretation of the (sub)state of
the driver and, ultimately, of the (global) state of the driver. An
example follows. The \nomenclature{HR}{heart rate}heart rate (HR)
is known to be an indicator of drowsiness. But, imagine that one relies
solely on the HR to monitor drowsiness, and that the driver must suddenly
brake to avoid an accident. Inevitably, this will cause \his HR to
undergo important variations. These particular variations have, however,
no direct link with \his level of drowsiness. Thus, while it is true
that the HR is an indicator of drowsiness, one cannot rely on it alone
to provide a reliable level of drowsiness. The environment, among
other things, needs to be considered.

The values of indicators are obtained through algorithms applied to
data collected via sensors.

\subsection{System view of characterization of a (sub)state}

\begin{figure}
\begin{centering}
\ifthenelse{\boolean{FORTHEWEB}}{\includegraphics[width=0.8\textwidth]{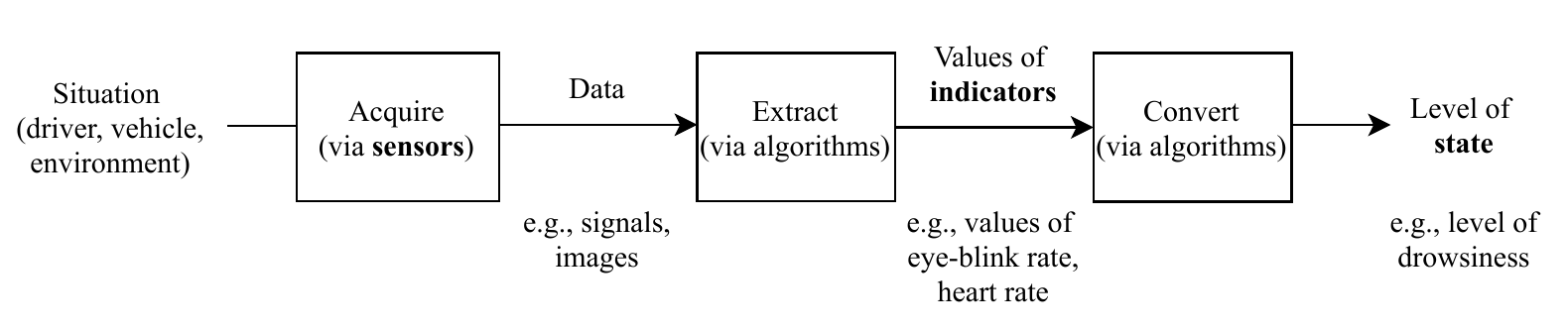}}{\includegraphics[width=1\textwidth]{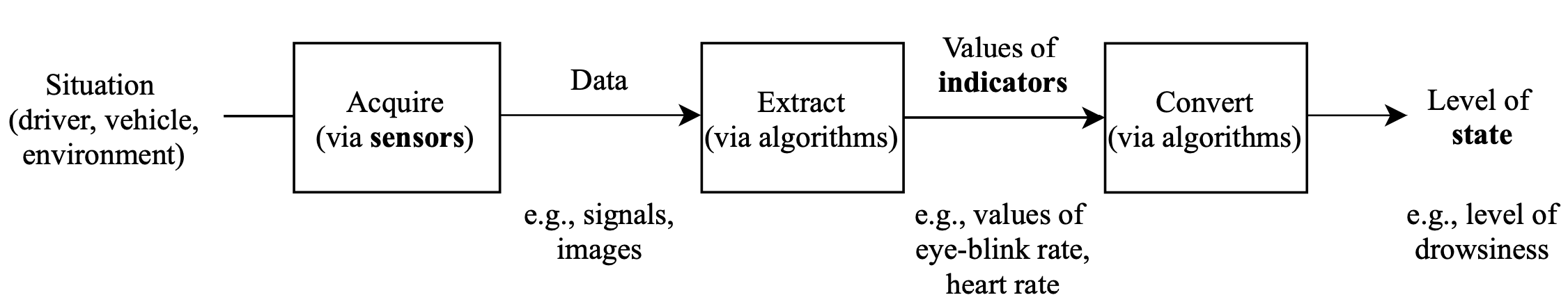}}
\par\end{centering}
\caption{The figure shows, for the context of driver monitoring (DM), the system
block diagram applicable to the characterization of a generic (sub)\construct.
The input is the situation of interest and the output is the level
of the \construct. The operation of each of the three subsystems
is described in the text. \label{fig-BD-DMS}}
\end{figure}

Figure~\ref{fig-BD-DMS} shows a system BD that uses the terminology
introduced above, \ie, sensors, indicators (and values thereof),
and \constructs (and levels thereof). The BD is drawn for a single,
generic \construct, and one must specialize it for each of the five
\constructs of interest (or others).

The BD is self-explanatory. The input is the situation of interest
(with the driver, vehicle, and environment). One or more sensors acquire
data, typically signals and images. Algorithms extract the values
of the indicators that are deemed relevant for the \construct of
interest. Other algorithms convert these values into a level of the
\construct. The three successive subsystems are labelled with the
operation they perform, \ie, acquire, extract, and convert. The input
and output of each subsystem should ideally be viewed as being functions
of time.

If several \constructs are used simultaneously, the value of a given
indicator can be used to compute the level of any \construct that
this indicator relates to.

\section{Synthesis of driver-state characterization via two interlocked tables\label{sec:Goal-and-approach}}

The previous section shows the key role played by the triad of \constructs,
indicators, and sensors (also emphasized in Figure~\ref{fig-BD-DMS})
in driver-state characterization, which is the first of two key steps
in DM, and the object of this paper. The present section describes
our approach to synthesize, in terms of this triad, the techniques
for driver-state characterization found in the literature.

\begin{figure}
\begin{centering}
\ifthenelse{\boolean{FORTHEWEB}}{\includegraphics[width=0.5\textwidth]{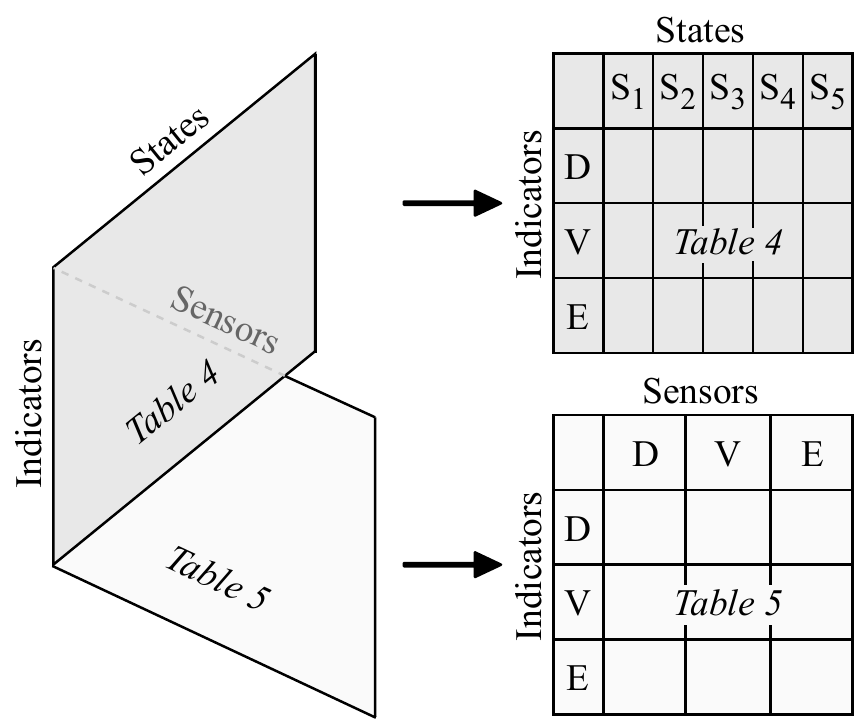}}{\includegraphics[width=0.5\columnwidth]{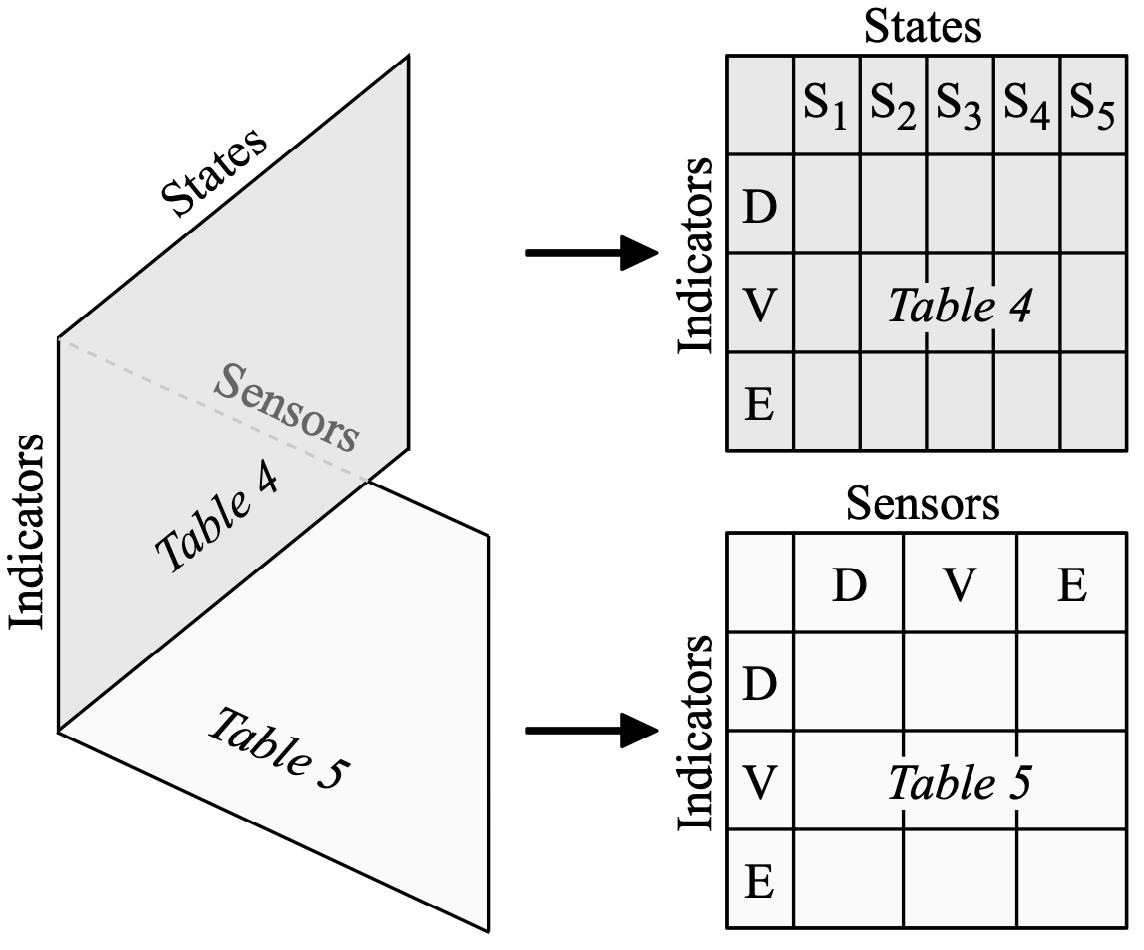}}
\par\end{centering}
\centering{}\caption{The figure shows simplified representations of key Table~\ref{tab-indicators}
(\constructs vs indicators) and Table~\ref{tab-sensors} (sensors
vs indicators). It also suggests that these tables can naturally be
interpreted as being two views of an underlying 3D array. $\text{S}_{\text{i}}$,
D, V, and E stand for “\Construct~i”, “Driver”, “Vehicle”, and “Environment'',
respectively.\label{fig:axes-analysis}}
\end{figure}

Our approach aims at answering, in a simple, visual way, the two following
questions: (1)~For a given \construct, what indicator(s) can one
use? (2)~For a given indicator, what sensor(s) can one use? We achieve
this goal by naturally providing two tables (or matrices) of “\constructs
vs indicators” and “sensors vs indicators”. These two tables can be
viewed as being two-dimensional (2D) views of a 3D table (or array)
of “\constructs vs indicators vs sensors”, as illustrated in Figure~\ref{fig:axes-analysis},
where the positions shown for the three dimensions and for the “dihedral”
they subtend make the tables on the right appear in numerical order
from top to bottom. The figure shows visually that the tables share
the “Indicators” dimension, and are thereby interlocked. It gives
a simplified representation of each of the tables that are progressively
filled in Sections~\ref{sec:drowsiness}-\ref{sec:under-the-influence},
\ie, Tables~\ref{tab-indicators} and \ref{tab-sensors}.

\subsection{Preview of two key tables}

In Figure~\ref{fig:axes-analysis}, the simplified representations
of Tables~\ref{tab-indicators} and \ref{tab-sensors} give the high-level
structures of these tables.

In Table~\ref{tab-reviews-summary-sumvers}, the megacolumn “Indicators”
is partitioned into the three columns “Driver”, “Vehicle”, and “Environment”.
Figure~\ref{fig:axes-analysis} shows, via the simplified representations,
that Tables~\ref{tab-indicators} and~\ref{tab-sensors} are also
partitioned in this way, but in megarows and with the corresponding
abbreviations D, V, and E. In Table~\ref{tab-reviews-summary-sumvers},
the megacolumn “Sensors” is partitioned in the same way as the megacolum
“Indicators”. This is reflected in Figure~\ref{fig:axes-analysis}
by the partitioning of Table~\ref{tab-sensors} into the megacolumns
D, V, and E. The figure shows that Table~\ref{tab-indicators} is
partitioned into the five megacolumns corresponding to the five \constructs,
denoted here by $\text{S}_{\text{1}},\,...,\,\text{S}_{\text{5}}$,
where $\text{S}_{\text{i}}$ stands for “\Construct~i”. This quoted
phrase appears at the beginning of the titles of the next five sections,
with the successive values of i.

Each lowest-level cell in both tables is destined to contain $0$,
$1$, or more related references.

The pair of tables allows one to answer other questions such as: (1)~If
one invests in the calculation of an indicator for a particular state,
what other state(s) can this indicator be useful for? (2)~If one
invests in a particular sensor for a particular state, what other
state(s) can this sensor be useful for?

\subsection{Further subdivision of rows and columns}

The rows and columns of Tables~\ref{tab-indicators} and~\ref{tab-sensors}
are further divided as follows. The D-megarows of Tables~\ref{tab-indicators}
and~\ref{tab-sensors} are subdivided as the D-megacolumns of Table~\ref{tab-reviews-summary-sumvers}
are, \ie, into the rows “Physiological”, “Behavioral”, and “Subjective”.

The D-megacolumns of Table~\ref{tab-sensors} are subdivided in a
way that does not already appear in Table~\ref{tab-reviews-summary-sumvers},
\ie, into the columns “Seat”, “Steering wheel”, “Safety belt”, “Internal
camera”, “Internal microphone”, and “Wearable”. Observe that the D-megarows
and D-megacolumns are not subdivided in the same way, even though
they correspond to the driver.

The V- and E- rows and columns are also further divided as necessary.

\subsection{Categories of indicators and sensors\label{subsec:Categories-of-indicators}}

We give examples of the various categories of indicators and sensors
that are further discussed in the next five sections. Below, we use
the self-explanatory terminology of ``X-based indicators'' and ``X-centric
sensors'', where X can be replaced by driver (or D), vehicle (or
V), or environment (or E).

\subsubsection{Indicators}

D-based indicators relate to the driver. They include physiological
indicators (\eg, heart activity, brain activity, electrodermal activity
(EDA)), behavioral indicators (\eg, eye blinks, gaze direction, hands
positions), and subjective indicators (which are not suited for real-world
operation, but can be used for validation at some point in the development
of a DMS).

V-based indicators relate to how the driver control \his vehicle,
for example, how \he controls the speed, steers, and brakes.

E-based indicators relate to the environment, viewed here as consisting
of three parts: (1) the outside environment (outside of vehicle),
(2) the inside environment (inside of vehicle), and (3) the contextual
environment (independent of the previous two). Examples of characteristics
of these parts of the environment are, respectively, (1) the road
type, weather conditions, and traffic density, (2) the temperature
and noise, and (3) the time of day and day of year. Each of these
characteristics (\eg, road type) can be used as an E-based indicator.

\subsubsection{Sensors}

Some D-centric sensors are placed in the seat (\eg, radar for breathing
activity), steering wheel (\eg, electrodes for electrocardiogram
(ECG)), and safety belt (\eg, \nomenclature{MI}{magnetic induction}magnetic
induction (MI) sensors). Some D-centric sensors, in particular cameras
(\eg, RGB) and microphones, are appropriately placed in the cockpit
to monitor the driver. We qualify these sensors of “internal”, to
distinguish them from similar sensors monitoring the external environment,
and qualified of “external”. Some D-centric sensors are wearables
(\eg, a smartwatch measuring HR and/or skin temperature). Since the
aim is to monitor the state of the driver, we assume throughout this
paper that the seat, safety belt, and similar items are related to
the driver.

V-centric sensors are mostly sensors---whether integrated in the
vehicle or not---that allow for the acquisition of vehicle parameters
such as speed, steering angle, and braking level. Such parameters
are often obtained via the CAN bus. Sensors (\eg, accelerometers,
gyroscopes) built into recent mobile devices can, however, also provide
some of this information.

E-centric sensors are sensors that allow for the acquisition of parameters
related to the environment. Cameras and radars can provide, for example,
information about the driving scene.

\subsection{Preview of next five sections}

The next five sections successively cover the five selected \constructs
in detail. In general, each section defines a \construct, the indicators
that characterize it, and the sensors that allow access to them, and
progressively fills Tables~\ref{tab-indicators} and~\ref{tab-sensors}
with relevant references.

At the end of the last of these five sections, both tables are complete.
They, together with the explanations in the five sections, constitute
the main contribution of this paper.

The structures of Tables~\ref{tab-reviews-summary-sumvers}, \ref{tab-indicators},
and \ref{tab-sensors} were obtained after a significant number of
iterations. This implies that the ultimate structure of Table~\ref{tab-reviews-summary-sumvers}
was informed by the content of Sections~\ref{sec:framework-DMS}~to~\ref{sec:under-the-influence}.

\section{\Construct 1: Drowsiness\label{sec:drowsiness}}

We provide a detailed description of (the \construct of) ``drowsiness'',
and we then present the indicators and sensors that can be used to
characterize it.

\subsection{Description}

\citet{Johns2000ASleep} appears to have given the earliest, accurate
definition of drowsiness, \ie, the state of being drowsy. \citet{Massoz2019NonInvasive}
provides useful, recent information about this state. Drowsiness is
an intermediate arousal state between wakefulness and sleep, \ie,
between being awake and being asleep; it thus refers to a state just
before potential sleep. A drowsy person has both a difficulty to stay
awake and a strong inclination to sleep. It is a continuous, fluctuating
state of (1) reduced awareness of the “here and now” \citep{Johns2001Assessing}
and (2) impaired cognitive and/or psychomotor performance. It is often
the result of a monotonous activity, such as a long drive on a monotonous
road. It can have a detrimental effect on the safety of driving. For
example, in the USA in 2018, there were $785$ fatal accidents due
to drowsiness for a total of $36,835$ people killed in motor vehicle
crashes and, in 2019, these numbers were $697$ vs $36,096$ \citep{NCSA2020verview}.
It can be viewed as a state of basic physiological need like hunger
and thirst, i.e., as an indication that one needs to sleep. It can
be considered to be synonymous with sleepiness, somnolence, and sleepening,
the latter being a less common term meaning “entry into sleep” \citep{Critchley1992OnSleepening}.

Drowsiness is, however, not synonymous with fatigue. These are two
distinct physiological states that are often confused, even in the
scientific literature. Fatigue corresponds to the feeling of being
tired or exhausted as a result of long periods of physical activity
and/or cognitive activity. It is characterized by an increasing difficulty
to accomplish an effort linked to a task. It can be considered to
be synonymous with tiredness. Talking about fatigue helps one to further
narrow down what drowsiness is and is not.

\citet{May2008Driver} suggest that, for driving, one should distinguish
between \nomenclature{SR}{sleep-related}sleep-related (SR) fatigue
and \nomenclature{TR}{task-related}task-related (TR) fatigue, based
on the causing factors. SR fatigue can be caused by sleep deprivation,
long wakefulness, and time of day (with effect of circadian rhythm),
while TR fatigue can be caused by certain characteristics of driving,
like task demand and duration, even in the absence of SR fatigue.
These suggested subcategories of fatigue clearly intersect with drowsiness,
but it is difficult to say exactly how.

Fatigue can be alleviated by taking a break (without necessarily sleeping),
while drowsiness can be alleviated by sleeping, even by taking a nap
or a power nap. One can be drowsy without being fatigued and vice-versa,
and one can be both. Fatigue and drowsiness both lead to decrements
in performance. In practice, it is difficult to distinguish between
them, and even more to quantify how much of a decrement is due to
each of them individually, especially in real time and non-invasively.
Their indicators appear to be mostly the same. In the driving context,
one focuses on monitoring drowsiness, with the main goal of preventing
the driver from falling asleep at the wheel.

There are many publications about the various ways of characterizing
drowsiness \citep{Ebrahimbabaie2020Prediction,Francois2018Development,Johns2005Monitoring,Massoz2019NonInvasive}
and apparently fewer for fatigue \citep{Aaronson2007Defining}. Very
few papers tackle both phenomena \citep{Shen2006Distinguishing}.

\subsection{Indicators}

We start with the driver-based indicators, divided into the three
categories of physiological, behavioral, and subjective indicators.

The most substantial changes in physiology associated with changes
in the \nomenclature{LoD}{level of drowsiness}level of drowsiness
(LoD) lie in the brain activity as measured by the \nomenclature{EEG}{electroencephalography}electroencephalogram
(EEG). \citet{Tantisatirapong2010Fractal} model EEG signals using
the \nomenclature{fBM}{fractal Brownian motion}fractal Brownian motion
(fBm) random process. They carried out experiments in a driving simulator,
and considered the three time periods of before, during, and after
sleep, where they mimic sleep by asking the driver to close \his
eyes, pretending to try to fall asleep. They saw corresponding changes
in the computed fractal dimension (related, for self-replicating random
processes, to the Hurst exponent), which allows them to classify the
driver as alert or drowsy. They conclude that the fractal dimension
of an EEG signal is a promising indicator of drowsiness. Changes in
physiology also manifest themselves in the heart activity, as measured
by the ECG. Indeed, as drowsiness increases, the HR decreases and
the \nomenclature{HRV}{heart rate variability}heart rate variability
(HRV) increases~\citep{Vicente2016Drowsiness}. However, HRV data
vary both between individuals and over time for each individual, depending
on both internal and external factors. Therefore, the many confounding
factors that also influence HRV must be accounted for in order to
use HRV as an indicator of drowsiness~\citep{Persson2021Heart}.
The breathing activity is an indicator of drowsiness, as changes in
breathing rate or inspiration-to-expiration ratio occur during the
transition from wakefulness to drowsiness~\citep{Kiashari2020Evaluation}.
Drowsiness leads to changes in EDA, also called skin conductance or
\nomenclature{GSR}{galvanic skin response}galvanic skin response
(GSR), which relates to the electrical resistance measured via electrodes
placed on the surface of the skin. The skin resistance fluctuates
with sweating, the level of which is controlled by the sympathetic
nervous system, which autonomously regulates emotional states such
as drowsiness~\citep{Michael2012Electrodermal}. The pupil diameter
instability has been linked to drowsiness. Indeed, several studies
found that the pupil diameter fluctuates at a low frequency and with
a high amplitude whenever a subject reports being drowsy~\citep{Lowenstein1963Pupillary,Nishiyama2007ThePupil,Wilhelm1998Pupillographic}.

Eye behavior is a good indicator of drowsiness. In a clinical setting,
one traditionally characterizes this behavior by \nomenclature{EOG}{electrooculography}electrooculography
(EOG)~\citep{Brown2006ISCEV}, which implies the use of electrodes.
In operational settings where a non-invasive characterization is highly
desirable, one generally uses video sequences of the eye(s) and applies
image-analysis methods to them. The dynamics of eye closures (in particular,
long and slow closures) is recognized as a strong and reliable indicator
of drowsiness~\citep{Schleicher2008Blinks}. The most-standard indicator
of spontaneous eye closure is the \nomenclature{PERCLOS}{percentage of closure}percentage
of closure (PERCLOS)~\citep{Dinges1998Evaluation,Dinges1998PERCLOS,Wierwille1994Research}.
It is usually defined as the proportion of time (over a given time
window) that the eyelids cover at least $70\%$ (or $80\%$) of the
pupils. As the LoD increases, the eye closures become slower and longer,
and the upper eyelid droops, and all of this contributes to an increase
in PERCLOS. Other reliable, standard indicators include mean blink
duration~\citep{Anund2008Driver,Schleicher2008Blinks}, mean blink
frequency or interval~\citep{Lisper1986Relation,Schleicher2008Blinks},
and eye closing and reopening speeds~\citep{Schleicher2008Blinks}.
Recently, \citet{Hultman2021Driver} used electrophysiological data
obtained by EOG and EEG to detect drowsiness with deep neural networks,
and found that, for driver-drowsiness classification, EOG data (and,
more precisely, the related blink data) are more informative than
EEG data.

All the above elements constitute objective indicators of drowsiness.
Besides these, there are subjective indicators, consisting of questionnaires
and self-reports. While they are not suitable for real-time characterization
of drowsiness, they can be used to validate other indicators, as ground
truth to train models, and/or to evaluate the performances of systems.
These subjective indicators include the \nomenclature{KSS}{Karolinska sleepiness scale}Karolinska
sleepiness scale (KSS)~\citep{Akerstedt1990Subjective}, the \nomenclature{SSS}{Stanford sleepiness scale}Stanford
sleepiness scale (SSS)~\citep{Hoddes1973SSS}, and the \nomenclature{VAS}{visual analog scale}visual
analog scale (VAS)~\citep{Monk1989VAS}.

The above information allows one to fill the cells of Table~\ref{tab-indicators}
at the intersection of the ``Drowsiness'' column and the ``Driver''
megarow. The latter lists a total of fourteen indicators. We stress
that these may or may not be relevant for each of the five \constructs.

A cell (at the lowest level) in the heart of Table~\ref{tab-indicators}
is either empty or filled with one or more related reference(s). For
example, this table shows that we found three significant references
about ``pupil diameter'' as an indicator of drowsiness, \ie, \citep{Lowenstein1963Pupillary,Nishiyama2007ThePupil,Wilhelm1998Pupillographic},
while we found no significant reference about ``gaze parameters''
as an indicator of drowsiness. The table shows, however, that we found
references reporting that this last indicator is useful for the \construct
of emotions (discussed later).

Below, as we progressively fill Tables~\ref{tab-indicators} and~\ref{tab-sensors},
we simply indicate which cell(s) is/are concerned. As we progress,
the discussion in the last two paragraphs remains valid, after proper
adaptation.

As should be clear from this discussion, the finer hierarchical partitioning
of Tables~\ref{tab-indicators} and \ref{tab-sensors} into the lowest-level
columns and rows is progressively obtained from the developments in
Sections~\ref{sec:bibliographic-study} to \ref{sec:under-the-influence}.

We now consider the vehicle-based indicators. In the literature, they
are often called measures of driving performance, the latter being
known to degrade with increasing drowsiness~\citep{Forsman2013Efficient,Kircher2002Vehicle,Wierwille1994Evaluation}.
These indicators characterize the driving behavior. Common such indicators
include speed, lateral control (or lane discipline), braking behavior,
and wheel steering. These last indicators are found in the central
part of Table~\ref{tab-indicators}, next to the ``Vehicle'' header.

The main vehicle-based indicator of drowsiness is the \nomenclature{SDLP}{standard deviation of lane position}standard
deviation of lane position (SDLP)~\citep{Godthelp1984TLC,Liang2019Prediction,Liu2009Predicting,Verwey2000Predicting}.
As the term suggests, SDLP measures the driver's ability to stay centered
in \his lane. Drowsiness can also produce greater variability in
driving speed~\citep{Arnedt2000Simulated}. Another important vehicle-based
indicator is the \nomenclature{SWM}{steering wheel movement}steering
wheel movement (SWM)~\citep{Liang2019Prediction}. It has been shown
that a drowsy driver makes fewer small SWMs and more large ones. When
a driver loses concentration, the vehicle begins to drift away from
the center of the lane, but, when the driver notices the drift, \he
compensates by large SWMs toward the lane center~\citep{Thiffault2003Monotony}.

\citet{JacobedeNaurois2019Detection} conducted a study in a driving
simulator, using different \nomenclature{ANN}{artificial neural network}artificial
neural networks (ANNs) based on various data, to detect drowsiness
and predict when a driver will reach a given LoD. The data used are
either (1) driver-based, physiological indicators (HR, breathing rate)
and behavioral indicators (blinks, PERCLOS, head pose), or (2) vehicle-based
indicators (lane deviation, steering wheel angle, acceleration, speed).
The results of the study show that the best performance is obtained
with behavioral data, successively followed by physiological data
and vehicle data, for both detection and prediction.

Most real-time, drowsiness-monitoring systems characterize the LoD
at the “present” time using sensor data located in a sliding time
window butting against this present time. Therefore, this LoD corresponds,
not to the present, but to roughly the center of the window, thus
several seconds, or tens of seconds, in the past. If this “present”
LoD is above a dangerous level, it may be too late for the driver
or the vehicle to take proper action. Given that, at $100\,\kmh$,
it takes about $2\,\second$ to get out of lane (then possibly hitting
an obstacle), predictions just $10$ to $20\,\second$ into the future
would already help. It is thus crucial to be able to predict (1) the
future evolution of the LoD and (2) the associated risks.

\citet{Ebrahimbabaie2020Prediction} and \citet{Ebrahimbabaie2018Excellent}
developed and tested a prediction system that (1) takes as input a
discrete-time, validated LoD signal consisting of the past LoD values
produced at regular intervals, up to just before the present time,
as in \citep{Francois2016Tests,Francois2018Development} (discussed
later), and (2) produces as output several types of predictions. Treating
the LoD signal as a realization of an underlying \nomenclature{RP}{random process}random
process (RP), the authors investigate the use of the RPs called \nomenclature{AR(I)MA}{autoregressive (integrated) moving average}“autoregressive
(integrated) moving average (AR(I)MA)” (from time-series analysis)
and \nomenclature{GBM}{geometric Brownian motion}“geometric Brownian
motion (GBM)” (found almost exclusively in finance). They show that
the LoD signal can generally be modeled as AR(I)MA and GBM within
each position of the sliding window (thus locally), they estimate
the parameters of the model for each position of the window, and they
use them to make predictions of one or more of the following three
types: future values of LoD signal, first hitting time (of a critical
LoD threshold), and survival probability.

We emphasize that “to predict” means “to tell beforehand”, and thus,
in the present context, to use past data to compute now a quantity
that describes some future situation. In the literature, this “future
situation” often turns out to be a “present situation”, so that no
prediction is performed.

The above information allows one to fill, in Table~\ref{tab-indicators},
the relevant cells of the ``Drowsiness'' column and the ``Vehicle''
megarow.

Note that there are no entries in the ``Environment'' megarow of
the ``Drowsiness'' column, which means that we did not find any
significant technique that uses one or more indicators related to
one of the three parts of the environment listed in Section~\ref{subsec:Categories-of-indicators}
(\ie, outside, inside, and contextual) to determine the level of
drowsiness of the driver. Some papers attempt to use the time of day
to try to capture the moments of the day where drowsiness tends to
peak. While the monotonicity of a road is known to increase driver
drowsiness, we have not found any paper using environment-based indicators
of road monotonicity (\eg, road geometry or traffic density), and
describing a way to give values to such indicators based upon available
data. As an aside, studies of drowsiness in a driving simulator often
use night driving and monotonous conditions to place the driver in
a situation conducive to drowsiness.

\subsection{Sensors}

Similarly to the indicators, we first address the driver-centric sensors.

In a vehicle, the HR can be monitored using electrodes that can be
placed at various locations, including the steering wheel (conductive
electrodes~\citep{Silva2012InVehicle}) and the seat (capacitive
electrodes~\citep{Leicht2015Capacitive}). ECG monitoring using steering-wheel-based
approaches is a feasible option for HR tracking, but requires both
hands to touch two different conductive parts of the steering wheel.

\nomenclature{BCG}{ballistocardiography}Ballistocardiography (BCG)
also allows for monitoring the cardiac activity unobtrusively. The
underlying sensing concept uses strain-gauge BCG sensors in the seat
or in the safety belt to detect both the cardiac activity and the
respiratory activity of the driver~\citep{Wusk2018NonInvasive}.
However, the vehicle vibrations make it difficult to use this sensor
in real driving conditions.

Information about the cardiac activity can be obtained using a camera
looking at the driver, in particular using \nomenclature{PPG}{photoplethysmography}photoplethysmography
(PPG) imaging~\citep{Zhang2017Webcam}.

Radar-based methods mainly provide information about movement, which
can of course be caused by both the cardiac activity and the respiratory
activity. Various sensor locations are possible, including integration
into the safety belt, the steering wheel, and the backrest of the
seat~\citep{Izumicontact,Schires2018Vital}.

Thermal imaging is a tool for analyzing respiration (or breathing)
non-intrusively. \citet{Kiashari2020Evaluation}~present a method
for the evaluation of driver drowsiness based on thermal imaging of
the face. Indeed, temperature changes in the region below the nose
and nostrils, caused by inspiration and expiration, can be detected
by this imaging modality. The procedure (1) uses a sequence of \nomenclature{IR}{infrared}infrared
(IR) images\footnote{Unless indicated otherwise, infrared (IR) means long-wave IR (LWIR),
\ie, with wavelengths of $8-14\,\micrometer$. LWIR is the ``thermal''
range of IR.} to produce a corresponding discrete-time signal of respiration, and
(2) extracts respiration information from it. The value of each successive
signal sample is the mean of the pixels in a rectangular window of
fixed size, representing the respiration region in the corresponding
IR image, adjusted frame-to-frame using a tracker. The initial respiration
region is determined based on the temporal variations of the first
few seconds of the sequence, and the region is tracked from frame-to-frame
by using the technique of ``spatio-temporal context learning''~\citep{Zhang2014Fast},
which is based on a Bayesian framework, and models the statistical
correlation between (1) the target (\ie, the tracked region) and
(2) its surrounding regions, based on the low-level characteristics
of the image (\ie, the intensity and position of each pixel). The
extracted information is the respiration rate and the inspiration-to-expiration
ratio. A classifier uses these rate and ratio to classify the driver
as awake or drowsy. A \nomenclature{SVM}{support vector machine}support
vector machine (SVM) classifier and a \nomenclature{KNN}{k-nearest neighbors}
$k$-nearest neighbors (KNN) classifier are used, and the first does
result in the best performance.

\citet{Francois2018Development} and \citet{Francois2016Tests} describe
a \nomenclature{POG}{photooculography}photooculographic (POG) system
that illuminates one eye with eye-safe IR light and uses as input
a sequence of images of this eye acquired by a monochrome camera that
is also sensitive in this IR range, and is head-mounted or dashboard-mounted.
A large number of ocular parameters, linked to the movements of the
eyelids (including blinks) and eyeball (including saccades), are extracted
from each video frame and combined into an LoD value, thus producing
an LoD signal. The output was validated using EEG, EOG, \nomenclature{EMG}{electromyography}EMG,
and reaction times. The head-mounted system is available commercially
as the Drowsimeter R100.

Using a camera, \citet{Massoz2018MultiTimescale}~characterize drowsiness
by using a multi-timescale system that is both accurate and responsive.
The system extracts, via \nomenclature{CNN}{convolutional neural network}convolutional
neural networks (CNNs), features related to eye-closure dynamics at
four timescales, \ie, using four time windows of four different lengths.
Accuracy is achieved at the longest timescales, whereas responsiveness
is achieved at the shortest ones. The system produces, from any 1-min
sequence of face images, four binary LoDs with diverse trades-offs
between accuracy and responsiveness. \citet{Massoz2018MultiTimescale}
also investigate the combination of these four LoDs into a single
LoD, which is more convenient for operational use.

\citet{Zin2018Vision} classify driver drowsiness by using a feature-extraction
method, the PERCLOS parameter, and an SVM classifier.

EDA is measured through electrodes placed on the skin of a person.
It can thus be measured through a wearable such as a smartwatch. Concerning
the other, relevant, physiological, driver-based indicators, (1) it
is challenging to get the pupil diameter in real conditions because
of issues with illumination conditions and camera resolution, among
others reasons, and (2) it is nearly impossible, as of this writing,
to characterize the brain activity in real time and in a non-intrusive,
reliable way.

\citet{Teyeb2015Vigilance} measure vigilance based on a video approach
calculating eye-closure duration and estimating head posture. \citet{Teyeb2016Towards}
monitor drowsiness by analyzing, via pressure sensors installed in
the driver seat, the changes in pressure distribution resulting from
the driver's body moving about in this seat. The authors suggest that
the techniques of these two papers can be usefully combined into a
multi-parameter system.

\citet{Bergasa2006RealTime} present a system to characterize drowsiness
in real time using images of the driver and extracting from them the
six visual parameters of PERCLOS, eye-closure duration, blink frequency,
nodding frequency, fixed gaze, and face pose. Using a camera, \citet{Baccour2020Camera}
and \citet{Dreissig2020Driver} monitor driver drowsiness based on
eye blinks and head movements.

Vehicle-based indicators can be collected in two main ways. Standard
indicators such as speed, acceleration, and steering wheel angle,
can be extracted from CAN-bus data~\citep{Li2008Design,Fridman2019MIT}.
The CAN bus enables intra-vehicle communications, linking the vehicle
sensors, warning lights, and \nomenclature{ECU}{electronic control unit}electronic
control units (ECUs). More advanced indicators can be obtained in
appropriately-equipped vehicles~\citep{Campbell2012TheSHRP,Fridman2019MIT}.
For example, speed and acceleration can be obtained via an \nomenclature{IMU}{inertial measurement unit}inertial
measurement unit (IMU), and following distance via a forward-looking
radar.

Since SDLP is considered to be a vehicle-based indicator of driver
drowsiness, one can quantify this indicator by examining the lane
discipline, \ie, the behavior of the vehicle in its lane. This is
traditionally done by using cameras (mounted inside, behind the windshield,
typically integrated beside the rear-view mirror)~\citep{Apostoloff2003Robust}
and/or laser sensors (mounted at the front of the vehicle) to track
the lane-delimiting lines when present. However, one can also use
the rumble strips (also called sleeper lines, audible lines, or alert
strips) when present. While these are designed to produce an audible,
acoustic signal intended to be sensed directly by the driver (as an
urgent warning or wake-up call), one could imagine using microphones
and/or vibration sensors to transform this acoustic/mechanical signal
into an electrical signal that is then analyzed via signal processing.

\citet{Bakker2021AMultiStage} describe a video-based system for detecting
drowsiness in real time. It uses computer vision and machine learning
(ML), and was developed and evaluated using naturalistic-driving data.
It has two stages. The first extracts, using data from the last 5
minutes, (1) driver-based indicators (\eg, blink duration, PERCLOS,
gaze direction, head pose, facial expressions) using an IR camera
looking at the driver's face, and (2) vehicle-based indicators (\eg,
lane positions, lane departures, lane changes) using an IR camera
looking at the scene ahead. This stage mostly uses pre-trained, \nomenclature{DNN}{deep neural network}deep-neural-network
(DNN) models. All indicators---also called deep features in DNNs---are
inputs to the second stage, which outputs an LoD, either binary (alert
or drowsy) or regression-like. This stage uses one KNN classifier,
trained and validated using KSS ratings as ground truth for the LoD,
and personalized for each driver by weighting more \his data during
training, thereby leading to higher performance during operation.

The above information allows one to fill the relevant cells of Table~\ref{tab-sensors}.

\section{\Construct 2: Mental workload\label{sec:mental-workload}}

We provide a detailed description of (the state of) “mental workload”,
and we then present the indicators and sensors that can be used to
characterize it.

\subsection{Description}

Mental workload, also known as cognitive\footnote{The present paper considers ``mental'' and ``cognitive'' as being
synonyms.} (work)load (or simply as driver workload in the driving context),
is one of the most important variables in psychology, ergonomics,
and human factors for understanding performance. This psychological
\construct is, however, challenging to monitor continuously~\citep{Marquat2015Workload}.

A commonly-used definition of mental workload is the one proposed
by~\citet{Hart1988Development}. They define mental workload as the
cost incurred by a person to achieve a particular level of performance
in the execution of a task. It is thus the portion of an individual's
mental capacity---necessarily limited---that is required by the
demands of this task~\citep{Borghini2014Measuring,Donnel1986Workload},
\ie, the ratio between the resources required to perform it and the
available resources of the person doing it~\citep{Sanders1998Human,Wickens2015Engineering}.

In the literature on mental workload, one often finds references to
another \construct called cognitive distraction. Mental workload
and cognitive distraction are two different concepts, even if they
can be linked when a driver performs secondary tasks while driving.
Cognitive distraction increases the mental workload of a driver. An
increase in mental workload is, however, not in itself an indication
of cognitive distraction. First, mental workload can increase in the
absence of distraction, for example, when a driver is focusing to
execute the primary task of driving correctly and safely. Second,
mental workload can increase significantly with an increasing complexity
of the driving environment~\citep{Schaap2013Relationship}. Cognitive
distraction is further considered later as a particular category of
(the \construct of) distraction.

Mental workload and stress are also linked since an increasing mental
workload usually induces some stress in the driver.

\subsection{Indicators}

In the driving context, visual tasks and mental tasks are closely
linked. Indeed, while driving, a driver is constantly perceiving \his
driving environment and analyzing what \he sees in order to make
the right decisions whenever required, for example, scanning a crossroad
and simultaneously judging the time and space relationships of other
road users to decide when it is safe to cross an intersection. Therefore,
it is logical that many researchers use eye-related parameters (\eg,
blinks, fixations, and pupil diameter) to assess the mental workload
of a driver~\citep{Marquat2015Review}.

Among the driver-based, physiological indicators, EDA~\citep{Kajiwara2014Evaluation},
HR~\citep{Gable2015Comparing}, and HRV~\citep{Paxion2014Mental}
are often used as indicators of mental workload. HR increases as a
task gets more difficult~\citep{Reimer2009Road} or if other tasks
are added~\citep{Fournier1999Electrophysiological}. EEG is also
a valuable indicator for studying mental workload because it records
the electrical activity of the brain itself, but it is complex to
analyze~\citep{Kim2013Highly}. The pupil diameter is considered
to be an indicator of mental workload~\citep{Gable2015Comparing,Kosch2018Look,Pfleging2016AModel}.
Indeed, \citet{Yokoyama2018Prediction} indicate that the mental workload
of a driver may be predicted from the slow fluctuations of the pupil
diameter in daylight driving. All physiological parameters mentioned
in this paragraph are, however, also influenced by other aspects of
the mental and physical situation of the driver (\eg, drowsiness
and TR fatigue) and by environmental situation (\eg, illumination
and temperature).

Among the driver-based, behavioral indicators, \citet{Fridman2018Cognitive}
have shown that the visual scanning by a driver decreases with an
increasing mental workload. Furthermore, since the interval of time
between saccades has been shown to decrease as the task complexity
increases, saccades may be a valuable indicator of mental workload~\citep{Liao2016Detection,May1990Eye}.

Subjective measures of mental workload exist, like the \nomenclature{NASA TLX}{NASA task load index}NASA
task load index (NASA TLX)~\citep{Hart1988Development}, which is
a workload questionnaire for self-report, and the \nomenclature{RSME}{rating scale mental effort}rating
scale mental effort (RSME).

Driving performance can diminish as a result of an increase in mental
workload. The vehicle-based indicators which are the most sensitive
to such an increase are SDLP and SWM~\citep{Schaap2013Relationship}.

\citet{Palasek2019Attentional} use the driving environment to estimate
the attentional demand required from the driver to drive. The features
extracted from the analysis of the driving environment are thus indicators
of the mental workload of the driver.

The above information allows one to fill, in Table~\ref{tab-indicators},
the relevant cells of the ``Mental workload'' column.

\subsection{Sensors}

Cameras are often used in the literature to characterize mental workload
as they are particularly well suited to extract driver-based, behavioral
indicators and are non-invasive.

\citet{Fridman2018Cognitive} describe a system for characterizing,
non-invasively, via a camera facing the driver, what they call \his
\nomenclature{CL}{cognitive load}cognitive load (CL). The system
exploits the well-documented, experimental observation that the angular
distribution of gaze direction (often characterized by the 2D pupil
position) tends to become more concentrated, especially vertically,
when the CL increases. Using video imagery, the system classifies
the CL of the driver into one of the three CL levels (low, medium,
high), as \he engages in activities other than the primary task of
driving, such as a conversation or the adjustment of the infotainment
system. The system extracts, from a 90-frame, 6-second video clip,
via computer vision, the face and the region of one eye of the driver.
It then uses one of two methods: (1) mainly \nomenclature{AAM}{active appearance model}active
appearance models (AAMs) for the face, eyelids, and pupil (when visible)
to produce a sequence of pupil 2D positions, and (2) one \nomenclature{HMM}{hidden Markov model}hidden
Markov model (HMM) for each of the three CL levels. The second method
uses a single 3D CNN with three output classes corresponding to these
levels. The two methods thus rely on a sequence of pupil positions
and on a sequence of eye images, respectively. The output of the system
is one of the three CL levels.

In order to develop this system, the authors first acquired training
data in real-driving conditions while imposing on the driver a secondary
task of a given CL level. This imposition of a given CL level while
performing a primary task (here driving) is commonly achieved in the
literature through the standard “$n$-back” task, where the three
values of $n$, \ie, $n=0$, $1$, and $2$, are viewed as corresponding
to low, medium, and high CL. For the $n$-back task, a sequence of
numbers is dictated to the subject, who is asked, for each number,
whether it matches the one dictated $n$ positions earlier in the
sequence. For example, for $n=2$, the subject must indicate whether
the current number is the same as the one \he heard 2 steps before,
all this while \he performs the primary task, here driving.

The authors indicate (1) that the differences in cognitive loading
for the three levels have been validated using, among others, physiological
measurements (e.g., HR, EDA, and pupil diameter), self-report ratings,
and detection-response tasks, and (2) that these levels have been
found to cover the usual range of secondary tasks while driving, such
as manipulating a radio or a navigation system.

It is noteworthy that the data used for building the system was acquired
through real driving, during which the driver repeatedly performed
$n$-back tasks, while a camera was recording \his face and surrounding
area, this by contrast with the many other developments made using
a driving simulator, in highly controlled conditions, and difficult
to implement in real-life conditions.

The authors indicate that, while they use the term “cognitive load”,
the literature often uses synonyms like “cognitive workload”, “driver
workload”, and “workload”.

\citet{Musabini2020Heatmap} describe another system that is also
based on eye-gaze dispersion. They use a camera facing the driver,
produce a heatmap representing the gaze activity, and train an SVM
classifier to estimate the mental workload based on the features extracted
from this representation.

\citet{Le2020Evaluating} characterize the mental workload based on
the involuntary eye movements of the driver, resulting from head vibrations
due to changing road conditions. They report that, as the mental workload
increases, these involuntarily eye movements become abnormal, resulting
in a mismatch between the actual eye movements measured via an eye-tracking
device and the predicted eye movements resulting from a VOR+OKR model\footnote{VOR and OKR are the abbreviations of vestibular-ocular reflex and
optokinetic response.}. For each driver, the VOR parameters are estimated during the first
$10\,\second$ of driving in condition of normal mental workload,
whereas the parameter in the OKR model is fixed. The hypothesis of
abnormal eye movements while driving under mental workload was validated
using a t-test analysis. Different levels of mental workload were
induced in a driving simulator using the $n$-back task.

\citet{Palasek2019Attentional} use an external camera recording the
driving environment to estimate the attentional demand using attentive-driving
models. Indeed, the task of driving can sometimes require the processing
of large amounts of visual information from the driving environment,
resulting in an overload of the perceptual systems of a human being.
Furthermore, traffic density is known to increase the mental workload~\citep{Hao2007Effect},
so that urban environments lead to a higher mental workload than rural
and highway environments do~\citep{Young2009Measuring}, all other
conditions being equal.

The above information allows one to fill the relevant cells of Table~\ref{tab-sensors}.

\section{\Construct 3: Distraction\label{sec:distraction}}

By contrast with the two previous sections, we start with some background
information (up to Section~\ref{subsec:manual-distraction}) on the
\construct of distraction.

The globally accepted definition of driver distraction follows: it
is a diversion of attention, away from activities critical for safe
driving (the primary task) and toward a competing activity~\citep{Ranney2000NHTSA,Regan2008Driver}.

Inattention, sometimes used---mistakenly---as a synonym of distraction,
is defined as a diminished attention to activities that are critical
for accomplishing a primary task, but not necessarily in the presence
of a competing activity~\citep{Regan2008Driver}. Therefore, driver
distraction is one particular form of driver inattention~\citep{Regan2011Driver}.
Inattention is a broader term as it can be caused, for example, by
drowsiness. It indeed occurs in a wide range of situations in which
the driver fails to attend to the demands of driving, such as when
a desire to sleep overcomes a drowsy driver.

Driver distraction can be caused by any cognitive process such as
daydreaming, mind wandering, logical and mathematical problem solving,
decision making, using any kind of in-vehicle system, for example,
for entertainment, navigation, communication (including a cell phone),
and any other activity that may affect the driver's attention to driving~\citep{Almahasneh2014Deep}.
It is helpful to distinguish between four types of distractions~\citep{Dong2011Driver,Goncalves2015Driver}:
(1) \textit{manual} distraction (\eg, manually adjusting the volume
of the radio), (2) \textit{visual} distraction (\eg, looking away
from the road), (3) \textit{auditory} distraction (\eg, answering
a ringing cell phone), and (4) \textit{cognitive} distraction (\eg,
being lost in thought). Several distracting activities may, however,
involve more than one type of distraction (\eg, talking on the phone
while driving creates at least an auditory distraction and a cognitive
distraction, under the assumption that a hands-free system is used,
thereby avoiding manual distraction).

When distracted, the driver looses awareness of the current driving
situation. Being aware of a situation (whether for driving or for
some other activity) is often called \nomenclature{SA}{situational awareness}situational
awareness (SA). A loss of SA while driving results in a reduction
of vigilance and in an increase of the risk of accident. In driving,
a major aspect of SA is the ability to scan the driving environment
and to sense dangers, challenges, and opportunities, in order to maintain
the ability to drive safely. As a driver moves through the environment,
\he must---to avoid getting into an accident---identify the relevant
information in rapidly changing traffic conditions (\eg, distance
to other vehicles, closing speed), and be prepared to react to suddenly-appearing
events (\eg, braking because of an obstacle, obeying a road sign).
To achieve SA, a driver must thus perceive correctly \his driving
environment~\citep{Durso1999Situation}, be attentive, and have a
working memory~\citep{Wickens2015Engineering}. It follows that any
distraction that harms the driver's attention may adversely impact
SA~\citep{Kass2007Effects}.

\citet{Kircher2016Minimum}~argue that existing definitions of distraction
have limitations because they are difficult to operationalize, and
they are either unreasonably strict and inflexible or suffering from
hindsight bias, the latter meaning that one needs to know the outcome
of the situation to be able (1) to tell what the driver should have
paid attention to and, then, (2) to judge whether \he was distracted
or not. The authors are also concerned that distraction-detection
algorithms (1) do not take into account the complexity of a situation,
and (2) generally cover only \nomenclature{EOR}{eyes-off-road}eyes-off-road
(EOR) and engagement in non-driving related activities (NDRA). They
thus developed a theory, named MiRA (minimum required attention),
that defines the attention of a driver in \his driving environment,
based on the notion of SA. Instead of trying to assess distraction
directly, one does it indirectly, by first trying to assess attention.
Recall that distraction is a form of inattention.

According to the MiRA theory, a driver is considered attentive at
any time when \he samples sufficient information to meet the demands
of the driving environment. This means that a driver should be classified
as distracted only if \he does not fulfill the minimum attentional
requirements to have sufficient SA. This occurs when the driver does
not sample enough information, whether or not simultaneously performing
an additional task. This theory thus acknowledges (1) that a driver
has some spare capacity at \his disposal in the less complex driving
environments, and (2) that some glances toward targets other than
the roadway in front of \him may, in some situations, be needed for
the driving task (like looking at, or for, a vehicle coming from each
of the branches at a crossroad). This means that EOR and engagement
in NDRA do not necessarily lead to driver distraction.

The MiRA theory does not conform to the traditional types of distraction
(manual, visual, auditory, cognitive) as it does not prescribe what
sensory channel a certain piece of information must be acquired through.

In an attempt to operationalize the MiRA theory, \citet{Ahlstrom2021Towards}~present
an algorithm for detecting driver distraction that is context dependent
and uses (1) eye-tracking data registered in the same coordinate system
as an accompanying model of the surrounding environment and (2) multiple
buffers. Each buffer is linked to a corresponding glance target of
relevance. Such targets include: windshield, left and right windows,
(rear-view) mirrors, and instrument cluster. Some targets and their
buffers are always present (like the roadway ahead via the windshield,
and behind via the mirrors), while some other targets and their buffers
appear as a function of encountered traffic-regulation indications
and infrastructural features. Each buffer is periodically updated,
and its update rate can vary in time according to requirements that
are either “static” (e.g., the presence of a specific on-ramp that
requires one to monitor the sides and mirrors) or “dynamic” (e.g.,
a reduced speed that lessens the need to monitor the speedometer).
At each scheduled update time, a buffer is incremented if the driver
looks at the corresponding target, and decremented otherwise; this
is a way of quantifying the “sampling” (of the environment) performed
by the driver. A buffer running empty is an indication that the driver
is not sampling enough the corresponding target; \he is then considered
to be inattentive (independently of which buffer has run empty). Until
declared inattentive, \he is considered attentive.

This completes the background information on the \construct of distraction.
We now successively consider the four types of distraction. For each
of the four corresponding substates, we provide a detailed description,
and we then present the indicators and sensors that can be used to
characterize it.

\subsection{\Construct 3.1: Manual distraction\label{subsec:manual-distraction}}

\subsubsection{Description}

Manual distraction, also called biomechanical distraction, occurs
when the driver is taking one or both of \his hands off the steering
wheel. The driver may do so to answer a call or send a text message,
grab food and eat, or grab a beverage and drink, all while driving.
According to the \nomenclature{NHTSA}{National Highway Traffic Safety Administration}National
Highway Traffic Safety Administration (NHTSA), texting while driving
is the most alarming distraction. It is mainly due to manual distraction,
but, inevitably, it also includes both visual distraction and cognitive
distraction.

\subsubsection{Indicators}

Unsurprisingly, the best indicator used to detect manual distraction
is the behavior of the driver's hands, mainly through their positions
and movements. For safe driving, these hands are expected to be, most
of the time, exclusively on the steering wheel, the gearshift, or
the turn-signal lever. On the contrary, a hand using a phone, adjusting
the radio, or trying to grab something on the passenger seat indicates
a manual distraction~\citep{Tijerina2000Issues}.

Vehicle-based indicators can also be used, as shown in~\citep{Li2017Visual}.
Using naturalistic-driving data, the authors studied the correlation
between (1) performance metrics linked to the steering-wheel behavior
and to the vehicle speed, and (2) manual and visual driver distractions
induced, for example, by texting. They found a good correlation between
the steering movements and the manual-visual distraction of the driver.

The above information allows one to fill, in Table~\ref{tab-indicators},
the relevant cells of the ``Manual distraction'' column.

\subsubsection{Sensors}

The most common solution to analyze the behavior of the driver's hands
is to use a camera placed inside the vehicle, usually near the central
mirror, looking down in the direction of the driver.

\citet{Le2016Robust,Le2017Robust} propose an approach to detecting~\citep{Le2016Robust}
and classifying~\citep{Le2017Robust} human-hand regions in a vehicle
using CNNs. Their technique for hands detection is robust in difficult
conditions caused, for example, by occlusions, low resolution, and/or
variations of illumination.

Using deep CNNs, \citet{Yan2016Driver} classify six actions involving
the driver's hands, \ie, calling, eating, smoking, keeping hands
on the steering wheel, operating the gearshift, and playing on the
phone. Similarly, both \citet{Baheti2018Detection} and \citet{Masood2018Detecting}
use ten classes to detect when the driver is engaged in activities
other than safe driving, and to identify the cause of distraction.

Vehicle-based indicators can be obtained from the CAN bus of the vehicle~\citep{Fridman2019MIT,Li2008Design}.

The above information allows one to fill the relevant cells of Table~\ref{tab-sensors}.

\subsection{\Construct 3.2: Visual distraction\label{subsec:visual-distraction}}

\subsubsection{Description}

Visual distraction occurs when the driver is looking away from the
road scene, even for a split second. It is often called EOR, and is
one of the most common distractions for a driver. Examples of activities
causing EOR are (1) adjusting devices in the vehicle (like a radio
or navigation system), (2) looking towards other seats, (3) regarding
a new message on the phone or glancing at the phone to see who is
calling, and (4) looking outside when there is a distraction by the
roadside. All generally result in the driver not looking straight
ahead, which is what \he needs to be doing for safe driving.

\subsubsection{Indicators}

The gaze is the main indicator used to detect a visual distraction
of a driver. The duration of EOR is probably the most-used metric.
The longer the EOR duration is, the lower the SA of the driver is,
and the higher the visual distraction of the driver is~\citep{Young2007Driver}.
The glance pattern and the mean glance duration are other metrics~\citep{Ranney2000NHTSA}.

Sometimes, the head direction is used to approximate the gaze direction
in order to characterize the driver visual distraction~\citep{Fridman2016Driver,Fridman2016Owl}.
For example, \citet{Fridman2016Driver} classify driver gaze regions
on the sole basis of the head pose of the driver. \citet{Fridman2016Owl}
compare classifications of driver gaze using either head pose alone
or both head pose and eye gaze. They classify, based on facial images,
the focus of the attention of the driver using $6$ gaze regions (road,
center stack, instrument cluster, rear-view mirror, left, and right).
To do so, they consecutively perform face detection, face alignment,
pupil detection, feature extraction and normalization, classification,
and decision pruning. \citet{Vicente2015Driver} similarly classify
the driver gaze, but use $18$ regions instead of $6$.

Visual distraction can also be inferred using vehicle-based indicators
such as wheel steering, braking behavior, and speed. Indeed, a driver
generally slows down when distracted by a visual stimulus~\citep{Engstrom2007Effects,Yusoff2017Selection},
and visual distraction impairs lateral control because the driver
needs to compensate for errors made when taking \his eyes off the
road, which leads to larger deviations in lane positioning~\citep{Liang2010Combining,Yusoff2017Selection}.
Such deviations have various causes, including drowsiness and visual
distraction. This re-emphasizes the need to use as many indicators
as possible. This also explains why more and more vehicles are equipped
with systems that keep the vehicle within its lane whenever possible.

The above information allows one to fill, in Table~\ref{tab-indicators},
the relevant cells of the ``Visual Distraction'' column.

\subsubsection{Sensors}

In order to monitor driver visual distraction, one mainly uses at
least one camera facing the driver, thus as for manual distraction.
The camera can be placed in various positions as long as the head
pose and/or gaze of the driver can be obtained.

\citet{Naqvi2018DeepLearning} use a \nomenclature{NIR}{near-infrared}near-infrared
(NIR) camera (with wavelengths of $0.75-1.4\,\micrometer$) placed
in the dashboard in conjunction with a deep-learning-based gaze detection
system, classifying the driver gaze into $17$ gaze zones.

\citet{Mukherjee2015Deep}, similarly to~\citet{Fridman2016Driver},
present a CNN-based model to estimate human head pose and to classify
human gaze direction. They use, however, low-resolution \nomenclature{RGB-D}{RGB-depth}RGB-depth
(RGB-D), thus with a camera providing depth information.

The above information allows one to fill the relevant cells of Table~\ref{tab-sensors}.

\subsection{\Construct 3.3: Auditory distraction\label{subsec:auditory-distraction}}

\subsubsection{Description}

Auditory distraction occurs when some sound prevents the driver from
making the best use of \his hearing, because \his attention is drawn
to the source of the sound. Hearing a phone ringing, listening to
a passenger, listening to music, and following navigation instructions
can all lead to auditory distraction.

This component of driver distraction is the least studied in the literature,
likely because (1) it is often accompanied by at least one other more-easily
detectable source of distraction falling among the other three types,
and (2) it poses lower safety risks in comparison to the other types
of distraction, in particular visual distraction~\citep{Sodnik2008User}.

The literature does not appear to introduce the concept of “auditory
indicators”, which would characterize (1) the sounds captured both
inside and outside of the vehicle, and, preferably, (2) the distraction
they create. By using several microphones (including arrays thereof),
and techniques for separating audio sources~\citep{Vincent2006Performance},
one could imagine breaking down and localizing the various sources
of sounds both inside and outside the vehicle.

\subsubsection{Indicators}

When the driver appears to be auditorily distracted, there occur changes
in pupil diameter~\citep{Goncalves2015Driver,Kahneman1969Pupillary}
and blink frequency~\citep{Goncalves2015Driver,Hargutt2001Eyelid}.
Brain activity (EEG)~\citep{Schroger2000Auditory} can also be used
as an indicator of auditory distraction. \citet{Sonnleitner2014EEG}
describe the impact of an auditory secondary task on a driver during
a primary driving task, and show changes in braking reaction and brain
activity.

The above information allows one to fill, in Table~\ref{tab-indicators},
the relevant cells of the ``Auditory distraction'' column.

\subsubsection{Sensors}

As already indicated, obtaining the pupil diameter is challenging
in real conditions due to illumination conditions and/or camera resolution,
among others. Furthermore, brain activity cannot, at this time, be
measured both in real time and in a non-intrusive, reliable way. Blink
frequency can, however, be monitored via a camera, and braking behavior
via the CAN bus.

Although microphones and, even better, arrays thereof, both inside
and outside the vehicle, would be natural sensors to provide values
for auditory indicators, we did not find any references considering
such sensors for characterizing auditory distraction. One can also
envision using the microphone(s) of a smartphone linked to a DMS.

The above information did not lead to the addition of any reference
to Table~\ref{tab-sensors}.

\subsection{\Construct 3.4: Cognitive distraction\label{subsec:cognitive-distraction}}

\subsubsection{Description}

In the context of driving, cognitive distraction is defined by \citet{NHTSA2010Overview}
as the mental workload associated with a task that involves thinking
about something other than the (primary) driving task. A driver who
is cognitively distracted due to a secondary task, such as mind wandering,
experiences an increase in \his mental workload (the \construct
discussed in Section~\ref{sec:mental-workload}). The characterization
of \his cognitive distraction could therefore be achieved (1) by
examining how \his mental workload evolves over time and (2) by finding
characteristics of this evolution allowing one to decide whether or
not it is caused by cognitive distraction. The monitoring of cognitive
distraction is thus, before all, a monitoring of the mental workload
and/or its time variations. Section~\ref{sec:mental-workload} shows
that there are (1) many ways to characterize mental workload, and
(2) many indicators thereof. The challenge is to be able to pinpoint
the components of, or changes in, the mental workload that are due
to distraction.

Cognitive distraction occurs when a driver is thinking about something
that is not related to the driving task. In the driving context, while
visual distraction can be summarized by EOR, cognitive distraction
can similarly be viewed as \nomenclature{MOR}{mind-off-road}``mind-off-road''
(MOR). While it is relatively easy to monitor EOR (with a camera facing
the driver), it is difficult to monitor MOR. It has, however, been
shown that, when a driver is cognitively distracted, \his visual
behavior is impacted. Mind-wandering and daydreaming are two causes
of cognitive distraction.

\subsubsection{Indicators}

As cognitive distraction induces mental workload, the indicators allowing
one to detect and characterize these two \constructs are similar,
if not identical. Therefore, it is difficult, if not impossible, to
distinguish, in the driving context (as well as others), between these
two \constructs since they have nearly the same influences on the
indicators.

Among the four types of distractions, cognitive distraction has proven
to be the most difficult to detect and characterize. This is because
it happens inside the brain, and, obviously, ``observing'' the brain
of a driver is more challenging than observing \his hands and eye(s).

As for visual distraction, cognitive distraction can be characterized
by indicators of both driving performance and eye movements~\citep{Liao2016Detection},
including (1) vehicle-based indicators, such as speed~\citep{Ranney2008Driver},
wheel steering~\citep{Liang2010Combining}, lane discipline~\citep{Liang2010Combining,Ranney2008Driver,Strayer2013Measuring},
and braking behavior~\citep{Harbluk2007AnOnRoad}, and (2) driver-based,
behavioral indicators, such as gaze parameters (\eg, fixation duration,
glance frequency, and gaze distribution)~\citep{Harbluk2007AnOnRoad,Liang2007Real,Son2018Evaluation,Strayer2015Assessing}
and head orientation. A driver makes significantly fewer high-speed
saccadic eye movements and spends less time looking to the relevant
periphery for impending hazards with increasing complexity of the
secondary task(s). \He also spends less time checking \his instruments
and mirrors~\citep{Harbluk2007AnOnRoad}.

Cognitive distraction can also be measured through a variety of driver-based,
physiological indicators. Among these, brain activity~\citep{Strayer2007Cell}
and pupil diameter may be the most convincing. Studies of EDA and
HR show only weak relationships between these indicators and cognitive
distraction~\citep{Yusoff2017Selection}.

Among the subjective measures, the NASA TLX~\citep{Hart1988Development}
is commonly used in driving-distraction studies even though it is
a subjective measure of mental workload, and, thus, not a measure
specific to cognitive distraction.

The above information allows one to fill, in Table~\ref{tab-indicators},
the relevant cells of the ``Cognitive distraction'' column.

\subsubsection{Sensors}

Since the main indicators of cognitive distraction are driving performance
and gaze parameters, the main sensors to characterize it are vehicle-centric
sensors, and cameras.

The above information did not lead to the addition of any reference
to Table~\ref{tab-sensors}.

\section{\Construct 4: Emotions\label{sec:emotions}}

We provide a detailed description of (the state of) “emotions”, and
we then present the indicators and sensors that can be used to characterize
it.

\subsection{Description}

While the concept of emotions is familiar to most people, it is difficult
to define. Emotions are associated with a strong feeling deriving
from one's circumstances, mood, and/or relationships with other people.
In the driving context, the emotions most commonly monitored for safety
purposes are stress and anger, as they have a negative impact on driving,
and create dangers~\citep{Hu2013Negative,Pecher2010Influence}.

Stress is a state of physical, emotional, or psychological tension
resulting from adverse or demanding circumstances. In biology, stress
is defined as a state of homeostasis being challenged due to a stressor~\citep{Lu2021TheEvolution}.

Anger is a strong feeling of annoyance, displeasure, and/or hostility.
It is a common negative emotion in the context of driving, where it
is often called road rage~\citep{Hu2018Analysis}.

\subsection{Indicators}

Emotion recognition is currently a hot topic in the field of affective
computing, and is gaining interest in the field of \nomenclature{ADAS}{advanced driver-assistance system}advanced
driver-assistance systems (ADASs). To recognize emotions, one can
use various behavioral features, for example, speech~\citep{Gavrilescu2019Feedforward}
and facial expressions~\citep{Ekman1993Facial,Russell1994IsThere}.

Among the driver-based indicators of both stress and anger, physiological
indicators are commonly used. Stress causes physiological responses~\citep{Diverrez2016Stress},
such as variations or modifications in HR~\citep{Diverrez2016Stress,Healey2005Detecting,DeSantos2011Stress,Zhao2016Emotion},
breathing activity~\citep{Diverrez2016Stress,Healey2005Detecting},
blood pressure, EDA~\citep{Healey2005Detecting,DeSantos2011Stress,Shi2007Galvanic},
and pupil activity~\citep{Partala2003Pupil}. The two physiological
features that exhibit the highest correlations with driver stress
are HR and EDA~\citep{Healey2005Detecting}.

For anger in the driving context, \citet{Wan2017Road} suggest to
identify it based on physiological indicators such as HR, EDA, breathing
rate, and EEG, with the obvious, current, practical limitations for
the latter.

The \nomenclature{SAM}{self-assessment manikin}self-assessment manikin
(SAM)~\citep{Bradley1994Measuring} is a subjective assessment technique
to characterize emotions.

The above information allows one to fill, in Table~\ref{tab-indicators},
the relevant cells of the ``Emotions'' column.

\subsection{Sensors}

The development of wearable devices with physiological sensors facilitates
the recognition of emotions in real-driving conditions, thus outside
of a laboratory context.

Facial expressions constitute a good indicator of emotions. The analysis
and recognition of facial expressions is currently a field of great
interest in scientific research~\citep{Li2017Multimodal,Zhang2016Deep}.
Facial expressions can be monitored in a vehicle via the use of a
camera facing the driver~\citep{Gao2014Detecting,Jeong2018Driver,Melnicuk2017Employing}.
Indeed, \citet{Jeong2018Driver} recently developed an algorithm for
monitoring the emotions of a driver based on the analysis of facial
expressions. Using DNNs performing \nomenclature{FER}{facial-expression recognition}facial-expression
recognition (FER), they can identify---in real time and in real-driving
situations---anger, disgust, fear, happiness, sadness, and surprise.
A smartphone with a camera facing the user can be used for FER, here
for estimating \his emotional state \citep{Melnicuk2017Employing}.

\nomenclature{FIR}{far infrared}Far-infrared (FIR) imaging (with
wavelengths of $15-1000\,\micrometer$), also called \nomenclature{IRT}{infrared thermography}infrared
thermography (IRT), can be used to quantify stress and emotions by
monitoring the breathing activity~\citep{Murthy2004Touchless}. This
can be done via the use of an IRT camera facing the driver.

The recognition of emotions can also be done using wearable sensors~\citep{Ragot2017Emotion}
such as the E4 wristband, which is a wearable research device that
provides the means to acquire physiological data in real time. Many
studies~\citep{Gouverneur2017Classification,Ollander2016Comparison,Sevil2017Social}
have indeed shown that one can detect stress by using the physiological
data that this device provides, in particular HR and EDA data.

\citet{Boril2012Towards} developed a stress detector employing a
combination of the driver's speech and some CAN-bus parameters, mainly
the steering-wheel angle and the speed. \citet{Basu2017AReview} review
various methods (that are not specific to the field of driving) for
recognizing emotions from speech. \citet{Zhang2018Speech} explore
how to utilize a deep CNN for the same purpose.

The above information allows one to fill the relevant cells of Table~\ref{tab-sensors}.

\section{\Construct 5: Under the influence\label{sec:under-the-influence}}

We provide a detailed description of (the state of) “under the influence”,
and we then present the indicators and sensors that can be used to
characterize it.

\subsection{Description}

\nomenclature{DUI}{driving under the influence}Driving under the
influence (DUI)---also called \nomenclature{DWI}{driving while intoxicated}driving
while intoxicated (DWI) and impaired driving---refers to the driving
of a vehicle by a person who has consumed a quantity of alcohol or
drugs (including prescription medication) that causes \him to function
in an impaired way. If the impaired driving is due only to alcohol,
one also talks about drunk driving. While DUI is obviously dangerous,
it is also illegal in most countries to drive under the influence
of alcohol, cannabis (or marijuana), opioids, methamphetamines, and
any potentially-impairing drug (\eg, a psychoactive drug), whether
prescribed or over-the-counter.

A psychoactive drug, also called a psychotropic drug, is a chemical
substance that changes a person's mental state and results in alterations
in perception, mood, and/or consciousness. Based on their effects,
psychoactive drugs can be classified into the three main categories
of stimulants, depressants, and hallucinogens \citep{Marillier2019Driving,Zapata2021Chemical}.
Yet, some drugs may fall under different categories at different times
(for example, cannabis is both a depressant drug and a hallucinogen
drug). Stimulants (\eg, methamphetamines, cocaine) speed up the activity
of the central nervous system, often resulting in the user feeling
more alert, euphoric, and energetic. Depressants (\eg, heroin) slow
down the activity of the central nervous system, often resulting in
the user feeling more relaxed, sleepier, and insensitive to pain.
Hallucinogens (\eg, LSD) are psychoactive substances that alter human
sensory perceptions in such a way that the user perceives a distorted
reality in which time, space, colors, and forms are altered.

The substances that are most frequently detected in impaired drivers
are alcohol followed by cannabis. Studies have shown that more than
one-third of adults and more than half of teenagers admit to DUI of
alcohol at some point in their lives \citep{Alonso2019Driving}. Alcohol
is a depressant drug that affects the central nervous system and slows
down brain functions. Any amount of alcohol can affect a person’s
abilities (1) by degrading attention, perception, information processing
skills, memory, reasoning, coordination, motor skills, and reaction
time, and (2) by altering the five senses and the emotions \citep{Alonso2015Driving,Attia2016Electronic,Berman1997Impairments,Garrisson2021TheEffects}.
A person's alcohol level is measured by the weight of the alcohol
in a specified volume of blood, called \nomenclature{BAC}{blood alcohol concentration}blood
alcohol concentration (BAC) and measured in grams of alcohol per deciliter
($\gdl$) of blood. According to NHTSA, the effects of alcohol vary
with BAC in the way shown in Table~\ref{tab-effects-of-BAC}, in
Appendix~\ref{appendix:effects-BAC}, and the risk of having an accident
after consuming alcohol increases exponentially as a function of BAC.
For example, every additional $0.08$~$\text{g}$ of alcohol per
deciliter ($\dl$) of blood multiplies by four the risk of accident
\citep{Alonso2019Driving}. According to the World Health Organization~\citep{WHO2018Global},
best practice for drunk--driving laws includes a BAC limit of $0.05\,\gdl$
for the general population and of $0.02\,\gdl$ for young or novice
drivers. Although studies show considerable differences among individuals
regarding their responses to alcohol consumption~\citep{Christoforou2013Reaction},
young drivers experience significantly stronger effects, putting them
at greater risk of accidents~\citep{Peck2008TheRelationship,Zador2000Alcohol}.
Hangovers, \ie, the after-effects occurring as a result of heavy
drinking and as the BAC subsequently approaches zero, are, however,
known to also affect the performance of daily-life tasks, such as
driving, by impairing cognitive functions, such as memory, psychomotor
speed, and sustained attention \citep{Gunn2018ASystematic,Verster2014Effects}.

\subsection{Indicators}

Several physiological indicators are used to monitor DUI such as heart
activity \citep{Berman1997Impairments,PAHO2018Drinking}, breathing
activity \citep{PAHO2018Drinking}, body temperature \citep{Berman1997Impairments,Rosero-Montalvo2021Hybrid},
and pupil diameter \citep{Rosero-Montalvo2021Hybrid}. Alcohol is
known to increase HR and breathing rate \citep{PAHO2018Drinking}.
Cannabis is known to increase HR and breathing difficulty. Alcohol
increases the activity of arteries and other blood vessels, therefore
increasing the temperature of the face of a drunk person \citep{Rosero-Montalvo2021Hybrid}.
The variations of temperature are visible on the nose, eyebrows, chin,
and forehead. When people drink alcohol, their irises become darker,
because the sclera is replete with blood vessels that increase in
temperature with alcohol consumption. In a sober person, the temperatures
of the sclera and the iris are the same, but with alcohol intoxication,
the temperature of the sclera increases compared to the one of the
iris because of the denser blood-vessel network in the sclera.

Behavioral indicators of DUI include parameters of gaze (due to the
impairment of some visual functions) and of slurred speech \citep{PAHO2018Drinking}.
Drunk speakers may use prosodic contours differently from sober speakers,
using more or less speech emphasis. Drunk speakers may pronounce words
differently, choose certain pronunciation variants more frequently
than others, and may even select more frequently certain words, the
latter affecting the phonotactic patterns \citep{Sanghvi2018Drunk}.

\citet{NHTSA1998TheVisual} defines four categories of cues to predict
that a driver is DUI, namely problems in (1) maintaining proper lane
position (\eg, weaving, drifting, swerving), (2) controlling speed
and brakes (\eg, varying speed, abnormally driving at low speed,
stopping beyond a limit line), (3) maintaining vigilance (\eg, driving
erroneously in opposing lanes, responding slowly to traffic signals),
and (4) exercising proper judgment (\eg, following too closely, turning
illegally). In congruence with the indication by NHTSA that a drunk
driver is prone to weaving, drifting, and swerving (and thus to having
difficulty keeping \his vehicle in the center of the lane), an increase
in SDLP is recognized in the literature to be an indicator of DUI
of alcohol \citep{Irwin2017Effects,Martin2013AReview,Mets2011Effects}
and hangovers \citep{Verster2014Effects}. Speed and acceleration
are other indicators, as drunk drivers often experience difficulty
in keeping an appropriate speed, with abrupt accelerations or decelerations,
erratic brakings, and jerky stops \citep{Irwin2017Effects,Mets2011Effects}.

The above information allows one to fill, in Table~\ref{tab-indicators},
the relevant cells of the ``Under the influence'' column.

\subsection{Sensors}

In police operations, alcohol levels are typically measured with a
breathalyzer using air exhaled through the mouth. The amount of alcohol
in breath can then be used to determine the BAC \citep{PAHO2018Drinking}.
If this BAC is above the legally authorized value, the results can,
if desired, be confirmed by a blood test. With just $100$ microliter
($\microliter$) of collected blood, one can, not only measure the
BAC precisely, but also identify and quantify $37$ substances that
are of interest in the context of drug-impaired driving \citep{Joye2020Driving}.
Many people, however, drive under the influence without necessarily
being stopped and checked by police every time they do so.

To solve the issue of DUI, the literature commonly suggests the use
of ignition-interlock devices \citep{Attia2016Electronic,Nair2017Drunk,Ray2017Prevention}.
When a driver enters \his vehicle, \he must provide a breath sample,
and an alcohol sensor then determines whether \he is drunk (\ie,
has a BAC above a specified threshold). If this is the case, the ignition-control
system prevents the driver from starting the engine. Ignition-interlock
devices are usually installed in the vehicles of people with prior
DUI convictions and in long-haul, commercial vehicles, for example,
trucks and buses \citep{Alonso2019Driving}. This solution does not,
however, allow for the real-time monitoring of the state of the driver,
and does not prevent the driver from drinking alcohol after starting
the engine.

To counter this problem, \citet{Sakairi2012Water} developed a system
using a water-cluster-detecting (WCD) breath sensor that can detect
breath from a distance of about $0.5\,\m$, allowing one to monitor
the driver's alcohol level while \he is operating \his vehicle.
The sensor detects breath by separating positively-charged water clusters
in breath from negatively-charged ones by using an electric field
and by measuring the two corresponding electric currents.

The detection of individuals DUI of alcohol can also be achieved based
on the heart activity. Indeed, \citet{Kojima2009Noninvasive} and
\citet{Murata2011Noninvasive} constructed a seat incorporating an
air-pack sensor that monitors, via a body-trunk plethysmogram, both
the heart activity and the breathing activity. The analysis, during
$5\,\minute$, of the extracted body-trunk plethysmogram signal, called
the air-pack pulse wave, reveals differences due to the consumption
of alcohol, allowing one to distinguish between sobriety and intoxication.
\citet{Wu2016AWearable,Wu2016APrecise} propose to use a wearable
ECG sensor, and an SVM to classify the corresponding ECG data as sober
or intoxicated.

Recognizing whether drivers are DUI of alcohol can also be achieved
using a camera that acquires IR images \citep{Hermosilla2018Face,Koukiou2017Local,Menon2019Driver}.
For an intoxicated person, vessels on the forehead become more active
so that, in an IR image, the intensities of the pixels in this region
are affected accordingly. \citet{Menon2019Driver} developed a system
that uses IR images of the driver's face in order to classify \him
as sober or drunk. The system successively (1) locates the face using
a CNN, and (2) performs the binary classification based on differences
in blood temperatures at 22 points on the face of the driver using
a supervised-learning-classification algorithm based on a probabilistic
model called Gaussian-mixture model.

\citet{Rosero-Montalvo2021Hybrid} introduce a non-invasive system
incorporating a gas sensor, a temperature sensor, and a camera to
identify a person having alcohol in the blood, through supervised
classification of the data from (1) the two sensors and (2) the results
of the analysis of the camera output via computer vision. The authors
use the concentration of alcohol in the vehicle environment, the facial
temperature of the driver, and the diameters of \his pupils.

According to NHTSA and its four, above-mentioned cues that a driver
is DUI, vehicle-based indicators and related vehicle-centric sensors
are of interest. Relevant CAN-bus parameters, and indicators such
as wheel steering and lane discipline, are widely used to detect instances
of DUI \citep{Berri2018ANonintrusive,ElBasiouniElMasri2017Toward,Harkous2018Application,Harkous2019ATwoStage,Li2015Drunk,Shirazi2014Detection}.
\citet{Harkous2018Application} identify drunk-driving behaviors using
HMMs based on vehicle-sensors data, available via the CAN bus. They
use wheel-steering parameters, speed, and lateral position as indicators.
They found that longitudinal-acceleration sensors achieve the best
average classification accuracy for distinguishing between sobriety
and intoxication. \citet{Harkous2019ATwoStage} extend the above work
by replacing each HMM by a \nomenclature{RNN}{recurrent neural network}recurrent
neural network (RNN). Likewise, \citet{Berri2018ANonintrusive} use
features such as speed, acceleration, braking, steering wheel angle,
distance to the center lane, and geometry of the road (straight or
curved) to detect DUI of alcohol. Their system can also be used to
detect the presence of any psychoactive drug that can cause a driver
to have abnormal driving behaviors. To detect an intoxicated driver,
\citet{Dai2010Mobile} describe a solution that only requires a mobile
phone placed in the vehicle. Using the phone's accelerometers, they
analyze the longitudinal and lateral accelerations of the vehicle
to detect any abnormal or dangerous driving maneuvers typically related
to DUI of alcohol.

The above information allows one to fill the relevant cells of Table~\ref{tab-sensors}.

\begin{table}
\caption{Detailed ``\constructs vs indicators'' table, introduced in simplified
form in Figure~\ref{fig:axes-analysis}. Each cell in the heart of
the table gives some references (if any) discussing how the corresponding
indicator is useful for characterizing the corresponding \construct.\label{tab-indicators}}

\raggedright{}\resizebox{\textwidth}{!}{{\scriptsize{}}%
\begin{tabular}{|l|l|l|l|c|>{\centering}p{1cm}|c|>{\centering}p{1cm}|c|c|>{\centering}p{1cm}|>{\raggedright}m{1.2cm}|}
\cline{5-12} \cline{6-12} \cline{7-12} \cline{8-12} \cline{9-12} \cline{10-12} \cline{11-12} \cline{12-12} 
\multicolumn{4}{c|}{} & \multicolumn{8}{c|}{\textbf{\tiny{}\Constructs}}\tabularnewline
\cline{5-12} \cline{6-12} \cline{7-12} \cline{8-12} \cline{9-12} \cline{10-12} \cline{11-12} \cline{12-12} 
\multicolumn{4}{c|}{} & \multirow{2}{*}{\textbf{\tiny{}Drowsiness}} & \multirow{2}{1cm}{\centering{}\textbf{\tiny{}Mental workload}} & \multicolumn{4}{c|}{\textbf{\tiny{}Distraction}} & \multirow{2}{1cm}{\centering{}\textbf{\tiny{}Emotions}} & \multirow{2}{1.2cm}{\centering{}\textbf{\tiny{}Under the influence}}\tabularnewline
\cline{7-10} \cline{8-10} \cline{9-10} \cline{10-10} 
\multicolumn{4}{c|}{} &  &  & \multirow{1}{*}{\textbf{\tiny{}Manual}} & \multirow{1}{1cm}{\centering{}\textbf{\tiny{}Visual}} & \multirow{1}{*}{\textbf{\tiny{}Auditory}} & \multicolumn{1}{c|}{\textbf{\tiny{}Cognitive}} &  & \tabularnewline
\hline 
\multirow{24}{*}{\begin{turn}{90}
\textbf{\tiny{}Indicators}
\end{turn}} & \multirow{14}{*}{\begin{turn}{90}
\textbf{\tiny{}Driver}
\end{turn}} & \multirow{6}{*}{\begin{turn}{90}
\textbf{\tiny{}Physiological}
\end{turn}} & \multicolumn{1}{>{\raggedright}m{0.08\textwidth}|}{\textbf{\tiny{}Heart activity}} & {\tiny{}\citep{JacobedeNaurois2019Detection,Persson2021Heart,Vicente2016Drowsiness}} & \centering{}{\tiny{}\citep{Fournier1999Electrophysiological,Gable2015Comparing,Paxion2014Mental,Reimer2009Road}} &  &  &  & {\tiny{}\citep{Yusoff2017Selection}} & {\tiny{}\citep{DeSantos2011Stress,Diverrez2016Stress,Healey2005Detecting,Zhao2016Emotion,Wan2017Road}} & \centering{}{\tiny{}\citep{Berman1997Impairments,PAHO2018Drinking}}\tabularnewline
\cline{4-12} \cline{5-12} \cline{6-12} \cline{7-12} \cline{8-12} \cline{9-12} \cline{10-12} \cline{11-12} \cline{12-12} 
 &  &  & \multicolumn{1}{>{\raggedright}m{0.08\textwidth}|}{\textbf{\tiny{}Breathing activity}} & {\tiny{}\citep{Kiashari2020Evaluation,JacobedeNaurois2019Detection}} & \centering{} &  &  &  &  & {\tiny{}\citep{Diverrez2016Stress,Healey2005Detecting,Wan2017Road}} & \centering{}{\tiny{}\citep{PAHO2018Drinking}}\tabularnewline
\cline{4-12} \cline{5-12} \cline{6-12} \cline{7-12} \cline{8-12} \cline{9-12} \cline{10-12} \cline{11-12} \cline{12-12} 
 &  &  & \multicolumn{1}{>{\raggedright}m{0.08\textwidth}|}{\textbf{\tiny{}Brain activity}} & {\tiny{}\citep{Akerstedt1990Subjective}} & \centering{}{\tiny{}\citep{Kim2013Highly}} &  &  & {\tiny{}\citep{Schroger2000Auditory,Sonnleitner2014EEG}} & {\tiny{}\citep{Strayer2007Cell}} & {\tiny{}\citep{Wan2017Road}} & \centering{}\tabularnewline
\cline{4-12} \cline{5-12} \cline{6-12} \cline{7-12} \cline{8-12} \cline{9-12} \cline{10-12} \cline{11-12} \cline{12-12} 
 &  &  & \multicolumn{1}{>{\raggedright}m{0.08\textwidth}|}{\textbf{\tiny{}Electrodermal activity}} & {\tiny{}\citep{Michael2012Electrodermal}} & \centering{}{\tiny{}\citep{Kajiwara2014Evaluation}} &  &  &  & {\tiny{}\citep{Yusoff2017Selection}} & {\tiny{}\citep{DeSantos2011Stress,Healey2005Detecting,Shi2007Galvanic,Wan2017Road}} & \centering{}\tabularnewline
\cline{4-12} \cline{5-12} \cline{6-12} \cline{7-12} \cline{8-12} \cline{9-12} \cline{10-12} \cline{11-12} \cline{12-12} 
 &  &  & \textbf{\tiny{}Body temperature} &  & \centering{} &  &  &  &  &  & \centering{}{\tiny{}\citep{Berman1997Impairments,Rosero-Montalvo2021Hybrid}}\tabularnewline
\cline{4-12} \cline{5-12} \cline{6-12} \cline{7-12} \cline{8-12} \cline{9-12} \cline{10-12} \cline{11-12} \cline{12-12} 
 &  &  & \multicolumn{1}{>{\raggedright}m{0.08\textwidth}|}{\textbf{\tiny{}Pupil diameter}} & {\tiny{}\citep{Lowenstein1963Pupillary,Nishiyama2007ThePupil,Wilhelm1998Pupillographic}} & \centering{}{\tiny{}\citep{Gable2015Comparing,Kosch2018Look,Marquat2015Review,Pfleging2016AModel,Yokoyama2018Prediction}} &  &  & {\tiny{}\citep{Goncalves2015Driver,Kahneman1969Pupillary}} &  & {\tiny{}\citep{Partala2003Pupil}} & \centering{}{\tiny{}\citep{Rosero-Montalvo2021Hybrid}}\tabularnewline
\cline{3-12} \cline{4-12} \cline{5-12} \cline{6-12} \cline{7-12} \cline{8-12} \cline{9-12} \cline{10-12} \cline{11-12} \cline{12-12} 
 &  & \multirow{7}{*}{\begin{turn}{90}
\textbf{\tiny{}Behavioral}
\end{turn}} & \multirow{1}{*}{\textbf{\tiny{}Gaze parameters}} & {\tiny{}\citep{Bakker2021AMultiStage}} & \centering{}{\tiny{}\citep{Fridman2018Cognitive,Le2020Evaluating,Liao2016Detection,Marquat2015Review,May1990Eye}} &  & {\tiny{}\citep{Fridman2016Owl,Ranney2000NHTSA,Vicente2015Driver,Young2007Driver}} &  & {\tiny{}\citep{Harbluk2007AnOnRoad,Liang2007Real,Son2018Evaluation,Strayer2015Assessing}} &  & \centering{}{\tiny{}\citep{PAHO2018Drinking}}\tabularnewline
\cline{4-12} \cline{5-12} \cline{6-12} \cline{7-12} \cline{8-12} \cline{9-12} \cline{10-12} \cline{11-12} \cline{12-12} 
 &  &  & \textbf{\tiny{}Blink dynamics} & {\tiny{}\citep{Anund2008Driver,Bakker2021AMultiStage,Hultman2021Driver,JacobedeNaurois2019Detection,Lisper1986Relation,Schleicher2008Blinks}} & \centering{}{\tiny{}\citep{Marquat2015Review}} &  &  & {\tiny{}\citep{Goncalves2015Driver,Hargutt2001Eyelid}} &  & {\tiny{}\citep{Partala2003Pupil}} & \centering{}\tabularnewline
\cline{4-12} \cline{5-12} \cline{6-12} \cline{7-12} \cline{8-12} \cline{9-12} \cline{10-12} \cline{11-12} \cline{12-12} 
 &  &  & \textbf{\tiny{}PERCLOS} & {\tiny{}\citep{Bakker2021AMultiStage,Dinges1998Evaluation,Dinges1998PERCLOS,JacobedeNaurois2019Detection,Wierwille1994Research}} & \centering{}{\tiny{}\citep{Marquat2015Review}} &  &  &  &  &  & \centering{}\tabularnewline
\cline{4-12} \cline{5-12} \cline{6-12} \cline{7-12} \cline{8-12} \cline{9-12} \cline{10-12} \cline{11-12} \cline{12-12} 
 &  &  & \multicolumn{1}{>{\raggedright}m{0.08\textwidth}|}{\textbf{\tiny{}Facial expressions}} & {\tiny{}\citep{Bakker2021AMultiStage}} & \centering{} &  &  &  &  & {\tiny{}\citep{Ekman1993Facial,Russell1994IsThere}} & \centering{}\tabularnewline
\cline{4-12} \cline{5-12} \cline{6-12} \cline{7-12} \cline{8-12} \cline{9-12} \cline{10-12} \cline{11-12} \cline{12-12} 
 &  &  & \multicolumn{1}{>{\raggedright}m{0.08\textwidth}|}{\textbf{\tiny{}Body posture}} & {\tiny{}\citep{Bakker2021AMultiStage,JacobedeNaurois2019Detection}} & \centering{} &  & {\tiny{}\citep{Fridman2016Driver,Fridman2016Owl}} &  &  &  & \centering{}\tabularnewline
\cline{4-12} \cline{5-12} \cline{6-12} \cline{7-12} \cline{8-12} \cline{9-12} \cline{10-12} \cline{11-12} \cline{12-12} 
 &  &  & \multicolumn{1}{>{\raggedright}m{0.08\textwidth}|}{\textbf{\tiny{}Hands parameters}} &  & \centering{} & {\tiny{}\citep{Tijerina2000Issues}} &  &  &  &  & \centering{}\tabularnewline
\cline{4-12} \cline{5-12} \cline{6-12} \cline{7-12} \cline{8-12} \cline{9-12} \cline{10-12} \cline{11-12} \cline{12-12} 
 &  &  & \textbf{\tiny{}Speech} &  & \centering{} &  &  &  &  & {\tiny{}\citep{Gavrilescu2019Feedforward,Boril2012Towards}} & \centering{}{\tiny{}\citep{PAHO2018Drinking,Sanghvi2018Drunk}}\tabularnewline
\cline{3-12} \cline{4-12} \cline{5-12} \cline{6-12} \cline{7-12} \cline{8-12} \cline{9-12} \cline{10-12} \cline{11-12} \cline{12-12} 
 &  & \multicolumn{2}{l|}{\textbf{\tiny{}Subjective}} & {\tiny{}\citep{Akerstedt1990Subjective,Hoddes1973SSS,Monk1989VAS}} & \centering{}{\tiny{}\citep{Hart1988Development}} &  &  &  & {\tiny{}\citep{Hart1988Development}} & {\tiny{}\citep{Bradley1994Measuring}} & \centering{}\tabularnewline
\cline{2-12} \cline{3-12} \cline{4-12} \cline{5-12} \cline{6-12} \cline{7-12} \cline{8-12} \cline{9-12} \cline{10-12} \cline{11-12} \cline{12-12} 
 & \multirow{4}{*}{\begin{turn}{90}
\textbf{\tiny{}Vehicle}
\end{turn}} & \multicolumn{2}{l|}{\textbf{\tiny{}Wheel steering}} & {\tiny{}\citep{JacobedeNaurois2019Detection,Liang2019Prediction,Thiffault2003Monotony}} & \centering{}{\tiny{}\citep{Schaap2013Relationship}} & {\tiny{}\citep{Li2017Visual}} & {\tiny{}\citep{Li2017Visual}} &  & {\tiny{}\citep{Liang2010Combining}} & {\tiny{}\citep{Boril2012Towards}} & \centering{}\tabularnewline
\cline{3-12} \cline{4-12} \cline{5-12} \cline{6-12} \cline{7-12} \cline{8-12} \cline{9-12} \cline{10-12} \cline{11-12} \cline{12-12} 
 &  & \multicolumn{2}{l|}{\textbf{\tiny{}Lane discipline}} & {\tiny{}\citep{Bakker2021AMultiStage,Godthelp1984TLC,JacobedeNaurois2019Detection,Liang2019Prediction,Verwey2000Predicting}} & \centering{}{\tiny{}\citep{Schaap2013Relationship}} &  & {\tiny{}\citep{Liang2010Combining,Yusoff2017Selection}} &  & {\tiny{}\citep{Liang2010Combining,Ranney2008Driver,Strayer2013Measuring}} &  & \centering{}{\tiny{}\citep{Irwin2017Effects,Martin2013AReview,Mets2011Effects,Verster2014Effects}}\tabularnewline
\cline{3-12} \cline{4-12} \cline{5-12} \cline{6-12} \cline{7-12} \cline{8-12} \cline{9-12} \cline{10-12} \cline{11-12} \cline{12-12} 
 &  & \multicolumn{2}{l|}{\textbf{\tiny{}Braking behavior}} &  & \centering{} &  &  & {\tiny{}\citep{Sonnleitner2014EEG}} & {\tiny{}\citep{Harbluk2007AnOnRoad}} &  & \centering{}\tabularnewline
\cline{3-12} \cline{4-12} \cline{5-12} \cline{6-12} \cline{7-12} \cline{8-12} \cline{9-12} \cline{10-12} \cline{11-12} \cline{12-12} 
 &  & \multicolumn{2}{l|}{\textbf{\tiny{}Speed}} & {\tiny{}\citep{Arnedt2000Simulated,JacobedeNaurois2019Detection}} & \centering{} &  & {\tiny{}\citep{Engstrom2007Effects,Yusoff2017Selection}} &  & {\tiny{}\citep{Ranney2008Driver}} & {\tiny{}\citep{Boril2012Towards}} & \centering{}{\tiny{}\citep{Irwin2017Effects,Mets2011Effects}}\tabularnewline
\cline{2-12} \cline{3-12} \cline{4-12} \cline{5-12} \cline{6-12} \cline{7-12} \cline{8-12} \cline{9-12} \cline{10-12} \cline{11-12} \cline{12-12} 
 & \multirow{6}{*}{\begin{turn}{90}
\textbf{\tiny{}Environment}
\end{turn}} & \multicolumn{2}{l|}{\textbf{\tiny{}Road geometry}} &  & \multirow{3}{1cm}{\centering{}{\tiny{}\citep{Palasek2019Attentional}}} &  &  &  &  &  & \centering{}\tabularnewline
\cline{3-5} \cline{4-5} \cline{5-5} \cline{7-12} \cline{8-12} \cline{9-12} \cline{10-12} \cline{11-12} \cline{12-12} 
 &  & \multicolumn{2}{l|}{\textbf{\tiny{}Traffic signs}} &  &  &  &  &  &  &  & \centering{}\tabularnewline
\cline{3-5} \cline{4-5} \cline{5-5} \cline{7-12} \cline{8-12} \cline{9-12} \cline{10-12} \cline{11-12} \cline{12-12} 
 &  & \multicolumn{2}{l|}{\textbf{\tiny{}Road work}} &  &  &  &  &  &  &  & \centering{}\tabularnewline
\cline{3-12} \cline{4-12} \cline{5-12} \cline{6-12} \cline{7-12} \cline{8-12} \cline{9-12} \cline{10-12} \cline{11-12} \cline{12-12} 
 &  & \multicolumn{2}{l|}{\textbf{\tiny{}Traffic density}} &  & \centering{}{\tiny{}\citep{Hao2007Effect,Palasek2019Attentional}} &  &  &  &  &  & \centering{}\tabularnewline
\cline{3-12} \cline{4-12} \cline{5-12} \cline{6-12} \cline{7-12} \cline{8-12} \cline{9-12} \cline{10-12} \cline{11-12} \cline{12-12} 
 &  & \multicolumn{2}{l|}{\textbf{\tiny{}Obstacles}} &  & \multirow{2}{1cm}{\centering{}{\tiny{}\citep{Palasek2019Attentional}}} &  &  &  &  &  & \centering{}\tabularnewline
\cline{3-5} \cline{4-5} \cline{5-5} \cline{7-12} \cline{8-12} \cline{9-12} \cline{10-12} \cline{11-12} \cline{12-12} 
 &  & \multicolumn{2}{l|}{\textbf{\tiny{}Weather}} &  &  &  &  &  &  &  & \centering{}\tabularnewline
\hline 
\end{tabular}}
\end{table}

\begin{table}
\caption{Detailed ``sensors vs indicators'' table, introduced in simplified
form in Figure~\ref{fig:axes-analysis}. Each cell in the heart of
the table gives some references (if any) discussing how the corresponding
sensor is useful for characterizing the corresponding indicator. The
indicators are identical to the ones in Table~\ref{tab-indicators},
thereby allowing one to link both tables.\label{tab-sensors}}

\resizebox{\textwidth}{!}{{\scriptsize{}}%
\begin{tabular}[t]{|l|l|l|l|>{\centering}p{0.8cm}|>{\centering}p{0.8cm}|>{\centering}p{0.6cm}|>{\centering}m{1cm}|>{\centering}p{1cm}|>{\centering}p{0.8cm}|>{\centering}p{1cm}|>{\centering}p{0.8cm}|c|}
\cline{5-13} \cline{6-13} \cline{7-13} \cline{8-13} \cline{9-13} \cline{10-13} \cline{11-13} \cline{12-13} \cline{13-13} 
\multicolumn{1}{l}{} & \multicolumn{1}{l}{} & \multicolumn{1}{l}{} &  & \multicolumn{9}{c|}{\textbf{\tiny{}Sensors}}\tabularnewline
\cline{5-13} \cline{6-13} \cline{7-13} \cline{8-13} \cline{9-13} \cline{10-13} \cline{11-13} \cline{12-13} \cline{13-13} 
\multicolumn{1}{l}{} & \multicolumn{1}{l}{} & \multicolumn{1}{l}{} &  & \multicolumn{6}{c|}{\textbf{\tiny{}Driver}} & \textbf{\tiny{}Vehicle} & \multicolumn{2}{c|}{\textbf{\tiny{}Environment}}\tabularnewline
\cline{5-13} \cline{6-13} \cline{7-13} \cline{8-13} \cline{9-13} \cline{10-13} \cline{11-13} \cline{12-13} \cline{13-13} 
\multicolumn{4}{c|}{} & \multicolumn{1}{>{\centering}p{0.8cm}|}{\textbf{\tiny{}Seat}} & \multicolumn{1}{>{\centering}p{0.8cm}|}{\textbf{\tiny{}Steering wheel}} & \multicolumn{1}{>{\centering}p{0.6cm}|}{\centering{}\textbf{\tiny{}Safety belt}} & \multirow{1}{1cm}{\centering{}\textbf{\tiny{}Internal camera}} & \multirow{1}{1cm}{\centering{}\textbf{\tiny{}Internal microphone}} & \multirow{1}{0.8cm}{\centering{}\textbf{\tiny{}Wearable}} & \textbf{\tiny{}CAN bus} & \textbf{\tiny{}External camera} & \textbf{\tiny{}Radar}\tabularnewline
\hline 
\multirow{24}{*}{\begin{turn}{90}
\textbf{\tiny{}Indicators}
\end{turn}} & \multirow{14}{*}{\begin{turn}{90}
\textbf{\tiny{}Driver}
\end{turn}} & \multirow{6}{*}{\begin{turn}{90}
\textbf{\tiny{}Physiological}
\end{turn}} & \multicolumn{1}{>{\raggedright}m{0.08\textwidth}|}{\textbf{\tiny{}Heart activity}} & {\tiny{}\citep{Kojima2009Noninvasive,Leicht2015Capacitive,Murata2011Noninvasive,Wusk2018NonInvasive}} & {\tiny{}\citep{Silva2012InVehicle}} & \centering{}{\tiny{}\citep{Izumicontact}} & {\tiny{}\citep{Zhang2017Webcam}} &  & {\tiny{}\citep{Gouverneur2017Classification,Ollander2016Comparison,Sevil2017Social,Wu2016APrecise,Wu2016AWearable}} &  &  & \tabularnewline
\cline{4-13} \cline{5-13} \cline{6-13} \cline{7-13} \cline{8-13} \cline{9-13} \cline{10-13} \cline{11-13} \cline{12-13} \cline{13-13} 
 &  &  & \multicolumn{1}{>{\raggedright}m{0.08\textwidth}|}{\textbf{\tiny{}Breathing activity}} &  &  & \centering{} & {\tiny{}\citep{Kiashari2020Evaluation,Murthy2004Touchless}} &  &  &  &  & \tabularnewline
\cline{4-13} \cline{5-13} \cline{6-13} \cline{7-13} \cline{8-13} \cline{9-13} \cline{10-13} \cline{11-13} \cline{12-13} \cline{13-13} 
 &  &  & \multicolumn{1}{>{\raggedright}m{0.08\textwidth}|}{\textbf{\tiny{}Brain activity}} &  &  & \centering{} &  &  &  &  &  & \tabularnewline
\cline{4-13} \cline{5-13} \cline{6-13} \cline{7-13} \cline{8-13} \cline{9-13} \cline{10-13} \cline{11-13} \cline{12-13} \cline{13-13} 
 &  &  & \multicolumn{1}{>{\raggedright}m{0.08\textwidth}|}{\textbf{\tiny{}Electrodermal activity}} &  &  & \centering{} &  &  & {\tiny{}\citep{Gouverneur2017Classification,Ollander2016Comparison,Sevil2017Social}} &  &  & \tabularnewline
\cline{4-13} \cline{5-13} \cline{6-13} \cline{7-13} \cline{8-13} \cline{9-13} \cline{10-13} \cline{11-13} \cline{12-13} \cline{13-13} 
 &  &  & \textbf{\tiny{}Body temperature} &  &  & \centering{} & {\tiny{}\citep{Hermosilla2018Face,Koukiou2017Local,Menon2019Driver}} &  &  &  &  & \tabularnewline
\cline{4-13} \cline{5-13} \cline{6-13} \cline{7-13} \cline{8-13} \cline{9-13} \cline{10-13} \cline{11-13} \cline{12-13} \cline{13-13} 
 &  &  & \multicolumn{1}{>{\raggedright}m{0.08\textwidth}|}{\textbf{\tiny{}Pupil diameter}} &  &  & \centering{} & {\tiny{}\citep{Rosero-Montalvo2021Hybrid,Yokoyama2018Prediction}} &  &  &  &  & \tabularnewline
\cline{3-13} \cline{4-13} \cline{5-13} \cline{6-13} \cline{7-13} \cline{8-13} \cline{9-13} \cline{10-13} \cline{11-13} \cline{12-13} \cline{13-13} 
 &  & \multirow{7}{*}{\begin{turn}{90}
\textbf{\tiny{}Behavioral}
\end{turn}} & \multirow{1}{*}{\textbf{\tiny{}Gaze parameters}} &  &  & \centering{} & {\tiny{}\citep{Bakker2021AMultiStage,Bergasa2006RealTime,Fridman2016Owl,Fridman2018Cognitive,Le2020Evaluating,Mukherjee2015Deep,Musabini2020Heatmap,Naqvi2018DeepLearning,Vicente2015Driver}} &  &  &  &  & \tabularnewline
\cline{4-13} \cline{5-13} \cline{6-13} \cline{7-13} \cline{8-13} \cline{9-13} \cline{10-13} \cline{11-13} \cline{12-13} \cline{13-13} 
 &  &  & \textbf{\tiny{}Blink dynamics} &  &  & \centering{} & {\tiny{}\citep{Baccour2020Camera,Bakker2021AMultiStage,Bergasa2006RealTime,Dreissig2020Driver,Francois2016Tests,Massoz2018MultiTimescale,Teyeb2015Vigilance}} &  &  &  &  & \tabularnewline
\cline{4-13} \cline{5-13} \cline{6-13} \cline{7-13} \cline{8-13} \cline{9-13} \cline{10-13} \cline{11-13} \cline{12-13} \cline{13-13} 
 &  &  & \textbf{\tiny{}PERCLOS} &  &  & \centering{} & {\tiny{}\citep{Bakker2021AMultiStage,Bergasa2006RealTime,Zin2018Vision}} &  &  &  &  & \tabularnewline
\cline{4-13} \cline{5-13} \cline{6-13} \cline{7-13} \cline{8-13} \cline{9-13} \cline{10-13} \cline{11-13} \cline{12-13} \cline{13-13} 
 &  &  & \multicolumn{1}{>{\raggedright}m{0.08\textwidth}|}{\textbf{\tiny{}Facial}{\tiny\par}

\textbf{\tiny{}expressions}} &  &  & \centering{} & {\tiny{}\citep{Bakker2021AMultiStage,Gao2014Detecting,Jeong2018Driver,Melnicuk2017Employing}} &  &  &  &  & \tabularnewline
\cline{4-13} \cline{5-13} \cline{6-13} \cline{7-13} \cline{8-13} \cline{9-13} \cline{10-13} \cline{11-13} \cline{12-13} \cline{13-13} 
 &  &  & \multicolumn{1}{>{\raggedright}m{0.08\textwidth}|}{\textbf{\tiny{}Body posture}} & {\tiny{}\citep{Teyeb2016Towards}} &  & \centering{} & {\tiny{}\citep{Baccour2020Camera,Bakker2021AMultiStage,Bergasa2006RealTime,Dreissig2020Driver,Teyeb2015Vigilance}} &  &  &  &  & \tabularnewline
\cline{4-13} \cline{5-13} \cline{6-13} \cline{7-13} \cline{8-13} \cline{9-13} \cline{10-13} \cline{11-13} \cline{12-13} \cline{13-13} 
 &  &  & \multicolumn{1}{>{\raggedright}m{0.08\textwidth}|}{\textbf{\tiny{}Hands parameters}} &  &  & \centering{} & {\tiny{}\citep{Baheti2018Detection,Le2017Robust,Le2016Robust,Masood2018Detecting,Yan2016Driver}} &  &  &  &  & \tabularnewline
\cline{4-13} \cline{5-13} \cline{6-13} \cline{7-13} \cline{8-13} \cline{9-13} \cline{10-13} \cline{11-13} \cline{12-13} \cline{13-13} 
 &  &  & \textbf{\tiny{}Speech} &  &  & \centering{} &  & {\tiny{}\citep{Gavrilescu2019Feedforward,Basu2017AReview,Boril2012Towards,Zhang2018Speech}} &  &  &  & \tabularnewline
\cline{3-13} \cline{4-13} \cline{5-13} \cline{6-13} \cline{7-13} \cline{8-13} \cline{9-13} \cline{10-13} \cline{11-13} \cline{12-13} \cline{13-13} 
 &  & \multicolumn{2}{l|}{\textbf{\tiny{}Subjective}} &  &  & \centering{} &  &  &  &  &  & \tabularnewline
\cline{2-13} \cline{3-13} \cline{4-13} \cline{5-13} \cline{6-13} \cline{7-13} \cline{8-13} \cline{9-13} \cline{10-13} \cline{11-13} \cline{12-13} \cline{13-13} 
 & \multirow{4}{*}{\begin{turn}{90}
\textbf{\tiny{}Vehicle}
\end{turn}} & \multicolumn{2}{l|}{\textbf{\tiny{}Wheel steering}} &  &  & \centering{} &  &  &  & {\tiny{}\citep{Berri2018ANonintrusive,Fridman2019MIT,Harkous2018Application,Harkous2019ATwoStage,Li2008Design,Li2015Drunk,Shirazi2014Detection}} &  & \tabularnewline
\cline{3-13} \cline{4-13} \cline{5-13} \cline{6-13} \cline{7-13} \cline{8-13} \cline{9-13} \cline{10-13} \cline{11-13} \cline{12-13} \cline{13-13} 
 &  & \multicolumn{2}{l|}{\textbf{\tiny{}Lane discipline}} &  &  & \centering{} &  &  &  & {\tiny{}\citep{Berri2018ANonintrusive,Harkous2019ATwoStage,Harkous2018Application,Shirazi2014Detection}} & {\tiny{}\citep{Apostoloff2003Robust,Bakker2021AMultiStage}} & \tabularnewline
\cline{3-13} \cline{4-13} \cline{5-13} \cline{6-13} \cline{7-13} \cline{8-13} \cline{9-13} \cline{10-13} \cline{11-13} \cline{12-13} \cline{13-13} 
 &  & \multicolumn{2}{l|}{\textbf{\tiny{}Braking behavior}} &  &  & \centering{} &  &  &  & {\tiny{}\citep{Berri2018ANonintrusive,Fridman2019MIT,Li2008Design}} &  & \tabularnewline
\cline{3-13} \cline{4-13} \cline{5-13} \cline{6-13} \cline{7-13} \cline{8-13} \cline{9-13} \cline{10-13} \cline{11-13} \cline{12-13} \cline{13-13} 
 &  & \multicolumn{2}{l|}{\textbf{\tiny{}Speed}} &  &  & \centering{} &  &  &  & {\tiny{}\citep{Berri2018ANonintrusive,Fridman2019MIT,Harkous2019ATwoStage,Harkous2018Application,Li2008Design}} & \centering{} & \tabularnewline
\cline{2-13} \cline{3-13} \cline{4-13} \cline{5-13} \cline{6-13} \cline{7-13} \cline{8-13} \cline{9-13} \cline{10-13} \cline{11-13} \cline{12-13} \cline{13-13} 
 & \multirow{6}{*}{\begin{turn}{90}
\textbf{\tiny{}Environment}
\end{turn}} & \multicolumn{2}{l|}{\textbf{\tiny{}Road geometry}} &  &  & \centering{} &  &  &  &  & \multirow{6}{0.8cm}{\centering{}{\tiny{}\citep{Palasek2019Attentional}}} & \tabularnewline
\cline{3-11} \cline{4-11} \cline{5-11} \cline{6-11} \cline{7-11} \cline{8-11} \cline{9-11} \cline{10-11} \cline{11-11} \cline{13-13} 
 &  & \multicolumn{2}{l|}{\textbf{\tiny{}Traffic signs}} &  &  & \centering{} &  &  &  &  &  & \tabularnewline
\cline{3-11} \cline{4-11} \cline{5-11} \cline{6-11} \cline{7-11} \cline{8-11} \cline{9-11} \cline{10-11} \cline{11-11} \cline{13-13} 
 &  & \multicolumn{2}{l|}{\textbf{\tiny{}Road work}} &  &  & \centering{} &  &  &  &  &  & \tabularnewline
\cline{3-11} \cline{4-11} \cline{5-11} \cline{6-11} \cline{7-11} \cline{8-11} \cline{9-11} \cline{10-11} \cline{11-11} \cline{13-13} 
 &  & \multicolumn{2}{l|}{\textbf{\tiny{}Traffic density}} &  &  & \centering{} &  &  &  &  &  & \tabularnewline
\cline{3-11} \cline{4-11} \cline{5-11} \cline{6-11} \cline{7-11} \cline{8-11} \cline{9-11} \cline{10-11} \cline{11-11} \cline{13-13} 
 &  & \multicolumn{2}{l|}{\textbf{\tiny{}Obstacles}} &  &  & \centering{} &  &  &  &  &  & {\tiny{}\citep{Saponara2019Radar}}\tabularnewline
\cline{3-11} \cline{4-11} \cline{5-11} \cline{6-11} \cline{7-11} \cline{8-11} \cline{9-11} \cline{10-11} \cline{11-11} \cline{13-13} 
 &  & \multicolumn{2}{l|}{\textbf{\tiny{}Weather}} &  &  & \centering{} &  &  &  &  &  & \tabularnewline
\hline 
\end{tabular}}
\end{table}

\section{Summary and conclusion\label{sec:conclusion}}

This paper focuses on the characterization of the state of a driver,
which is the first key step for driver monitoring (DM) and driver
monitoring systems (DMSs). It surveys (in Section~\ref{sec:bibliographic-study})
the relevant scientific and technical literature on driver-state characterization,
and subsequently provides a synthesis (in Sections~\ref{sec:framework-DMS}-\ref{sec:under-the-influence})
of the main, published techniques for this characterization.

The survey yielded $\numberOfSurveys$ publications in scientific/technical
journals and conference proceedings. Their examination led to the
conclusion that the state of a driver should be characterized according
to five main dimensions---called here “(sub)\constructs”---of drowsiness,
mental workload, distraction (further subdivided into four types qualified
of manual, visual, auditory, and cognitive), emotions, and under the
influence.

In comparison with standard physical quantities, such as voltage and
power, these \constructs are not well defined and/or are very difficult---if
at all possible---to quantify or to label, not only in a validated
way, but also in real time and non-invasively, as is required in the
driving context. The only reasonable approach, found almost universally
in the literature, is to have recourse to indicators (of each of these
\constructs), the value of which can be obtained in a practical and
validated way. Examples of indicators are the eye-blink rate, the
standard deviation of lane departure (SDLP), and the outside temperature.
The values of many indicators (but not all) are obtained by applying
algorithms, often complex, to data (typically signals and images)
collected from sensors.

The last paragraph brings to light the three ingredients that, in
our view, lie at the heart of DM and DMSs, \ie, the triad of \constructs,
indicators (of these \constructs), and sensors (providing data, which
are the source of the values of these indicators). Figure~\ref{fig-BD-DMS}
links these three ingredients.

Our survey confirmed the intuition that one should monitor, not only
the driver (D), but also the (driven) vehicle (V) and the (driving)
environment (E). Accordingly, we partitioned both the indicators and
the sensors into D, V, and E categories, leading to the phrases “X-based
indicators” and “X-centric sensors'', where X can be D, V, or E.
For the D-based indicators, we further distinguished between three
types: physiological, behavioral, and subjective. The three examples
of indicators given earlier correspond to D, V, and E, respectively.

The major outcome of the paper is the pair of interlocked tables “\constructs
vs indicators” (Table~\ref{tab-indicators}) and “sensors vs indicators”
(Table~\ref{tab-sensors}), where each cell contains zero, one, or
more references. These tables bring together, in an organized way,
most of the useful information found in the literature, up to the
time of this writing, about driver-state characterization, for DM
and DMSs. These tables constitute an up-to-date, at-a-glance, visual
reference guide for anyone active in this field. They provide immediate
answers to key questions that arise in the design of DMSs, such as
the four questions posed in Section~\ref{sec:Goal-and-approach}.

The pair of tables and the references they contain lead to the following
main conclusions:
\begin{enumerate}
\item Each \construct can be inferred from several indicators (which are
often far from perfect), thereby encouraging multimodal fusion.
\item The internal camera (possibly with several instances) appears to be
the most-commonly-used sensor.
\item Wearable sensors (\eg, smartwatches) are increasingly used to obtain
driver-based, physiological indicators and vehicle-based indicators.
\item Environment-based indicators are often ignored, even though there
is an agreement that they should be used.
\item Driver-based, subjective indicators, although sometimes alluded to,
cannot be used in real driving, but are essential for the validation
of some indicators of some \constructs.
\item Brain activity is a recognized indicator of several \constructs,
but cannot be accessed today in a non-invasive, reliable, and inexpensive
way in real driving.
\item Several methods for characterizing each of the 5 states use, without
surprise, techniques of \nomenclature{ML}{machine learning}machine
learning (ML) and, especially, of deep learning.
\item The term ``predict(ion)'' often refers to a present state rather
than to a future state, and few papers describe techniques ``to tell
beforehand'', for example, the future values of indicators and levels
of \constructs.
\end{enumerate}
The next two paragraphs respectively elaborate on the last two points.

For driving safety, it is paramount that the processing and decisions
made by any algorithm used in a vehicle, including for DM, be fully
explainable (to a human being) at the time of design and certification
of this algorithm. Most algorithms using ML do not, however, have
this necessary feature of explainability, or interpretability, and
this is certainly the case for ML-based algorithms that would learn
on-the-fly during one or more trips. Therefore, while ML algorithms
and, especially deep-learning algorithms, often provide stellar performances
on specific datasets in comparison with other types of algorithms,
they will almost certainly not be acceptable to an equipment provider
or a vehicle manufacturer. There is, however, a trend toward designing
ML algorithms that produce results that can be explained~\citep{Linardatos2021Explainable,Zablocki2021Explainability}.
The above remarks apply not only to ML but also to any approach whose
operation cannot be explained simply. Our framework, which implies
the use of indicators and states, supports the desired explainability.
It indeed prevents any algorithm from going, in one fell swoop, from
(nearly-)raw sensor data to driver characterization, by forcing it
to estimate both the values of indicators and the levels of states,
as a stepping stone toward the ultimate characterization of the state
of a driver.

The literature on DM focuses almost exclusively on characterizing
the “present” state of the driver. We use quotes because the characterization
is typically based on data from the recent past, for example, in a
window that extends over several tens of seconds and butts against
almost the present time. This results in a characterization of the
“recent-past” state of the driver. If the driver is in control, a
DMS using this characterization may not have sufficient lead time
to take proper emergency action (to issue an alarm and/or to take
back the control) and, if the vehicle is in control, such a DMS may
hand the control over to the driver even though \he might be falling
asleep or getting distracted in a few tens of seconds or more. A major
missing link in current DMS research and development is thus the true
prediction of the future state of the driver, at least a few tens
of seconds into the future.

On the one hand, Tables~\ref{tab-indicators} and \ref{tab-sensors}
show, at a glance, which areas of driver-state characterization have
been the object of research and with what intensity (as measured by
the number of references listed in each cell). For example, Table~\ref{tab-indicators}
shows that significant research has been performed to analyze the
emotions of the driver using the driver-based, physiological indicators
of heart rate, breathing activity, and electrodermal activity. On
the other hand, the two tables show, also at a glance, where little
or no research has been performed to date, thereby suggesting new,
potentially-fruitful research areas. The two tables should thus prove
to be a rich source of information for both research and product development.

Starting from a set of $\numberOfSurveys$ initial references, our
exploration of the field of DM led us to examine a total of $\numberOfArticles$
references. While our criss-crossing of the field, at several different
times, led us to identify many relevant publications, our search cannot,
obviously, be exhaustive. In any case, the two histograms of “number
of references vs year” of Figure~\ref{fig:Histograms} (for the $\numberOfSurveys$
and $\numberOfArticles$ references, respectively) constitute a clue
that the research activity in DM has been accelerating over the past
decade.

The methodology used in this paper can be applied to update the tables
at various times in the future to take into accounts new developments.
This can be done by adding and/or removing rows, columns, and/or references,
as appropriate.

Characterizing the state of a driver and, more generally, DM will
remain important despite the progressive increase in vehicle automation.
SAE Level 3 enables vehicles to drive by themselves under certain
conditions such as on a highway and in sunny weather, but a driver
must still be present and able to take back the control of the vehicle
at any time and in a relatively short lapse of time. In order to ensure
that the driver is able to take back the control, technologies for
monitoring the state of the driver will become even more critical.
These technologies are also needed to monitor the driver during the
time \he is driving, and to possibly allow the vehicle to take back
the control if necessary.

Currently, some vehicle manufacturers offer DMSs based on the behavior
of the driver and/or the behavior of the vehicle, such as the detection
of steering-wheel movements and lane deviations, respectively. These
systems can be useful in current vehicles with automation up to (SAE)
Level $2$, but will become obsolete at higher levels of automation.
Indeed, when a vehicle drives autonomously, monitoring its behavior
does not give any information about the state of the driver, and technologies
that directly monitor both the driver and the driving environment
are a necessity as long as the driver is involved in the driving task,
at least partially.

To date, the development of driving-automation systems (DASs) has
moved at a faster pace than has the development of DMSs. This is,
in major part, a consequence of the long-held belief by some automotive-industry
players that they would be able to easily leapfrog Levels 3 and 4,
and move on directly to Level 5, where there is no need to monitor
the driver. But, most experts now agree that it will be decades before
most privately-owned vehicles are fully automated, if ever. Along
the long and winding road to Level 5, the automotive industry will
need to significantly boost the research on, and the development of,
DMSs. For Levels 3 and 4, the same industry will need to develop automated-driving
systems (ADSs) and DMSs in full synergy. The future could thus not
be brighter for the field of DM and DMSs.

\subsubsection*{Acknowledgments}

This work was supported in part by the European Regional Development
Fund (ERDF).

\subsubsection*{Conflict of interest}

The authors declare no conflict of interest.

{\footnotesize{}}{\footnotesize\par}

{\footnotesize{}
}{\footnotesize\par}

\ifthenelse{\boolean{FORTHEWEB} \OR \boolean{FORARXIV}}{

\section*{Short biography of authors}

\begin{wrapfigure}{l}{25mm}
	\includegraphics[width=1in,clip,keepaspectratio]{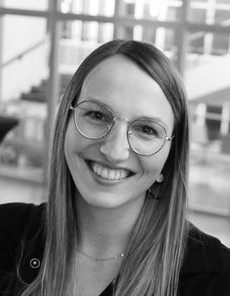}
\end{wrapfigure}\par
\textbf{Ana\"is Halin} received the M.S. degree in Electrical Engineering from the University of Li\`ege (ULi\`ege), Belgium, in 2017. Since then, she has been a Research and Teaching Assistant in the Department of Electrical Engineering and Computer Science of ULi\`ege’s School of Applied Sciences. She is currently a Ph.D. candidate in the same EECS department. Her current research interests include the real-time characterization of the state of alertness of car drivers. Since 2018, she has been a member of the Technical Program Committee of the International Conference on 3D Immersion (IC3D), which is technically co-sponsored by the IEEE Signal Processing Society, and she has played a key role in reviewing papers, preparing the proceedings, chairing sessions, and organizing the logistics. Since 2017, she has also played a major role in the logistics of the many Thematic Conferences of Stereopsia, the World Immersion Forum. In 2020, she also received the M.S. degree in Management in part-time study from HEC Liège.\par

\vspace{1em}

\begin{wrapfigure}{l}{25mm}
	\includegraphics[width=1in,clip,keepaspectratio]{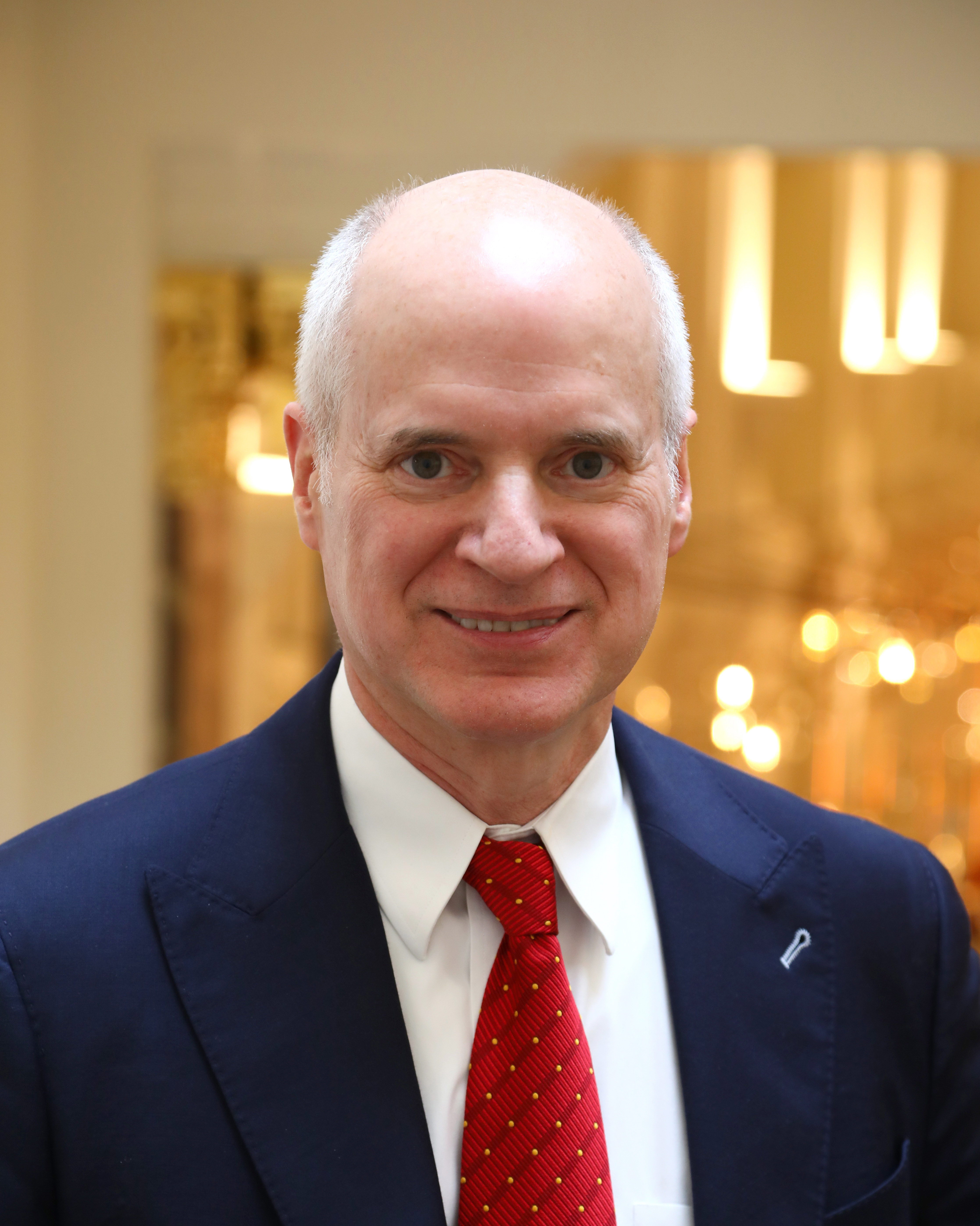}
\end{wrapfigure}\par
\noindent \textbf{Jacques G. Verly} received the Ing\'enieur \'Electronicien degree from University of Li\`ege, Belgium. Sponsored by the Belgian American Educational Foundation (BAEF), he attended Stanford University, where he received the M.S. and Ph.D. degrees in electrical engineering. From 1980 to 2000, he was at MIT Lincoln Laboratory, doing research in many areas, including signal and image processing for several imaging sensors (visible, IR, laser radar, SAR). From 2000 to 2017, he was a full professor in the Department of EECS of the University of Li\`ege. He is currently an honorary full professor. His current research areas include immersive technologies, drowsiness and cognitive state monitoring, and terahertz imaging. He is the instigator of Stereopsia (Europe and LATAM), SomnoSafe, and International Conference on 3D Immersion (IC3D). He is an architect and partner of the European Horizon-2020 XR4ALL project. He is the instigator and co-founder of the Phasya company, specialized in driver cognitive state monitoring.\par

\vspace{1em}

\begin{wrapfigure}{l}{25mm}
	\includegraphics[width=1in,clip,keepaspectratio]{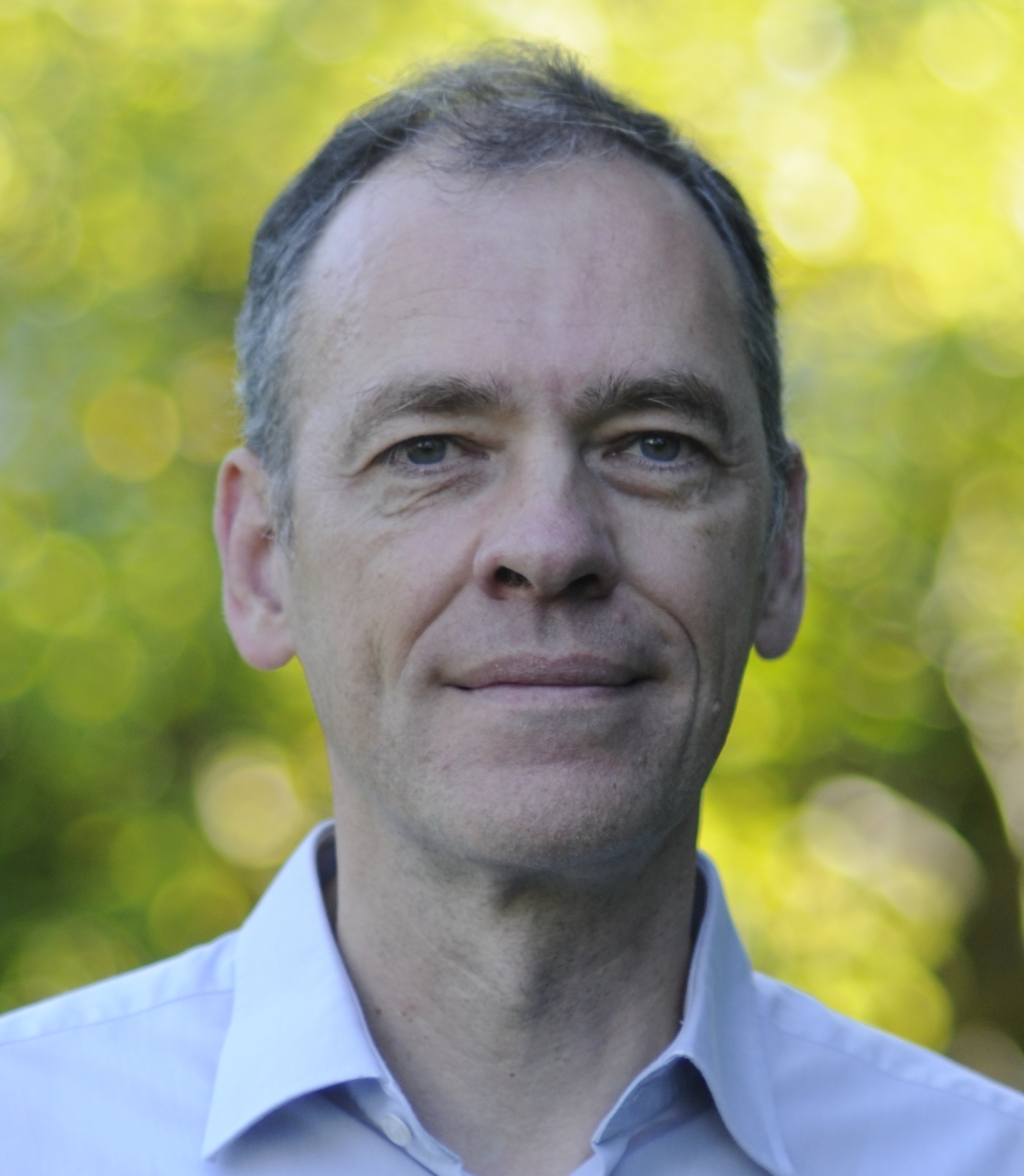}
\end{wrapfigure}\par
\noindent \textbf{Marc Van Droogenbroeck} received the degree in Electrical Engineering and the Ph.D. degree from the University of Louvain (UCLouvain), Belgium, in 1990 and 1994, respectively. While working toward the Ph.D, he spent two years with the Center of Mathematical Morphology (CMM), School of Mines of Paris, Fontainebleau, France. In April 1994, he joined the New Development Department of Belgacom. He was the Head of the Belgian Delegation within the ISO/MPEG Committee and served as a Representative to the World Wide Web Consortium for two years. In 2003, he was a Visiting Scientist at CSIRO, Sydney, Australia. Since 1998, he has been a Member of the Faculty of Applied Sciences at the University of Li\`ege, Belgium, where he is a full Professor. His current interests include computer vision, deep learning, machine learning, real-time processing, motion analysis, soccer analysis, driver monitoring, and exoplanet detection.\par

}{}

\newpage{}

\appendix

\section{List of acronyms~\label{list-of-acronyms}}

\settowidth{\nomlabelwidth}{NASA TLX}
\printnomenclature{}\newpage{}

\section{Printable version of Table~\ref{tab-reviews-summary-sumvers}\label{appendix:table-for-printing}}

Tables~\ref{tab-reviews-summary-printvers-1} and~\ref{tab-reviews-summary-printvers-2}
constitute a version of Table~\ref{tab-reviews-summary-sumvers}
suitable for printing.
\begin{table}[H]
\caption{This table gives the three main columns of Table~\ref{tab-reviews-summary-sumvers}
labelled ``\Constructs'', ``Sensors'', and ``Tests''. The remaining
main column ``Indicators'' is provided in Table~\ref{tab-reviews-summary-printvers-2}.
This partitioning of Table~\ref{tab-reviews-summary-sumvers} allows
for more comfortable visualization of its content when printed. \label{tab-reviews-summary-printvers-1}}

\centering{}\resizebox{\textwidth}{!}{%
\begin{tabular}{|>{\raggedright}p{4cm}|c|>{\centering}m{1.5cm}|c|c|>{\centering}p{1.5cm}|c|c|c|>{\centering}p{1.5cm}|}
\hline 
\multirow{3}{4cm}{\textbf{References}} & \multicolumn{5}{c|}{\textbf{\Constructs}} & \multicolumn{3}{c|}{\textbf{Sensors}} & \multirow{3}{1.5cm}{\centering{}\textbf{\small{}Tests}}\tabularnewline
\cline{2-9} \cline{3-9} \cline{4-9} \cline{5-9} \cline{6-9} \cline{7-9} \cline{8-9} \cline{9-9} 
 & \multirow{2}{*}{\textbf{\small{}Drowsiness}} & \multirow{2}{1.5cm}{\centering{}\textbf{\small{}Mental workload}} & \multirow{2}{*}{\textbf{Distraction}} & \multirow{2}{*}{\textbf{Emotions}} & \multirow{2}{1.5cm}{\centering{}\textbf{\small{}Under the influence}} & \multirow{2}{*}{\textbf{Driver}} & \multirow{2}{*}{\textbf{\small{}Vehicle}} & \multirow{2}{*}{\textbf{\small{}Environment}} & \tabularnewline
 &  &  &  &  &  &  &  &  & \tabularnewline
\hline 
\citet{Ahir2019Driver} & $\check$ &  &  &  &  & $\sensors{\cameraDriver,\electrodesDriver}$ &  & $\sensors{\cameraEnvironment}$ & $\testConditions{\realConditions,\simulatorConditions}$\tabularnewline
\hline 
\citet{Alluhaibi2018Driver} & $\check$ &  & $\check$ & $\driverStateConstruct{\angerEmotion}$ &  & $\sensors{\cameraMobileDriver,\microphoneMobileDriver}$ & $\indirectVehicleSensor$ &  & \tabularnewline
\hline 
\citet{Arun2012Driver} &  &  & $\driverStateConstruct{\visualDriverState,\cognitiveDriverState}$ &  &  & $\sensors{\cameraDriver,\wearableDriver,\eyeTrackerDriver}$ & $\check$ &  & $\testConditions{\simulatorConditions}$\tabularnewline
\hline 
\citet{Balandong2018AReview} & $\check$ &  &  &  &  & $\sensors{\electrodesDriver}$ &  &  & $\testConditions{\simulatorConditions}$\tabularnewline
\hline 
\multirow{1}{4cm}{\citet{Begum2013Intelligent}} & $\check$ &  & $\check$ & $\driverStateConstruct{\stressEmotion}$ &  & $\sensors{\seatDriver,\steeringWheelDriver,\safetyBeltDriver,\wearableDriver}$ &  &  & $\testConditions{\realConditions,\simulatorConditions}$\tabularnewline
\hline 
\citet{Chacon-murguia2015Detecting} & $\check$ &  &  &  &  & $\sensors{\steeringWheelDriver,\cameraDriver}$ &  & $\sensors{\radarEnvironment}$ & $\testConditions{\realConditions}$\tabularnewline
\hline 
\multirow{1}{4cm}{\citet{Chan2019AComprehensive}} & $\check$ &  &  &  &  & $\sensors{\cameraMobileDriver,\microphoneMobileDriver}$ &  &  & $\testConditions{\realConditions}$\tabularnewline
\hline 
\multirow{1}{4cm}{\citet{Chhabra2017ASurvey}} & $\check$ &  & $\check$ &  & $\driverStateConstruct{\alcoholUnderInfluence}$ & $\sensors{\seatDriver,\cameraMobileDriver,\microphoneMobileDriver}$ & $\indirectVehicleSensor$ &  & $\testConditions{\realConditions,\simulatorConditions}$\tabularnewline
\hline 
\citet{Chowdhury2018Sensor} & $\check$ &  &  &  &  &  &  &  & $\testConditions{\simulatorConditions}$\tabularnewline
\hline 
\citet{Chung2019Methods} &  &  &  & $\driverStateConstruct{\stressEmotion}$ &  & $\sensors{\cameraDriver,\wearableDriver}$ & $\check$ &  & $\testConditions{\realConditions,\simulatorConditions}$\tabularnewline
\hline 
\citet{Coetzer2009Driver} & $\check$ &  &  &  &  & $\sensors{\cameraDriver}$ & $\check$ &  & $\testConditions{\realConditions,\simulatorConditions}$\tabularnewline
\hline 
\citet{Dababneh2015Driver} & $\check$ &  &  &  &  & $\sensors{\cameraDriver,\wearableDriver}$ &  & $\sensors{\radarEnvironment}$ & $\testConditions{\realConditions,\simulatorConditions}$\tabularnewline
\hline 
\citet{Dahiphale2015AReview} & $\check$ &  & $\check$ &  &  & $\sensors{\cameraDriver}$ &  &  & $\testConditions{\realConditions}$\tabularnewline
\hline 
\multirow{1}{4cm}{\citet{Dong2011Driver}} & $\check$ &  & $\check$ &  &  & $\sensors{\cameraDriver}$ & $\check$ &  & $\testConditions{\realConditions}$\tabularnewline
\hline 
\multirow{1}{4cm}{\citet{ElKhatib2020Driver}} & $\check$ &  & $\driverStateConstruct{\manualDriverState,\visualDriverState,\cognitiveDriverState}$ &  &  & $\sensors{\cameraDriver}$ & $\indirectVehicleSensor$ & $\sensors{\cameraEnvironment,\radarEnvironment}$ & $\testConditions{\realConditions,\simulatorConditions}$\tabularnewline
\hline 
\multirow{1}{4cm}{\citet{Ghandour2020Driver}} &  &  & $\driverStateConstruct{\manualDriverState,\visualDriverState,\auditoryDriverState,\cognitiveDriverState}$ & $\driverStateConstruct{\stressEmotion}$ &  & $\sensors{\cameraDriver,\wearableDriver}$ &  &  & $\testConditions{\realConditions,\simulatorConditions}$\tabularnewline
\hline 
\multirow{1}{4cm}{\citet{Hecht2018AReview}} & $\check$ & $\check$ & $\check$ &  &  & $\sensors{\electrodesDriver,\eyeTrackerDriver}$ &  &  & $\testConditions{\realConditions,\simulatorConditions}$\tabularnewline
\hline 
\multirow{1}{4cm}{\citet{Kang2013Various}} & $\check$ &  & $\check$ &  &  & $\sensors{\seatDriver,\steeringWheelDriver,\cameraDriver}$ & $\check$ &  & $\testConditions{\realConditions,\simulatorConditions}$\tabularnewline
\hline 
\multirow{1}{4cm}{\citet{Kaplan2015Driver}} & $\check$ &  & $\check$ &  &  & $\sensors{\steeringWheelDriver,\cameraMobileDriver,\microphoneMobileDriver,\wearableDriver}$ & $\check$ &  & $\testConditions{\realConditions,\simulatorConditions}$\tabularnewline
\hline 
\citet{Kaye2018Comparison} & $\check$ &  &  & $\driverStateConstruct{\stressEmotion}$ &  &  &  &  & $\testConditions{\realConditions,\simulatorConditions}$\tabularnewline
\hline 
\citet{Khan2019AComprehensive} & $\check$ &  & $\driverStateConstruct{\manualDriverState,\visualDriverState,\auditoryDriverState,\cognitiveDriverState}$ &  &  & $\sensors{\wearableDriver}$ &  &  & $\testConditions{\realConditions}$\tabularnewline
\hline 
\citet{Kumari2017ASurvey} & $\check$ &  &  &  &  & $\sensors{\cameraDriver}$ &  &  & \tabularnewline
\hline 
\citet{Lal2001ACritical} & $\check$ &  &  &  &  & $\sensors{\cameraDriver}$ &  &  & $\testConditions{\simulatorConditions}$\tabularnewline
\hline 
\citet{Laouz2020Literature} & $\check$ &  &  &  &  & $\sensors{\seatDriver,\cameraDriver,\wearableDriver}$ &  & $\sensors{\cameraEnvironment}$ & $\testConditions{\realConditions}$\tabularnewline
\hline 
\citet{Leonhardt2018Unobtrusive} &  &  &  &  &  & $\sensors{\seatDriver,\steeringWheelDriver,\safetyBeltDriver,\cameraDriver}$ &  &  & $\testConditions{\realConditions}$\tabularnewline
\hline 
\citet{Liu2019AReview} & $\check$ &  &  &  &  & $\sensors{\cameraDriver}$ & $\check$ &  & $\testConditions{\realConditions}$\tabularnewline
\hline 
\citet{Marquat2015Review} &  & $\check$ &  &  &  & $\sensors{\eyeTrackerDriver}$ &  &  & $\testConditions{\realConditions,\simulatorConditions}$\tabularnewline
\hline 
\citet{Martinez2017Driving} &  &  &  & $\driverStateConstruct{\angerEmotion}$ &  &  & $\indirectVehicleSensor$ &  & \tabularnewline
\hline 
\citet{Mashko2015Review} & $\check$ &  &  &  &  & $\sensors{\cameraDriver,\wearableDriver}$ & $\check$ & $\sensors{\cameraEnvironment,\radarEnvironment}$ & $\testConditions{\realConditions,\simulatorConditions}$\tabularnewline
\hline 
\citet{Mashru2018Detection} & $\check$ &  &  &  &  & $\sensors{\seatDriver,\steeringWheelDriver,\cameraDriver,\wearableDriver}$ &  &  & $\testConditions{\simulatorConditions}$\tabularnewline
\hline 
\citet{Melnicuk2016Towards} & $\check$ & $\check$ & $\driverStateConstruct{\cognitiveDriverState}$ & $\driverStateConstruct{\stressEmotion,\angerEmotion}$ &  & $\sensors{\seatDriver,\steeringWheelDriver,\safetyBeltDriver,\cameraMobileDriver,\wearableDriver}$ & $\indirectVehicleSensor$ &  & $\testConditions{\realConditions}$\tabularnewline
\hline 
\citet{Mittal2016Head} & $\check$ &  &  &  &  & $\sensors{\cameraDriver,\electrodesDriver}$ & $\check$ & $\sensors{\cameraEnvironment}$ & $\testConditions{\realConditions}$\tabularnewline
\hline 
\citet{Murugan2019Analysis} & $\check$ &  &  &  &  & $\sensors{\cameraDriver,\electrodesDriver}$ & $\check$ &  & $\testConditions{\simulatorConditions}$\tabularnewline
\hline 
\citet{Nair2016ASurvey} & $\check$ &  & $\check$ &  & $\driverStateConstruct{\alcoholUnderInfluence}$ & $\sensors{\seatDriver,\cameraMobileDriver}$ & $\check$ & $\sensors{\radarEnvironment}$ & \tabularnewline
\hline 
\citet{Nemcova2021Multimodal} & $\check$ &  &  & $\driverStateConstruct{\stressEmotion}$ &  & $\sensors{\seatDriver,\steeringWheelDriver,\cameraDriver,\wearableDriver,\eyeTrackerDriver}$ & $\check$ &  & $\testConditions{\realConditions,\simulatorConditions}$\tabularnewline
\hline 
\citet{Ngxande2017Driver} & $\check$ &  &  &  &  & $\sensors{\cameraDriver}$ &  &  & \tabularnewline
\hline 
\citet{Oviedo-Trespalacios2016Understanding} &  & $\check$ & $\check$ &  &  &  &  &  & $\testConditions{\realConditions,\simulatorConditions}$\tabularnewline
\hline 
\citet{Papantoniou2017Review} &  & $\check$ & $\check$ &  &  & $\sensors{\cameraDriver}$ &  & $\sensors{\cameraEnvironment,\radarEnvironment}$ & $\testConditions{\realConditions,\simulatorConditions}$\tabularnewline
\hline 
\citet{Pratama2017AReview} & $\check$ &  &  &  &  & $\sensors{\cameraDriver,\wearableDriver,\electrodesDriver}$ &  & $\sensors{\cameraEnvironment}$ & $\testConditions{\realConditions,\simulatorConditions}$\tabularnewline
\hline 
\citet{Ramzan2019ASurvey} & $\check$ &  &  &  &  & $\sensors{\cameraDriver,\wearableDriver,\electrodesDriver}$ & $\check$ &  & $\testConditions{\realConditions,\simulatorConditions}$\tabularnewline
\hline 
\citet{Sahayadhas2012Detecting} & $\check$ &  &  &  &  & $\sensors{\seatDriver,\steeringWheelDriver,\cameraDriver,\wearableDriver}$ & $\check$ &  & $\testConditions{\realConditions,\simulatorConditions}$\tabularnewline
\hline 
\citet{Scott-Parker2017Emotions} &  &  &  & $\driverStateConstruct{\stressEmotion,\angerEmotion}$ &  & $\sensors{\eyeTrackerDriver}$ &  & $\sensors{\cameraEnvironment}$ & $\testConditions{\realConditions,\simulatorConditions}$\tabularnewline
\hline 
\citet{Seth2020ASurvey} & $\check$ &  &  &  &  & $\sensors{\cameraDriver}$ & $\check$ &  & $\testConditions{\realConditions}$\tabularnewline
\hline 
\citet{Shameen2018Electroencephalography} & $\check$ &  &  &  &  & $\sensors{\electrodesDriver}$ &  &  & $\testConditions{\simulatorConditions}$\tabularnewline
\hline 
\citet{Sigari2014AReview} & $\check$ &  &  &  &  & $\sensors{\cameraDriver}$ &  &  & $\testConditions{\realConditions}$\tabularnewline
\hline 
\citet{Sikander2019Driver} & $\check$ &  &  &  &  & $\sensors{\seatDriver,\steeringWheelDriver,\safetyBeltDriver,\cameraDriver,\wearableDriver,\electrodesDriver}$ &  &  & $\testConditions{\realConditions}$\tabularnewline
\hline 
\citet{Singh2021Analyzing} & $\check$ & $\check$ & $\check$ & $\check$ &  & $\sensors{\cameraDriver,\wearableDriver}$ & $\check$ & $\sensors{\cameraEnvironment,\radarEnvironment}$ & $\testConditions{\realConditions}$\tabularnewline
\hline 
\citet{Subbaiah2019Driver} & $\check$ &  &  &  &  & $\sensors{\cameraDriver}$ &  &  & $\testConditions{\realConditions,\simulatorConditions}$\tabularnewline
\hline 
\citet{Tu2016ASurvey} & $\check$ &  &  &  &  & $\sensors{\cameraMobileDriver,\wearableDriver,\electrodesDriver}$ & $\check$ &  & $\testConditions{\realConditions,\simulatorConditions}$\tabularnewline
\hline 
\citet{Ukwuoma2019Deep} & $\check$ &  &  &  &  & $\sensors{\cameraDriver,\wearableDriver,\electrodesDriver}$ &  &  & $\testConditions{\realConditions}$\tabularnewline
\hline 
\citet{Vilaca2017Systematic} & $\check$ &  & $\check$ &  &  & $\sensors{\cameraDriver,\microphoneDriver}$ & $\check$ & $\sensors{\cameraEnvironment}$ & \tabularnewline
\hline 
\citet{Vismaya2020AReview} &  &  & $\check$ &  &  & $\sensors{\cameraDriver,\eyeTrackerDriver}$ &  &  & $\testConditions{\realConditions,\simulatorConditions}$\tabularnewline
\hline 
\citet{Wang2006Driver} & $\check$ &  &  &  &  & $\sensors{\cameraDriver,\wearableDriver}$ &  &  & $\testConditions{\realConditions,\simulatorConditions}$\tabularnewline
\hline 
\citet{Welch2019AReview} &  &  &  & $\driverStateConstruct{\stressEmotion,\angerEmotion}$ &  & $\sensors{\seatDriver,\steeringWheelDriver,\cameraDriver,\wearableDriver}$ & $\check$ &  & $\testConditions{\realConditions,\simulatorConditions}$\tabularnewline
\hline 
\citet{Yusoff2017Selection} &  &  & $\driverStateConstruct{\visualDriverState,\cognitiveDriverState}$ &  &  & $\sensors{\eyeTrackerDriver}$ &  &  & \tabularnewline
\hline 
\citet{Zhang2013Review} & $\check$ &  &  &  &  & $\sensors{\cameraDriver}$ &  & $\sensors{\cameraEnvironment}$ & $\testConditions{\realConditions,\simulatorConditions}$\tabularnewline
\hline 
\end{tabular}}
\end{table}

\begin{table}
\caption{This table gives the main column of Table~\ref{tab-reviews-summary-sumvers}
labelled ``Indicators''. The three remaining main columns ``\Constructs'',
``Sensors'', and ``Tests'' are provided in Table~\ref{tab-reviews-summary-printvers-1}.
This partitioning of Table~\ref{tab-reviews-summary-sumvers} allows
for more comfortable visualization of its content when printed. \label{tab-reviews-summary-printvers-2}}

\centering{}\resizebox{\textwidth}{!}{%
\begin{tabular}{|>{\raggedright}p{4cm}|c|c|c|c|c|}
\hline 
\multirow{3}{4cm}{\textbf{References}} & \multicolumn{5}{c|}{\textbf{Indicators}}\tabularnewline
\cline{2-6} \cline{3-6} \cline{4-6} \cline{5-6} \cline{6-6} 
 & \multicolumn{3}{c|}{\textbf{Driver}} & \multirow{2}{*}{\textbf{Vehicle}} & \multirow{2}{*}{\textbf{Environment}}\tabularnewline
\cline{2-4} \cline{3-4} \cline{4-4} 
 & \textbf{Physiological} & \textbf{Behavioral} & \textbf{\small{}Subjective} &  & \tabularnewline
\hline 
\citet{Ahir2019Driver} & $\indicators{\heartRatePhysiological,\brainActivityPhysiological}$ & $\indicators{\gazeParametersBehavioral,\blinksBehavioral,\percentageClosureBehavioral,\facialExpressionsBehavioral,\bodyPostureBehavioral}$ &  & $\indicators{\wheelSteeringVehicle,\laneDisciplineVehicle,\speedVehicle}$ & \tabularnewline
\hline 
\citet{Alluhaibi2018Driver} &  & $\indicators{\speechBehavioral}$ &  & $\indicators{\wheelSteeringVehicle,\laneDisciplineVehicle,\brakingBehaviorVehicle,\speedVehicle}$ & \tabularnewline
\hline 
\citet{Arun2012Driver} & $\indicators{\heartRatePhysiological,\brainActivityPhysiological,\electrodermalActivityPhysiological,\pupilDiameterPhysiological}$ & $\indicators{\gazeParametersBehavioral,\blinksBehavioral,\bodyPostureBehavioral}$ & $\check$ & $\indicators{\wheelSteeringVehicle,\laneDisciplineVehicle,\brakingBehaviorVehicle,\speedVehicle}$ & \tabularnewline
\hline 
\citet{Balandong2018AReview} & $\indicators{\heartRatePhysiological,\brainActivityPhysiological}$ & $\indicators{\gazeParametersBehavioral,\blinksBehavioral,\percentageClosureBehavioral,\bodyPostureBehavioral}$ & $\check$ & $\indicators{\wheelSteeringVehicle,\laneDisciplineVehicle,\brakingBehaviorVehicle,\speedVehicle}$ & \tabularnewline
\hline 
\multirow{1}{4cm}{\citet{Begum2013Intelligent}} & $\indicators{\heartRatePhysiological,\brainActivityPhysiological}$ &  &  &  & \tabularnewline
\hline 
\citet{Chacon-murguia2015Detecting} & $\indicators{\heartRatePhysiological,\brainActivityPhysiological,\electrodermalActivityPhysiological}$ & $\indicators{\gazeParametersBehavioral,\blinksBehavioral,\bodyPostureBehavioral}$ &  & $\indicators{\wheelSteeringVehicle,\laneDisciplineVehicle,\brakingBehaviorVehicle,\speedVehicle}$ & \tabularnewline
\hline 
\multirow{1}{4cm}{\citet{Chan2019AComprehensive}} & $\indicators{\heartRatePhysiological,\brainActivityPhysiological}$ & $\indicators{\blinksBehavioral,\percentageClosureBehavioral,\facialExpressionsBehavioral,\bodyPostureBehavioral}$ &  & $\indicators{\wheelSteeringVehicle,\brakingBehaviorVehicle,\speedVehicle}$ & \tabularnewline
\hline 
\multirow{1}{4cm}{\citet{Chhabra2017ASurvey}} & $\indicators{\breathingActivityPhysiological}$ & $\indicators{\gazeParametersBehavioral,\percentageClosureBehavioral,\facialExpressionsBehavioral,\bodyPostureBehavioral}$ &  & $\indicators{\wheelSteeringVehicle}$ & $\indicators{\roadGeometryEnvironment}$\tabularnewline
\hline 
\citet{Chowdhury2018Sensor} & $\indicators{\heartRatePhysiological,\brainActivityPhysiological,\electrodermalActivityPhysiological}$ & $\indicators{\blinksBehavioral,\percentageClosureBehavioral}$ &  &  & \tabularnewline
\hline 
\citet{Chung2019Methods} & $\indicators{\heartRatePhysiological,\breathingActivityPhysiological,\brainActivityPhysiological,\electrodermalActivityPhysiological,\pupilDiameterPhysiological}$ & $\indicators{\speechBehavioral}$ & $\check$ & $\indicators{\wheelSteeringVehicle,\laneDisciplineVehicle,\brakingBehaviorVehicle,\speedVehicle}$ & \tabularnewline
\hline 
\citet{Coetzer2009Driver} & $\indicators{\brainActivityPhysiological}$ & $\indicators{\gazeParametersBehavioral,\percentageClosureBehavioral,\facialExpressionsBehavioral,\bodyPostureBehavioral}$ &  & $\indicators{\wheelSteeringVehicle,\laneDisciplineVehicle,\speedVehicle}$ & \tabularnewline
\hline 
\citet{Dababneh2015Driver} & $\indicators{\brainActivityPhysiological,\electrodermalActivityPhysiological,\pupilDiameterPhysiological}$ & $\indicators{\blinksBehavioral,\percentageClosureBehavioral,\bodyPostureBehavioral}$ &  & $\indicators{\wheelSteeringVehicle,\laneDisciplineVehicle,\speedVehicle}$ & $\indicators{\roadGeometryEnvironment}$\tabularnewline
\hline 
\citet{Dahiphale2015AReview} &  & $\indicators{\gazeParametersBehavioral,\blinksBehavioral,\facialExpressionsBehavioral,\bodyPostureBehavioral}$ &  & $\indicators{\wheelSteeringVehicle}$ & \tabularnewline
\hline 
\multirow{1}{4cm}{\citet{Dong2011Driver}} & $\indicators{\heartRatePhysiological,\brainActivityPhysiological,\pupilDiameterPhysiological}$ & $\indicators{\gazeParametersBehavioral,\blinksBehavioral,\percentageClosureBehavioral,\facialExpressionsBehavioral,\bodyPostureBehavioral}$ & $\check$ & $\indicators{\wheelSteeringVehicle,\laneDisciplineVehicle,\speedVehicle}$ & $\indicators{\roadGeometryEnvironment,\weatherEnvironment}$\tabularnewline
\hline 
\multirow{1}{4cm}{\citet{ElKhatib2020Driver}} & $\indicators{\heartRatePhysiological,\breathingActivityPhysiological,\brainActivityPhysiological,\electrodermalActivityPhysiological,\pupilDiameterPhysiological}$ & $\indicators{\gazeParametersBehavioral,\blinksBehavioral,\percentageClosureBehavioral,\facialExpressionsBehavioral,\bodyPostureBehavioral,\handsParametersBehavioral}$ &  & $\indicators{\wheelSteeringVehicle,\laneDisciplineVehicle,\speedVehicle}$ & \tabularnewline
\hline 
\multirow{1}{4cm}{\citet{Ghandour2020Driver}} & $\indicators{\heartRatePhysiological,\breathingActivityPhysiological,\brainActivityPhysiological,\electrodermalActivityPhysiological}$ & $\indicators{\gazeParametersBehavioral,\facialExpressionsBehavioral,\bodyPostureBehavioral,\speechBehavioral}$ & $\check$ & $\indicators{\wheelSteeringVehicle,\brakingBehaviorVehicle,\speedVehicle}$ & \tabularnewline
\hline 
\multirow{1}{4cm}{\citet{Hecht2018AReview}} & $\indicators{\heartRatePhysiological,\brainActivityPhysiological,\electrodermalActivityPhysiological,\pupilDiameterPhysiological}$ & $\indicators{\gazeParametersBehavioral,\blinksBehavioral,\percentageClosureBehavioral,\facialExpressionsBehavioral,\bodyPostureBehavioral}$ & $\check$ &  & \tabularnewline
\hline 
\multirow{1}{4cm}{\citet{Kang2013Various}} & $\indicators{\heartRatePhysiological,\breathingActivityPhysiological,\brainActivityPhysiological,\electrodermalActivityPhysiological}$ & $\indicators{\gazeParametersBehavioral,\blinksBehavioral,\facialExpressionsBehavioral,\bodyPostureBehavioral}$ &  & $\indicators{\wheelSteeringVehicle,\laneDisciplineVehicle,\brakingBehaviorVehicle,\speedVehicle}$ & \tabularnewline
\hline 
\multirow{1}{4cm}{\citet{Kaplan2015Driver}} & $\indicators{\heartRatePhysiological,\brainActivityPhysiological}$ & $\indicators{\gazeParametersBehavioral,\blinksBehavioral,\percentageClosureBehavioral,\facialExpressionsBehavioral,\bodyPostureBehavioral,\speechBehavioral}$ &  & $\indicators{\wheelSteeringVehicle,\laneDisciplineVehicle,\brakingBehaviorVehicle,\speedVehicle}$ & \tabularnewline
\hline 
\citet{Kaye2018Comparison} & $\indicators{\heartRatePhysiological,\breathingActivityPhysiological,\brainActivityPhysiological,\electrodermalActivityPhysiological}$ &  & $\check$ &  & \tabularnewline
\hline 
\citet{Khan2019AComprehensive} & $\indicators{\heartRatePhysiological,\brainActivityPhysiological,\electrodermalActivityPhysiological}$ & $\indicators{\gazeParametersBehavioral,\percentageClosureBehavioral,\bodyPostureBehavioral}$ &  & $\indicators{\wheelSteeringVehicle,\laneDisciplineVehicle,\brakingBehaviorVehicle,\speedVehicle}$ & \tabularnewline
\hline 
\citet{Kumari2017ASurvey} & $\indicators{\heartRatePhysiological,\brainActivityPhysiological}$ & $\indicators{\gazeParametersBehavioral,\blinksBehavioral,\percentageClosureBehavioral,\bodyPostureBehavioral}$ & $\check$ & $\indicators{\wheelSteeringVehicle,\laneDisciplineVehicle}$ & \tabularnewline
\hline 
\citet{Lal2001ACritical} & $\indicators{\heartRatePhysiological,\brainActivityPhysiological,\electrodermalActivityPhysiological}$ & $\indicators{\percentageClosureBehavioral,\facialExpressionsBehavioral}$ &  &  & \tabularnewline
\hline 
\citet{Laouz2020Literature} & $\indicators{\heartRatePhysiological,\brainActivityPhysiological,\electrodermalActivityPhysiological}$ & $\indicators{\blinksBehavioral,\percentageClosureBehavioral,\facialExpressionsBehavioral,\bodyPostureBehavioral}$ & $\check$ & $\indicators{\wheelSteeringVehicle,\speedVehicle}$ & \tabularnewline
\hline 
\citet{Leonhardt2018Unobtrusive} & $\indicators{\heartRatePhysiological,\breathingActivityPhysiological}$ &  &  &  & \tabularnewline
\hline 
\citet{Liu2019AReview} & $\indicators{\heartRatePhysiological,\brainActivityPhysiological,\pupilDiameterPhysiological}$ & $\indicators{\gazeParametersBehavioral,\blinksBehavioral,\percentageClosureBehavioral,\bodyPostureBehavioral}$ &  & $\indicators{\wheelSteeringVehicle,\laneDisciplineVehicle,\speedVehicle}$ & \tabularnewline
\hline 
\citet{Marquat2015Review} & $\indicators{\pupilDiameterPhysiological}$ & $\indicators{\gazeParametersBehavioral,\blinksBehavioral,\percentageClosureBehavioral}$ & $\check$ &  & \tabularnewline
\hline 
\citet{Martinez2017Driving} &  &  &  & $\indicators{\brakingBehaviorVehicle,\speedVehicle}$ & \tabularnewline
\hline 
\citet{Mashko2015Review} & $\indicators{\heartRatePhysiological,\brainActivityPhysiological,\electrodermalActivityPhysiological}$ & $\indicators{\gazeParametersBehavioral,\blinksBehavioral,\bodyPostureBehavioral}$ &  & $\indicators{\wheelSteeringVehicle,\laneDisciplineVehicle,\brakingBehaviorVehicle,\speedVehicle}$ & \tabularnewline
\hline 
\citet{Mashru2018Detection} & $\indicators{\heartRatePhysiological,\breathingActivityPhysiological}$ & $\indicators{\blinksBehavioral,\percentageClosureBehavioral,\facialExpressionsBehavioral,\bodyPostureBehavioral}$ & $\check$ & $\indicators{\wheelSteeringVehicle,\laneDisciplineVehicle}$ & \tabularnewline
\hline 
\citet{Melnicuk2016Towards} & $\indicators{\heartRatePhysiological,\brainActivityPhysiological}$ & $\indicators{\blinksBehavioral,\percentageClosureBehavioral,\facialExpressionsBehavioral}$ &  & $\indicators{\wheelSteeringVehicle,\brakingBehaviorVehicle,\speedVehicle}$ & $\indicators{\roadGeometryEnvironment,\trafficDensityEnvironment,\weatherEnvironment}$\tabularnewline
\hline 
\citet{Mittal2016Head} & $\indicators{\heartRatePhysiological,\brainActivityPhysiological,\pupilDiameterPhysiological}$ & $\indicators{\blinksBehavioral,\percentageClosureBehavioral,\bodyPostureBehavioral}$ & $\check$ & $\indicators{\wheelSteeringVehicle,\laneDisciplineVehicle,\brakingBehaviorVehicle,\speedVehicle}$ & \tabularnewline
\hline 
\citet{Murugan2019Analysis} & $\indicators{\heartRatePhysiological,\breathingActivityPhysiological,\brainActivityPhysiological,\electrodermalActivityPhysiological,\pupilDiameterPhysiological}$ & $\indicators{\blinksBehavioral,\percentageClosureBehavioral,\bodyPostureBehavioral}$ & $\check$ & $\indicators{\wheelSteeringVehicle,\laneDisciplineVehicle,\speedVehicle}$ & \tabularnewline
\hline 
\citet{Nair2016ASurvey} &  & $\indicators{\gazeParametersBehavioral,\percentageClosureBehavioral,\facialExpressionsBehavioral,\bodyPostureBehavioral}$ &  & $\indicators{\laneDisciplineVehicle}$ & \tabularnewline
\hline 
\citet{Nemcova2021Multimodal} & $\indicators{\heartRatePhysiological,\breathingActivityPhysiological,\brainActivityPhysiological,\electrodermalActivityPhysiological}$ & $\indicators{\gazeParametersBehavioral,\blinksBehavioral,\percentageClosureBehavioral,\facialExpressionsBehavioral,\bodyPostureBehavioral}$ &  & $\indicators{\wheelSteeringVehicle,\laneDisciplineVehicle,\brakingBehaviorVehicle,\speedVehicle}$ & \tabularnewline
\hline 
\citet{Ngxande2017Driver} &  & $\indicators{\blinksBehavioral,\percentageClosureBehavioral,\facialExpressionsBehavioral,\bodyPostureBehavioral}$ &  &  & \tabularnewline
\hline 
\citet{Oviedo-Trespalacios2016Understanding} &  & $\indicators{\gazeParametersBehavioral}$ &  & $\indicators{\wheelSteeringVehicle,\laneDisciplineVehicle,\brakingBehaviorVehicle,\speedVehicle}$ & \tabularnewline
\hline 
\citet{Papantoniou2017Review} & $\indicators{\heartRatePhysiological,\breathingActivityPhysiological,\brainActivityPhysiological}$ & $\indicators{\gazeParametersBehavioral,\blinksBehavioral,\speechBehavioral}$ & $\check$ & $\indicators{\wheelSteeringVehicle,\laneDisciplineVehicle,\speedVehicle}$ & \tabularnewline
\hline 
\citet{Pratama2017AReview} & $\indicators{\heartRatePhysiological,\brainActivityPhysiological,\electrodermalActivityPhysiological}$ & $\indicators{\gazeParametersBehavioral,\blinksBehavioral,\percentageClosureBehavioral,\facialExpressionsBehavioral,\bodyPostureBehavioral,\handsParametersBehavioral}$ & $\check$ & $\indicators{\wheelSteeringVehicle,\laneDisciplineVehicle}$ & \tabularnewline
\hline 
\citet{Ramzan2019ASurvey} & $\indicators{\heartRatePhysiological,\breathingActivityPhysiological,\brainActivityPhysiological}$ & $\indicators{\blinksBehavioral,\percentageClosureBehavioral,\facialExpressionsBehavioral,\bodyPostureBehavioral}$ &  & $\indicators{\wheelSteeringVehicle,\laneDisciplineVehicle,\speedVehicle}$ & \tabularnewline
\hline 
\citet{Sahayadhas2012Detecting} & $\indicators{\heartRatePhysiological,\brainActivityPhysiological,\pupilDiameterPhysiological}$ & $\indicators{\gazeParametersBehavioral,\blinksBehavioral,\percentageClosureBehavioral,\bodyPostureBehavioral}$ & $\check$ & $\indicators{\wheelSteeringVehicle,\laneDisciplineVehicle}$ & \tabularnewline
\hline 
\citet{Scott-Parker2017Emotions} & $\indicators{\heartRatePhysiological,\brainActivityPhysiological,\electrodermalActivityPhysiological}$ & $\indicators{\gazeParametersBehavioral,\facialExpressionsBehavioral}$ & $\check$ & $\indicators{\wheelSteeringVehicle,\laneDisciplineVehicle,\brakingBehaviorVehicle,\speedVehicle}$ & $\indicators{\trafficDensityEnvironment}$\tabularnewline
\hline 
\citet{Seth2020ASurvey} &  &  &  &  & \tabularnewline
\hline 
\citet{Shameen2018Electroencephalography} & $\indicators{\brainActivityPhysiological}$ & $\indicators{\gazeParametersBehavioral,\blinksBehavioral}$ &  &  & \tabularnewline
\hline 
\citet{Sigari2014AReview} &  & $\indicators{\gazeParametersBehavioral,\blinksBehavioral,\percentageClosureBehavioral,\facialExpressionsBehavioral,\bodyPostureBehavioral}$ &  &  & \tabularnewline
\hline 
\citet{Sikander2019Driver} & $\indicators{\heartRatePhysiological,\brainActivityPhysiological,\pupilDiameterPhysiological}$ & $\indicators{\gazeParametersBehavioral,\blinksBehavioral,\percentageClosureBehavioral,\bodyPostureBehavioral}$ & $\check$ & $\indicators{\wheelSteeringVehicle,\laneDisciplineVehicle}$ & \tabularnewline
\hline 
\citet{Singh2021Analyzing} & $\indicators{\pupilDiameterPhysiological}$ & $\indicators{\gazeParametersBehavioral,\blinksBehavioral,\percentageClosureBehavioral,\facialExpressionsBehavioral}$ &  & $\indicators{\wheelSteeringVehicle,\brakingBehaviorVehicle,\speedVehicle}$ & $\indicators{\roadGeometryEnvironment,\trafficDensityEnvironment}$\tabularnewline
\hline 
\citet{Subbaiah2019Driver} & $\indicators{\heartRatePhysiological,\brainActivityPhysiological,\pupilDiameterPhysiological}$ & $\indicators{\blinksBehavioral,\percentageClosureBehavioral,\facialExpressionsBehavioral,\bodyPostureBehavioral}$ &  &  & \tabularnewline
\hline 
\citet{Tu2016ASurvey} & $\indicators{\heartRatePhysiological,\brainActivityPhysiological}$ & $\indicators{\blinksBehavioral,\percentageClosureBehavioral,\facialExpressionsBehavioral,\bodyPostureBehavioral}$ &  & $\indicators{\wheelSteeringVehicle,\laneDisciplineVehicle,\speedVehicle}$ & \tabularnewline
\hline 
\citet{Ukwuoma2019Deep} & $\indicators{\heartRatePhysiological,\breathingActivityPhysiological,\brainActivityPhysiological}$ & $\indicators{\blinksBehavioral,\percentageClosureBehavioral,\facialExpressionsBehavioral,\bodyPostureBehavioral}$ &  & $\indicators{\wheelSteeringVehicle,\laneDisciplineVehicle,\brakingBehaviorVehicle}$ & \tabularnewline
\hline 
\citet{Vilaca2017Systematic} & $\indicators{\brainActivityPhysiological}$ & $\indicators{\gazeParametersBehavioral,\bodyPostureBehavioral}$ &  & $\indicators{\wheelSteeringVehicle,\laneDisciplineVehicle,\brakingBehaviorVehicle,\speedVehicle}$ & \tabularnewline
\hline 
\citet{Vismaya2020AReview} &  & $\indicators{\gazeParametersBehavioral,\blinksBehavioral,\percentageClosureBehavioral,\bodyPostureBehavioral}$ &  &  & \tabularnewline
\hline 
\citet{Wang2006Driver} & $\indicators{\brainActivityPhysiological,\pupilDiameterPhysiological}$ & $\indicators{\gazeParametersBehavioral,\blinksBehavioral,\percentageClosureBehavioral,\bodyPostureBehavioral}$ &  & $\indicators{\laneDisciplineVehicle}$ & \tabularnewline
\hline 
\citet{Welch2019AReview} & $\indicators{\heartRatePhysiological,\breathingActivityPhysiological,\brainActivityPhysiological,\electrodermalActivityPhysiological}$ & $\indicators{\blinksBehavioral,\facialExpressionsBehavioral,\speechBehavioral}$ &  & $\indicators{\wheelSteeringVehicle,\brakingBehaviorVehicle,\speedVehicle}$ & \tabularnewline
\hline 
\citet{Yusoff2017Selection} & $\indicators{\heartRatePhysiological,\brainActivityPhysiological,\electrodermalActivityPhysiological,\pupilDiameterPhysiological}$ & $\indicators{\gazeParametersBehavioral,\bodyPostureBehavioral}$ & $\check$ & $\indicators{\laneDisciplineVehicle,\speedVehicle}$ & \tabularnewline
\hline 
\citet{Zhang2013Review} & $\indicators{\heartRatePhysiological,\brainActivityPhysiological}$ & $\indicators{\gazeParametersBehavioral,\blinksBehavioral,\percentageClosureBehavioral,\bodyPostureBehavioral}$ &  & $\indicators{\laneDisciplineVehicle,\speedVehicle}$ & \tabularnewline
\hline 
\end{tabular}}
\end{table}

\pagebreak{}

\section{Effects of blood alcohol concentration\label{appendix:effects-BAC}}

As of this writing (in mid 2021), the NHTSA website contains a webpage
about “Drunk Driving”, which features a table entitled “The Effects
of Blood Alcohol Concentration”. Table~\ref{tab-effects-of-BAC}
reproduces this table, nearly verbatim, in compliance with the “Terms
of Use” of the website. The table shows, as a function of the level
of blood alcohol concentration (BAC) (in $\gdl$), (1) the typical
effects, independently of any task, and (2) the predictable effects
for the specific task of driving.

\begin{table}
\caption{This table gives the effects of blood alcohol concentration (BAC).
It is a nearly-verbatim reproduction of a table present on the NHTSA
website in mid 2021.\label{tab-effects-of-BAC}}

\centering{}\resizebox{\textwidth}{!}{%
\begin{tabular}{|>{\centering}m{0.1\textwidth}|>{\raggedright}p{0.45\textwidth}|>{\raggedright}p{0.45\textwidth}|}
\hline 
\textbf{\small{}BAC}{\small\par}

\textbf{\small{}(in $\gdl$)} & \textbf{\small{}Typical effects} & \textbf{\small{}Predictable effects on driving}\tabularnewline
\hline 
\centering{}{\small{}$0.02$} & {\small{}Some loss of judgment; relaxation, slight body warmth, altered
mood} & {\small{}Decline in visual functions (rapid tracking of a moving target),
decline in ability to perform two tasks at the same time (divided
attention)}\tabularnewline
\hline 
\centering{}{\small{}$0.05$} & {\small{}Exaggerated behavior, may have loss of small-muscle control
(\eg, focusing your eyes), impaired judgment, usually good feeling,
lowered alertness, release of inhibition} & {\small{}Reduced coordination, reduced ability to track moving objects,
difficulty steering, reduced response to emergency driving situations}\tabularnewline
\hline 
\centering{}{\small{}$0.08$} & {\small{}Muscle coordination becomes poor (\eg, balance, speech,
vision, reaction time, and hearing), harder to detect danger; judgment,
self-control, reasoning, and memory are impaired} & {\small{}Concentration, short-term memory loss, speed control, reduced
information processing capability (\eg, signal detection, visual
search), impaired perception}\tabularnewline
\hline 
\centering{}{\small{}$0.10$} & {\small{}Clear deterioration of reaction time and control, slurred
speech, poor coordination, and slowed thinking} & {\small{}Reduced ability to maintain lane position and brake appropriately}\tabularnewline
\hline 
\centering{}{\small{}$0.15$} & {\small{}Far less muscle control than normal, vomiting may occur (unless
this level is reached slowly or a person has developed a tolerance
for alcohol), major loss of balance} & {\small{}Substantial impairment in vehicle control, attention to driving
task, and in necessary visual and auditory information processing}\tabularnewline
\hline 
\end{tabular}}
\end{table}

\section{Growth of literature on driver monitoring}

The survey of Section~\ref{sec:bibliographic-study} provided an
initial set of $\numberOfSurveys$ references for the field of DM.
They appear in Table~\ref{tab-reviews-summary-sumvers}. Our overall
analysis and synthesis of the field led us to examine in detail a
total of $\numberOfArticles$ references, including the $\numberOfSurveys$
initial ones. They all appear in the “References” section.

\begin{figure}
\centering{}%
\begin{tabular}{cc}
\includegraphics[width=0.48\columnwidth]{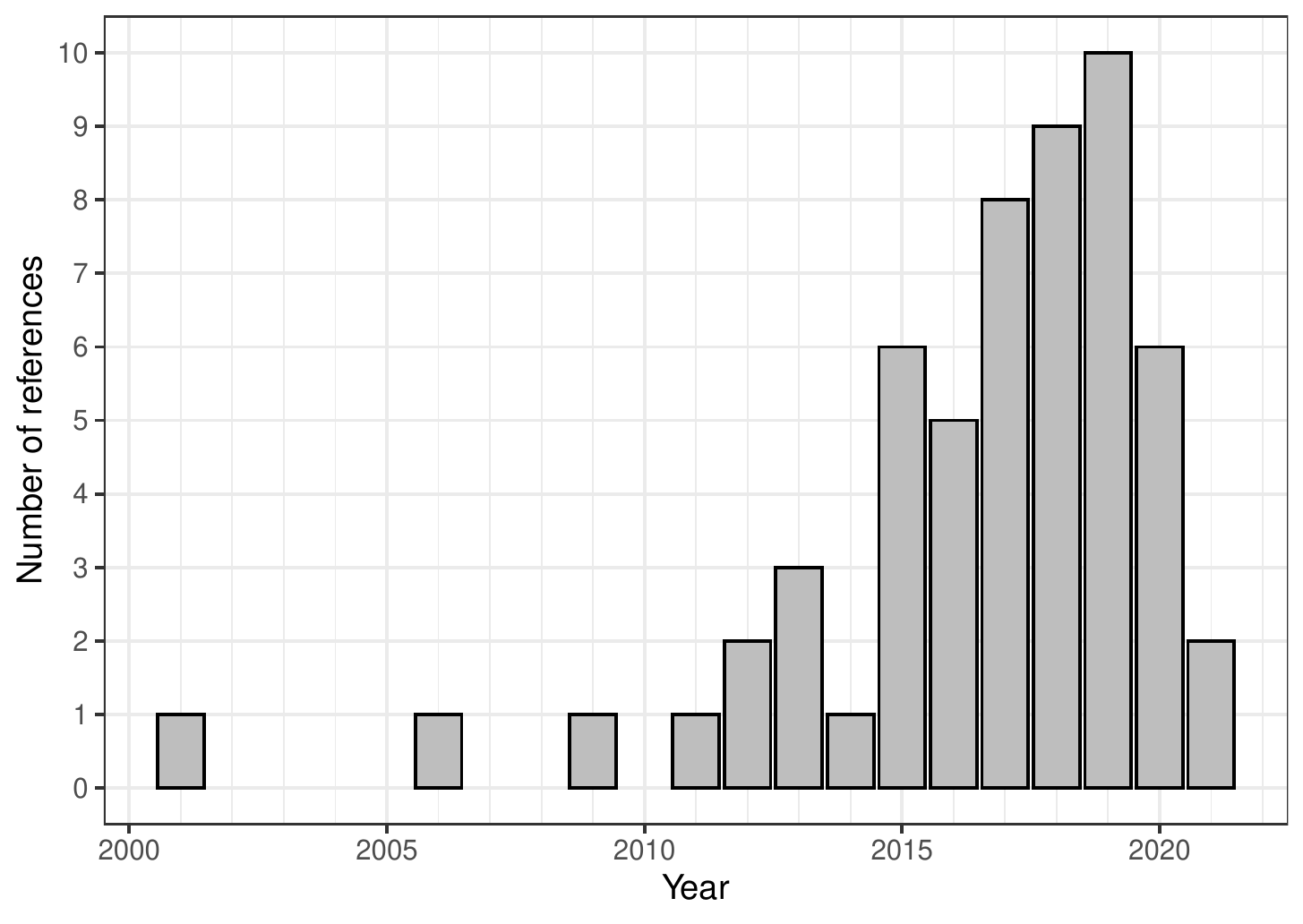} & \includegraphics[width=0.48\columnwidth]{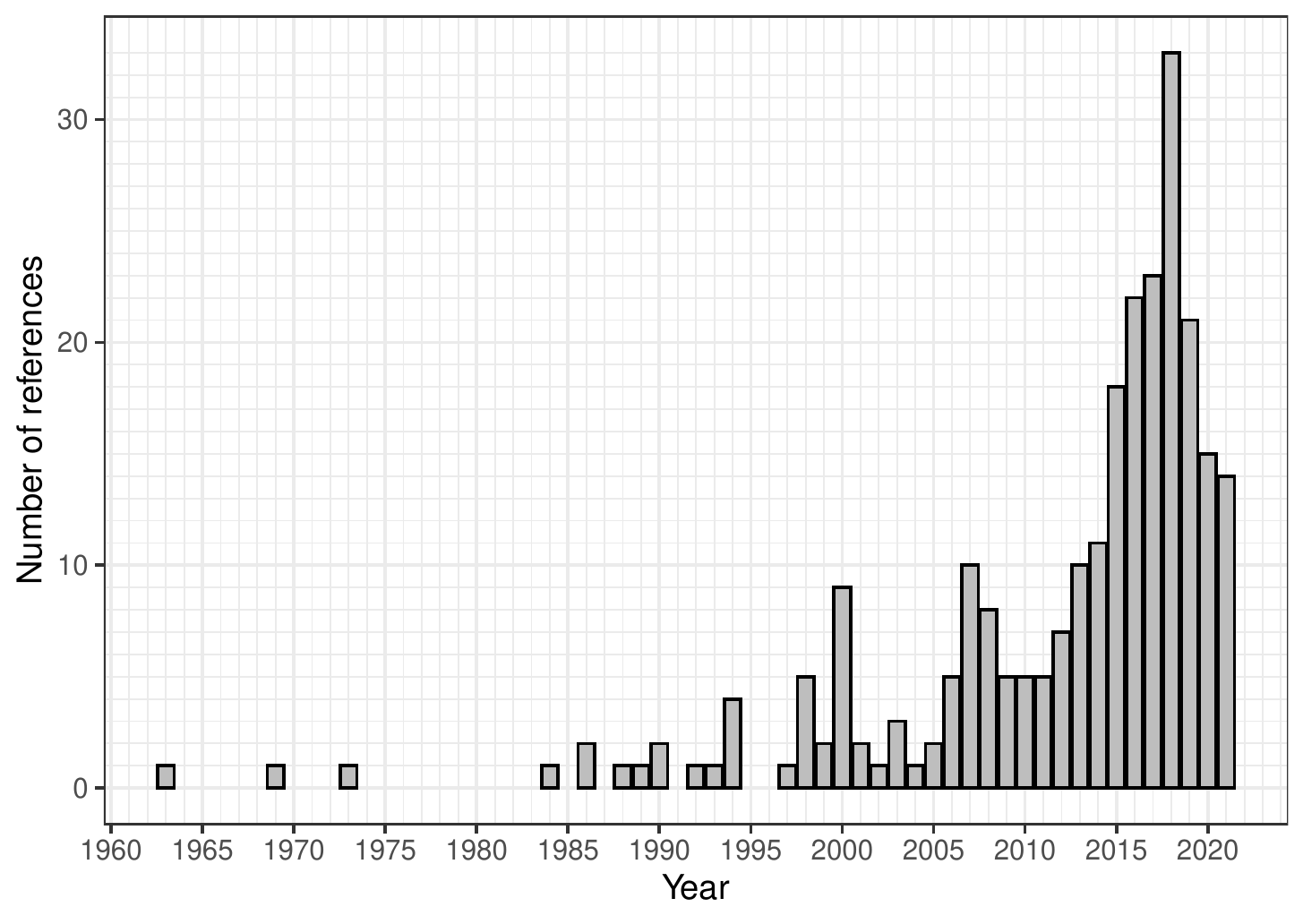}\tabularnewline
(a) & (b)\tabularnewline
\end{tabular}\caption{Graph~(a) is the histogram of the number of references vs year for
the $\protect\numberOfSurveys$ initial references on driver monitoring
(DM), and graph~(b) is the corresponding histogram for the $\protect\numberOfArticles$
examined references. These histograms suggest that the field of DM
has been the object of growing interest over the years and, in particular,
over the last $10$ years.\label{fig:Histograms}}
\end{figure}

To characterize, in an approximate way, the evolution of the number
of publications on DM over recent years, (1) we computed, for the
$\numberOfSurveys$ initial references, the number of them published
during each of the years they cover, and (2) we did the same for the
$\numberOfArticles$ examined references. Figure~\ref{fig:Histograms}
gives the corresponding graphs, or histograms, of ``number of references
vs year''. Each histogram shows a significant growth over the last
$10$ years or so. The significant dip in 2020 could be an effect
of the difficult worldwide situation during that year.

\begin{thebibliography}{254}
\providecommand{\natexlab}[1]{#1}
\providecommand{\url}[1]{\texttt{#1}}
\expandafter\ifx\csname urlstyle\endcsname\relax
  \providecommand{\doi}[1]{doi: #1}\else
  \providecommand{\doi}{doi: \begingroup \urlstyle{rm}\Url}\fi

\bibitem[Aaronson et~al.(2007)Aaronson, Teel, Cassmeyer, Neuberger,
  Pallikkathayil, Pierce, Press, Williams, and Wingate]{Aaronson2007Defining}
L.~Aaronson, C.~Teel, V.~Cassmeyer, G.~Neuberger, L.~Pallikkathayil, J.~Pierce,
  A.~Press, P.~Williams, and A.~Wingate.
\newblock Defining and measuring fatigue.
\newblock \emph{Journal of Nursing Scholarship}, 31\penalty0 (1):\penalty0
  45--50, June 2007.
\newblock \doi{10.1111/j.1547-5069.1999.tb00420.x}.
\newblock URL \url{https://doi.org/10.1111/j.1547-5069.1999.tb00420.x}.

\bibitem[Ahir and Gohokar(2019)]{Ahir2019Driver}
A.~Ahir and V.~Gohokar.
\newblock Driver inattention monitoring system: A review.
\newblock In \emph{International Conference on Innovative Trends and Advances
  in Engineering and Technology (ICITAET)}, pages 188--194, Shegoaon, India,
  December 2019. IEEE.
\newblock \doi{10.1109/ICITAET47105.2019.9170249}.
\newblock URL \url{https://doi.org/10.1109/ICITAET47105.2019.9170249}.

\bibitem[Ahlstr{\"o}m et~al.(2021)Ahlstr{\"o}m, Georgoulas, and
  Kircher]{Ahlstrom2021Towards}
C.~Ahlstr{\"o}m, G.~Georgoulas, and K.~Kircher.
\newblock Towards a context-dependent multi-buffer driver distraction detection
  algorithm.
\newblock \emph{IEEE Transactions on Intelligent Transportation Systems}, 2021.
\newblock \doi{10.1109/TITS.2021.3060168}.
\newblock URL \url{https://doi.org/10.1109/TITS.2021.3060168}.

\bibitem[Aidman et~al.(2015)Aidman, Chadunow, Johnson, and
  Reece]{Aidman2015RealTime}
E.~Aidman, C.~Chadunow, K.~Johnson, and J.~Reece.
\newblock Real-time driver drowsiness feedback improves driver alertness and
  self-reported driving performance.
\newblock \emph{Accident Analysis \& Prevention}, 81:\penalty0 8--13, August
  2015.
\newblock \doi{10.1016/j.aap.2015.03.041}.
\newblock URL \url{https://doi.org/10.1016/j.aap.2015.03.041}.

\bibitem[{\AA}kerstedt and Gillberg(1990)]{Akerstedt1990Subjective}
T.~{\AA}kerstedt and M.~Gillberg.
\newblock Subjective and objective sleepiness in the active individual.
\newblock \emph{International Journal of Neuroscience}, 52\penalty0
  (1-2):\penalty0 29--37, June 1990.
\newblock \doi{10.3109/00207459008994241}.
\newblock URL \url{https://doi.org/10.3109/00207459008994241}.

\bibitem[Alluhaibi et~al.(2018)Alluhaibi, {Al-Din}, and
  Moyaid]{Alluhaibi2018Driver}
S.~Alluhaibi, M.~{Al-Din}, and A.~Moyaid.
\newblock Driver behavior detection techniques: A survey.
\newblock \emph{International Journal of Applied Engineering Research},
  13\penalty0 (11):\penalty0 8856--8861, 2018.
\newblock URL \url{https://www.ripublication.com/Volume/ijaerv13n11.htm}.

\bibitem[Almahasneh et~al.(2014)Almahasneh, Chooi, Kamel, and
  Malik]{Almahasneh2014Deep}
H.~Almahasneh, W.-T. Chooi, N.~Kamel, and A.~Malik.
\newblock Deep in thought while driving: An {EEG} study on drivers' cognitive
  distraction.
\newblock \emph{Transportation Research Part F: Traffic Psychology and
  Behaviour}, 26, Part A:\penalty0 218--226, September 2014.
\newblock \doi{10.1016/j.trf.2014.08.001}.
\newblock URL \url{https://doi.org/10.1016/j.trf.2014.08.001}.

\bibitem[Alonso(2019)]{Alonso2019Driving}
F.~Alonso.
\newblock \emph{Driving under the influence}, volume~1, pages 392--394.
\newblock SAGE, September 2019.
\newblock \doi{10.4135/9781483392240.n130}.
\newblock URL \url{https://doi.org/10.4135/9781483392240.n130}.

\bibitem[Alonso et~al.(2015)Alonso, Pastor, Montoro, and
  Esteban]{Alonso2015Driving}
F.~Alonso, J.~Pastor, L.~Montoro, and C.~Esteban.
\newblock Driving under the influence of alcohol: frequency, reasons, perceived
  risk and punishment.
\newblock \emph{Substance Abuse Treatment, Prevention, and Policy}, 10\penalty0
  (11):\penalty0 1--9, March 2015.
\newblock \doi{10.1186/s13011-015-0007-4}.
\newblock URL \url{https://doi.org/10.1186/s13011-015-0007-4}.

\bibitem[Anund et~al.(2008)Anund, Kecklund, Peters, Forsman, Arne, and
  {\AA}kerstedt]{Anund2008Driver}
A.~Anund, G.~Kecklund, B.~Peters, {\AA}.~Forsman, L.~Arne, and
  T.~{\AA}kerstedt.
\newblock Driver impairment at night and its relation to physiological
  sleepiness.
\newblock \emph{Scandinavian Journal of Work, Environment \& Health},
  34\penalty0 (2):\penalty0 142--150, April 2008.
\newblock \doi{10.5271/sjweh.1193}.
\newblock URL \url{https://doi.org/10.5271/sjweh.1193}.

\bibitem[Apostoloff and Zelinsky(2003)]{Apostoloff2003Robust}
N.~Apostoloff and A.~Zelinsky.
\newblock Robust vision based lane tracking using multiple cues and particle
  filtering.
\newblock In \emph{IEEE Intelligent Vehicles Symposium. Proceedings (Cat.
  No.03TH8683)}, pages 558--563, Columbus, Ohio, USA, June 2003.
\newblock \doi{10.1109/IVS.2003.1212973}.
\newblock URL \url{https://doi.org/10.1109/IVS.2003.1212973}.

\bibitem[Arnedt et~al.(2000)Arnedt, Wilde, Munt, and
  MacLean]{Arnedt2000Simulated}
J.~Arnedt, G.~Wilde, P.~Munt, and A.~MacLean.
\newblock Simulated driving performance following prolonged wakefulness and
  alcohol consumption: separate and combined contributions to impairment.
\newblock \emph{Journal of Sleep Research}, 9\penalty0 (3):\penalty0 233--241,
  September 2000.
\newblock \doi{10.1046/j.1365-2869.2000.00216.x}.
\newblock URL \url{https://doi.org/10.1046/j.1365-2869.2000.00216.x}.

\bibitem[Arun et~al.(2012)Arun, Sundaraj, and Murugappan]{Arun2012Driver}
S.~Arun, K.~Sundaraj, and M.~Murugappan.
\newblock Driver inattention detection methods: A review.
\newblock In \emph{IEEE Conference on Sustainable Utilization and Development
  in Engineering and Technology (STUDENT)}, pages 1--6, Kuala Lumpur, Malaysia,
  October 2012.
\newblock \doi{10.1109/STUDENT.2012.6408351}.
\newblock URL \url{https://doi.org/10.1109/STUDENT.2012.6408351}.

\bibitem[Attia et~al.(2016)Attia, Takruri, and Ali]{Attia2016Electronic}
H.~Attia, M.~Takruri, and H.~Ali.
\newblock Electronic monitoring and protection system for drunk driver based on
  breath sample testing.
\newblock In \emph{International Conference on Electronic Devices, Systems and
  Applications (ICEDSA)}, pages 1--4, Ras Al Khaimah, United Arab Emirates,
  December 2016. IEEE.
\newblock \doi{10.1109/ICEDSA.2016.7818477}.
\newblock URL \url{https://doi.org/10.1109/ICEDSA.2016.7818477}.

\bibitem[Baccour et~al.(2020)Baccour, Driewer, Schack, and
  Kasneci]{Baccour2020Camera}
M.~Baccour, F.~Driewer, T.~Schack, and E.~Kasneci.
\newblock Camera-based driver drowsiness state classification using logistic
  regression models.
\newblock In \emph{IEEE International Conference on Systems, Man, and
  Cybernetics (SMC)}, pages 1--8, Toronto, Ontario, Canada, October 2020. IEEE.
\newblock \doi{10.1109/SMC42975.2020.9282918}.
\newblock URL \url{https://doi.org/10.1109/SMC42975.2020.9282918}.

\bibitem[Baheti et~al.(2018)Baheti, Gajre, and Talbar]{Baheti2018Detection}
B.~Baheti, S.~Gajre, and S.~Talbar.
\newblock Detection of distracted driver using convolutional neural network.
\newblock In \emph{IEEE International Conference on Computer Vision and Pattern
  Recognition Workshops (CVPRW)}, pages 1032--1038, Salt Lake City, Utah, USA,
  June 2018.
\newblock \doi{10.1109/CVPRW.2018.00150}.
\newblock URL \url{https://doi.org/10.1109/CVPRW.2018.00150}.

\bibitem[Bakker et~al.(2021)Bakker, Zab{\l}ocki, Baker, Riethmeister, Marx,
  Iyer, Anund, and Ahlstr{\"o}m]{Bakker2021AMultiStage}
B.~Bakker, B.~Zab{\l}ocki, A.~Baker, V.~Riethmeister, B.~Marx, G.~Iyer,
  A.~Anund, and C.~Ahlstr{\"o}m.
\newblock A multi-stage, multi-feature machine learning approach to detect
  driver sleepiness in naturalistic road driving conditions.
\newblock \emph{IEEE Transactions on Intelligent Transportation Systems}, Early
  access:\penalty0 1--10, June 2021.
\newblock \doi{10.1109/TITS.2021.3090272}.
\newblock URL \url{https://doi.org/10.1109/TITS.2021.3090272}.

\bibitem[Balandong et~al.(2018)Balandong, Ahmad, Saad, and
  Malik]{Balandong2018AReview}
R.~Balandong, R.~Ahmad, M.~Saad, and A.~Malik.
\newblock A review on {EEG}-based automatic sleepiness detection systems for
  driver.
\newblock \emph{IEEE Access}, 6:\penalty0 22908--22919, December 2018.
\newblock \doi{10.1109/ACCESS.2018.2811723}.
\newblock URL \url{https://doi.org/10.1109/ACCESS.2018.2811723}.

\bibitem[Basu et~al.(2017)Basu, Chakraborty, Bagb, and
  Aftabuddin]{Basu2017AReview}
S.~Basu, J.~Chakraborty, A.~Bagb, and Md. Aftabuddin.
\newblock A review on emotion recognition using speech.
\newblock In \emph{International Conference on Inventive Communication and
  Computational Technologies (ICICCT)}, pages 109--114, Coimbatore, India,
  March 2017.
\newblock \doi{10.1109/ICICCT.2017.7975169}.
\newblock URL \url{https://doi.org/10.1109/ICICCT.2017.7975169}.

\bibitem[Begum(2013)]{Begum2013Intelligent}
S.~Begum.
\newblock Intelligent driver monitoring systems based on physiological sensor
  signals: A review.
\newblock In \emph{International Conference on Intelligent Transportation
  Systems (ITSC)}, pages 282--289, The Hague, The Netherlands, October 2013.
  IEEE.
\newblock \doi{10.1109/ITSC.2013.6728246}.
\newblock URL \url{https://doi.org/10.1109/ITSC.2013.6728246}.

\bibitem[Bergasa et~al.(2006)Bergasa, Nuevo, Sotelo, Barea, and
  Lopez]{Bergasa2006RealTime}
L.~Bergasa, J.~Nuevo, M.~Sotelo, R.~Barea, and M.~Lopez.
\newblock Real-time system for monitoring driver vigilance.
\newblock \emph{IEEE Transactions on Intelligent Transportation Systems},
  7\penalty0 (1):\penalty0 63--77, 2006.
\newblock \doi{10.1109/TITS.2006.869598}.
\newblock URL \url{https://dx.doi.org/10.1109/TITS.2006.869598}.

\bibitem[Berri and Os{\'o}rio(2018)]{Berri2018ANonintrusive}
R.~Berri and F.~Os{\'o}rio.
\newblock A nonintrusive system for detecting drunk drivers in modern vehicles.
\newblock In \emph{Brazilian Conference on Intelligent Systems (BRACIS)}, pages
  73--78, S{\~a}o Paulo, Brazil, October 2018. IEEE.
\newblock \doi{10.1109/BRACIS.2018.00021}.
\newblock URL \url{https:/doi.org/10.1109/BRACIS.2018.00021}.

\bibitem[Borghini et~al.(2014)Borghini, Astolfi, Vecchiato, Mattia, and
  Babiloni]{Borghini2014Measuring}
G.~Borghini, L.~Astolfi, G.~Vecchiato, D.~Mattia, and F.~Babiloni.
\newblock Measuring neurophysiological signals in aircraft pilots and car
  drivers for the assessment of mental workload, fatigue and drowsiness.
\newblock \emph{Neuroscience \& Biobehavioral Reviews}, 44:\penalty0 58--75,
  July 2014.
\newblock \doi{10.1016/j.neubiorev.2012.10.003}.
\newblock URL \url{https://doi.org/10.1016/j.neubiorev.2012.10.003}.

\bibitem[Bo{\v{r}}il et~al.(2012)Bo{\v{r}}il, Boyraz, and
  Hansen]{Boril2012Towards}
H.~Bo{\v{r}}il, P.~Boyraz, and J.~Hansen.
\newblock Towards multimodal driver's stress detection.
\newblock In \emph{Digital Signal Processing for In-Vehicle Systems and
  Safety}, pages 3--19. Springer, New York City, New York, USA, 2012.
\newblock \doi{10.1007/978-1-4419-9607-7_1}.
\newblock URL \url{https://doi.org/10.1007/978-1-4419-9607-7_1}.

\bibitem[Bradley and Lang(1994)]{Bradley1994Measuring}
M.~Bradley and P.~Lang.
\newblock Measuring emotion: the self-assessment manikin and the semantic
  differential.
\newblock \emph{Journal of Behavior Therapy and Experimental Psychiatry},
  25\penalty0 (1):\penalty0 49--59, 1994.
\newblock \doi{10.1016/0005-7916(94)90063-9}.
\newblock URL \url{https://doi.org/10.1016/0005-7916(94)90063-9}.

\bibitem[Brown et~al.(2006)Brown, Marmor, Vaegan, Zrenner, Brigell, and
  Bach]{Brown2006ISCEV}
M.~Brown, M.~Marmor, Vaegan, E.~Zrenner, M.~Brigell, and M.~Bach.
\newblock {ISCEV} standard for clinical electro-oculography ({EOG}) 2006.
\newblock \emph{Documenta Ophthalmologica}, 113:\penalty0 205--212, November
  2006.
\newblock \doi{10.1007/s10633-006-9030-0}.
\newblock URL \url{https://doi.org/10.1007/s10633-006-9030-0}.

\bibitem[Campbell(2012)]{Campbell2012TheSHRP}
K.~Campbell.
\newblock The {SHRP} 2 naturalistic driving study.
\newblock \emph{TR News}, 282:\penalty0 30--35, September 2012.
\newblock URL \url{https://insight.shrp2nds.us/documents/shrp2_background.pdf}.

\bibitem[{Chacon-Murguia} and
  {Prieto-Resendiz}(2015)]{Chacon-murguia2015Detecting}
M.~{Chacon-Murguia} and C.~{Prieto-Resendiz}.
\newblock Detecting driver drowsiness: A survey of system designs and
  technology.
\newblock \emph{IEEE Consumer Electronics Magazine}, 4\penalty0 (4):\penalty0
  107--119, October 2015.
\newblock \doi{10.1109/MCE.2015.2463373}.
\newblock URL \url{https://doi.org/10.1109/MCE.2015.2463373}.

\bibitem[Chan et~al.(2020)Chan, Chin, Chen, and Zhong]{Chan2019AComprehensive}
T.~Chan, C.~Chin, H.~Chen, and X.~Zhong.
\newblock A comprehensive review of driver behavior analysis utilizing
  smartphones.
\newblock \emph{IEEE Transactions on Intelligent Transportation Systems},
  21\penalty0 (10):\penalty0 4444--4475, 2020.
\newblock \doi{10.1109/TITS.2019.2940481}.
\newblock URL \url{https://doi.org/10.1109/TITS.2019.2940481}.

\bibitem[Charniya and Nair(2017)]{Nair2017Drunk}
N.~Charniya and V.~Nair.
\newblock Drunk driving and drowsiness detection.
\newblock In \emph{International Conference on Intelligent Computing and
  Control (I2C2)}, pages 1--6, Coimbatore, India, June 2017. IEEE.
\newblock \doi{10.1109/I2C2.2017.8321811}.
\newblock URL \url{https://doi.org/10.1109/I2C2.2017.8321811}.

\bibitem[Chhabra et~al.(2017)Chhabra, Verma, and Krishna]{Chhabra2017ASurvey}
R.~Chhabra, S.~Verma, and C.~R. Krishna.
\newblock A survey on driver behavior detection techniques for intelligent
  transportation systems.
\newblock In \emph{International Conference on Cloud Computing, Data Science \&
  Engineering -- Confluence}, pages 36--41, Noida, India, January 2017. IEEE.
\newblock \doi{10.1109/CONFLUENCE.2017.7943120}.
\newblock URL \url{https://doi.org/10.1109/CONFLUENCE.2017.7943120}.

\bibitem[Chowdhury et~al.(2018)Chowdhury, Shankaran, Kavakli, and
  Haque]{Chowdhury2018Sensor}
A.~Chowdhury, R.~Shankaran, M.~Kavakli, and M.~M. Haque.
\newblock Sensor applications and physiological features in drivers' drowsiness
  detection: A review.
\newblock \emph{IEEE Sensors Journal}, 18\penalty0 (8):\penalty0 3055--3067,
  April 2018.
\newblock \doi{10.1109/JSEN.2018.2807245}.
\newblock URL \url{https://doi.org/10.1109/JSEN.2018.2807245}.

\bibitem[Christoforou et~al.(2013)Christoforou, Karlaftis, and
  Yannis]{Christoforou2013Reaction}
Z.~Christoforou, M.~Karlaftis, and G.~Yannis.
\newblock Reaction times of young alcohol-impaired drivers.
\newblock \emph{Accident Analysis \& Prevention}, 61:\penalty0 54--62, December
  2013.
\newblock \doi{10.1016/j.aap.2012.12.030}.
\newblock URL \url{https://doi.org/10.1016/j.aap.2012.12.030}.

\bibitem[Chung et~al.(2019)Chung, Chong, and Lee]{Chung2019Methods}
W.-Y. Chung, T.-W. Chong, and B.-G. Lee.
\newblock Methods to detect and reduce driver stress: A review.
\newblock \emph{International Journal of Automotive Technology}, 20\penalty0
  (5):\penalty0 1051--1063, October 2019.
\newblock \doi{10.1007/s12239-019-0099-3}.
\newblock URL \url{https://doi.org/10.1007/s12239-019-0099-3}.

\bibitem[Coetzer and Hancke(2009)]{Coetzer2009Driver}
R.~Coetzer and G.~Hancke.
\newblock Driver fatigue fetection: A survey.
\newblock In \emph{AFRICON}, pages 1--6, Nairobi, Kenya, September 2009. IEEE.
\newblock \doi{10.1109/AFRCON.2009.5308101}.
\newblock URL \url{https://doi.org/10.1109/AFRCON.2009.5308101}.

\bibitem[Critchley(1992)]{Critchley1992OnSleepening}
M.~Critchley.
\newblock On sleepening.
\newblock \emph{Clinical Neurology and Neurosurgery}, 94:\penalty0 121--122,
  1992.
\newblock \doi{10.1016/0303-8467(92)90044-4}.
\newblock URL \url{https://doi.org/10.1016/0303-8467(92)90044-4}.

\bibitem[Dababneh and {El-Gindy}(2015)]{Dababneh2015Driver}
L.~Dababneh and M.~{El-Gindy}.
\newblock Driver vigilance level detection systems: A literature survey.
\newblock \emph{International Journal of Vehicle Performance (IJVP)},
  2\penalty0 (1):\penalty0 1--29, 2015.
\newblock \doi{10.1504/IJVP.2015.074120}.
\newblock URL \url{https://doi.org/10.1504/IJVP.2015.074120}.

\bibitem[Dahiphale and Rao(2015)]{Dahiphale2015AReview}
V.~Dahiphale and S.~Rao.
\newblock A review paper on portable driver monitoring system for teal time
  fatigue.
\newblock In \emph{International Conference on Computing Communication Control
  and Automation}, pages 558--560, Pune, India, February 2015. IEEE.
\newblock \doi{10.1109/ICCUBEA.2015.115}.
\newblock URL \url{https://doi.org/10.1109/ICCUBEA.2015.115}.

\bibitem[Dai et~al.(2010)Dai, Teng, Bai, Shen, and Xuan]{Dai2010Mobile}
J.~Dai, J.~Teng, X.~Bai, Z.~Shen, and D.~Xuan.
\newblock Mobile phone based drunk driving detection.
\newblock In \emph{International ICST Conference on Pervasive Computing
  Technologies for Healthcare}, pages 1--8, Munich, Germany, March 2010.
\newblock \doi{10.4108/ICST.PERVASIVEHEALTH2010.8901}.
\newblock URL \url{https://doi.org/10.4108/ICST.PERVASIVEHEALTH2010.8901}.

\bibitem[{de Santos Sierra} et~al.(2011){de Santos Sierra}, {\'A}vila, {del
  Pozo}, and Casanova]{DeSantos2011Stress}
A.~{de Santos Sierra}, C.~{\'A}vila, G.~{del Pozo}, and J.~Casanova.
\newblock Stress detection by means of stress physiological template.
\newblock In \emph{World Congress on Nature and Biologically Inspired
  Computing}, pages 131--136, Salamanca, Spain, October 2011. IEEE.
\newblock \doi{10.1109/NaBIC.2011.6089448}.
\newblock URL \url{https://doi.org/10.1109/NaBIC.2011.6089448}.

\bibitem[Dinges et~al.(1998{\natexlab{a}})Dinges, Mallis, Maislin, and
  Powell]{Dinges1998Evaluation}
D.~Dinges, M.~Mallis, G.~Maislin, and J.~Powell.
\newblock Evaluation of techniques for ocular measurement as an index of
  fatigue and the basis for alertness management.
\newblock Technical Report DOT HS 808 762, National Highway Traffic Safety
  Administration, Washington, District of Columbia, USA, April
  1998{\natexlab{a}}.

\bibitem[Dinges et~al.(1998{\natexlab{b}})Dinges, Mallis, Maislin, and
  Powell]{Dinges1998PERCLOS}
D.~Dinges, M.~Mallis, G.~Maislin, and J.~Powell.
\newblock {PERCLOS}, a valid psychophysiological measure of alertness as
  assessed by psychomotor vigilance.
\newblock Technical Report FHWA-MCRT-98-006, FHWA, Washington, District of
  Columbia, USA, October 1998{\natexlab{b}}.

\bibitem[Diverrez et~al.(2016)Diverrez, Martin, and
  Pallamin]{Diverrez2016Stress}
J.-M. Diverrez, N.~Martin, and N.~Pallamin.
\newblock Stress interface inducer, a way to generate stress in laboratory
  conditions.
\newblock In \emph{International Conference on Methods and Techniques in
  Behavioral Research (Measuring Behavior)}, pages 25--27, Dublin, Ireland, May
  2016.
\newblock URL \url{https://hal.archives-ouvertes.fr/hal-01525678}.

\bibitem[Dong et~al.(2011)Dong, Hu, Uchimura, and Murayama]{Dong2011Driver}
Y.~Dong, Z.~Hu, K.~Uchimura, and N.~Murayama.
\newblock Driver inattention monitoring system for intelligent vehicles: A
  review.
\newblock \emph{IEEE Transactions on Intelligent Transportation Systems},
  12\penalty0 (2):\penalty0 596--614, June 2011.
\newblock \doi{10.1109/TITS.2010.2092770}.
\newblock URL \url{https://doi.org/10.1109/TITS.2010.2092770}.

\bibitem[Drei{\ss}ig et~al.(2020)Drei{\ss}ig, Baccour, Sch{\"a}ck, and
  Kasneci]{Dreissig2020Driver}
M.~Drei{\ss}ig, M.~Baccour, T.~Sch{\"a}ck, and E.~Kasneci.
\newblock Driver drowsiness classification based on eye blink and head movement
  features using the {k-NN} algorithm.
\newblock In \emph{Symposium Series on Computational Intelligence (SSCI)},
  pages 889--896, Canberra, Australia, December 2020. IEEE.
\newblock \doi{10.1109/SSCI47803.2020.9308133}.
\newblock URL \url{https://doi.org/10.1109/SSCI47803.2020.9308133}.

\bibitem[Durso and Gronlund(1999)]{Durso1999Situation}
F.~Durso and S.~Gronlund.
\newblock \emph{Situation awareness}, pages 283--314.
\newblock John Wiley \& Sons Ltd, 1999.

\bibitem[Ebrahimbabaie(2020)]{Ebrahimbabaie2020Prediction}
P.~Ebrahimbabaie.
\newblock \emph{Prediction of risk of an event using sensor signals, with
  application to the prevention of driving accidents due to drowsiness}.
\newblock PhD thesis, University of Li{\`e}ge, Belgium, December 2020.
\newblock URL \url{https://orbi.uliege.be/handle/2268/250646}.

\bibitem[Ebrahimbabaie and Verly(2018)]{Ebrahimbabaie2018Excellent}
P.~Ebrahimbabaie and J.~Verly.
\newblock Excellent potential of geometric {B}rownian motion ({GBM}) as a
  random process model for level of drowsiness signals.
\newblock In \emph{International Joint Conference on Biomedical Engineering
  Systems and Technologies - BIOSIGNAL}, pages 105--112, Madeira, Portugal,
  2018. SciTePress.
\newblock \doi{10.5220/0006545101050112}.
\newblock URL \url{https://doi.org/10.5220/0006545101050112}.

\bibitem[Ekman(1993)]{Ekman1993Facial}
P.~Ekman.
\newblock Facial expression and emotion.
\newblock \emph{American Psychologist}, 48\penalty0 (4):\penalty0 384--392,
  April 1993.
\newblock \doi{10.1037/0003-066X.48.4.384}.
\newblock URL \url{https://doi.org/10.1037/0003-066X.48.4.384}.

\bibitem[{El Basiouni El Masri} et~al.(2017){El Basiouni El Masri}, Artail, and
  Akkary]{ElBasiouniElMasri2017Toward}
A.~{El Basiouni El Masri}, H.~Artail, and H.~Akkary.
\newblock Toward self-policing: detecting drunk driving behaviors through
  sampling {CAN} bus data.
\newblock In \emph{International Conference on Electrical and Computing
  Technologies and Applications (ICECTA)}, pages 1--5, Ras Al Khaimah, United
  Arab Emirates, November 2017. IEEE.
\newblock \doi{10.1109/ICECTA.2017.8252037}.
\newblock URL \url{https://doi.org/10.1109/ICECTA.2017.8252037}.

\bibitem[{El Khatib} et~al.(2020){El Khatib}, Ou, and
  Karray]{ElKhatib2020Driver}
A.~{El Khatib}, C.~Ou, and F.~Karray.
\newblock Driver inattention detection in the context of next-generation
  autonomous vehicles design: A survey.
\newblock \emph{IEEE Transactions on Intelligent Transportation Systems},
  21\penalty0 (11):\penalty0 4483--4496, November 2020.
\newblock \doi{10.1109/TITS.2019.2940874}.
\newblock URL \url{https://doi.org/10.1109/TITS.2019.2940874}.

\bibitem[Engstr{\"o}m and Markkula(2007)]{Engstrom2007Effects}
J.~Engstr{\"o}m and G.~Markkula.
\newblock Effects of visual and cognitive distraction on lane change test
  performance.
\newblock In \emph{International Driving Symposium on Human Factors in Driver
  Assessment}, pages 199--205, Stevenson, Washington, USA, July 2007.
\newblock \doi{10.17077/drivingassessment.1237}.
\newblock URL \url{https://doi.org/10.17077/drivingassessment.1237}.

\bibitem[Forsman et~al.(2013)Forsman, Vila, Short, Mott, and {Van
  Dongen}]{Forsman2013Efficient}
P.~Forsman, B.~Vila, R.~Short, C.~Mott, and H.~{Van Dongen}.
\newblock Efficient driver drowsiness detection at moderate levels of
  drowsiness.
\newblock \emph{Accident Analysis \& Prevention}, 50\penalty0 (Supplement
  C):\penalty0 341--350, 2013.
\newblock \doi{10.1016/j.aap.2012.05.005}.
\newblock URL
  \url{http://www.sciencedirect.com/science/article/pii/S0001457512001571}.

\bibitem[Fournier et~al.(1999)Fournier, Wilson, and
  Swain]{Fournier1999Electrophysiological}
L.~Fournier, G.~Wilson, and C.~Swain.
\newblock Electrophysiological, behavioral, and subjective indexes of workload
  when performing multiple tasks: manipulations of task difficulty and
  training.
\newblock \emph{International Journal of Psychophysiology}, 31\penalty0
  (2):\penalty0 129--145, 1999.
\newblock \doi{10.1016/s0167-8760(98)00049-x}.
\newblock URL \url{https://doi.org/10.1016/s0167-8760(98)00049-x}.

\bibitem[Fran{\c{c}}ois(2018)]{Francois2018Development}
C.~Fran{\c{c}}ois.
\newblock \emph{Development and validation of algorithms for automatic and
  real-time characterization of drowsiness}.
\newblock PhD thesis, University of Li{\`e}ge, Belgium, October 2018.
\newblock URL
  \url{https://orbi.uliege.be/bitstream/2268/228889/1/DoctoralThesis_ClementineFrancois_October2018.pdf}.

\bibitem[Fran{\c{c}}ois et~al.(2016)Fran{\c{c}}ois, Hoyoux, Langohr, Wertz, and
  Verly]{Francois2016Tests}
C.~Fran{\c{c}}ois, T.~Hoyoux, T.~Langohr, J.~Wertz, and J.~Verly.
\newblock Tests of a new drowsiness characterization and monitoring system
  based on ocular parameters.
\newblock \emph{International Journal of Environmental Research and Public
  Health}, 13\penalty0 (2):\penalty0 174, January 2016.
\newblock \doi{10.3390/ijerph13020174}.
\newblock URL \url{https://doi.org/10.3390/ijerph13020174}.

\bibitem[Fridman et~al.(2016{\natexlab{a}})Fridman, Langhans, Lee, and
  Reimer]{Fridman2016Driver}
L.~Fridman, P.~Langhans, J.~Lee, and B.~Reimer.
\newblock Driver gaze region estimation without use of eye movement.
\newblock \emph{IEEE Transactions on Intelligent Transportation Systems},
  31\penalty0 (3):\penalty0 49--56, May 2016{\natexlab{a}}.
\newblock \doi{10.1109/MIS.2016.47}.
\newblock URL \url{http://doi.org/10.1109/MIS.2016.47}.

\bibitem[Fridman et~al.(2016{\natexlab{b}})Fridman, Lee, Reimer, and
  Victor]{Fridman2016Owl}
L.~Fridman, J.~Lee, B.~Reimer, and T.~Victor.
\newblock `{O}wl' and `{L}izard': patterns of head pose and eye pose in driver
  gaze classification.
\newblock \emph{IET Computer Vision}, 10\penalty0 (4):\penalty0 308--313,
  2016{\natexlab{b}}.
\newblock \doi{10.1049/iet-cvi.2015.0296}.
\newblock URL \url{http://doi.org/10.1049/iet-cvi.2015.0296}.

\bibitem[Fridman et~al.(2018)Fridman, Reimer, Mehler, and
  Freeman]{Fridman2018Cognitive}
L.~Fridman, B.~Reimer, B.~Mehler, and W.~Freeman.
\newblock Cognitive load estimation in the wild.
\newblock In \emph{Proceedings of the 2018 CHI Conference on Human Factors in
  Computing Systems}, pages 1--9, Montr{\'e}al, Canada, April 2018. ACM.
\newblock \doi{10.1145/3173574.3174226}.
\newblock URL \url{https://doi.org/10.1145/3173574.3174226}.

\bibitem[Fridman et~al.(2019)Fridman, Brown, Glazer, Angell, Dodd, Jenik,
  Terwilliger, Patsekin, Kindelsberger, Ding, Seaman, Mehler, Sipperley,
  Pettinato, Seppelt, Angell, Mehler, and Reimer]{Fridman2019MIT}
L.~Fridman, D.~Brown, M.~Glazer, W.~Angell, S.~Dodd, B.~Jenik, J.~Terwilliger,
  A.~Patsekin, J.~Kindelsberger, L.~Ding, S.~Seaman, A.~Mehler, A.~Sipperley,
  A.~Pettinato, B.~Seppelt, L.~Angell, B.~Mehler, and B.~Reimer.
\newblock {MIT} advanced vehicle technology study: large-scale naturalistic
  driving study of driver behavior and interaction with automation.
\newblock \emph{IEEE Access}, 7:\penalty0 102021--102038, July 2019.
\newblock \doi{10.1109/ACCESS.2019.2926040}.
\newblock URL \url{https://doi.org/10.1109/ACCESS.2019.2926040}.

\bibitem[Gable et~al.(2015)Gable, Kun, Walker, and Winton]{Gable2015Comparing}
T.~Gable, A.~Kun, B.~Walker, and R.~Winton.
\newblock Comparing heart rate and pupil size as objective measures of workload
  in the driving context: initial look.
\newblock In \emph{Adjunct Proceedings of the 7th International Conference on
  Automotive User Interfaces and Interactive Vehicular Applications}, pages
  20--25, Nottingham, England, UK, September 2015. ACM.
\newblock \doi{10.1145/2809730.2809745}.
\newblock URL \url{https://doi.org/10.1145/2809730.2809745}.

\bibitem[Gao et~al.(2014)Gao, Y{\"u}ce, and Thiran]{Gao2014Detecting}
H.~Gao, A.~Y{\"u}ce, and J.-P. Thiran.
\newblock Detecting emotional stress from facial expressions for driving
  safety.
\newblock In \emph{IEEE International Conference on Image Processing (ICIP)},
  pages 5961--5965, Paris, France, October 2014.
\newblock \doi{10.1109/ICIP.2014.7026203}.
\newblock URL \url{https://doi.org/10.1109/ICIP.2014.7026203}.

\bibitem[Garrisson et~al.(2021)Garrisson, Scholey, Ogden, and
  Benson]{Garrisson2021TheEffects}
H.~Garrisson, A.~Scholey, E.~Ogden, and S.~Benson.
\newblock The effects of alcohol intoxication on cognitive functions critical
  for driving: A systematic review.
\newblock \emph{Accident Analysis \& Prevention}, 154:\penalty0 1--11, May
  2021.
\newblock \doi{10.1016/j.aap.2021.106052}.
\newblock URL \url{https://doi.org/10.1016/j.aap.2021.106052}.

\bibitem[Gavrilescu and Vizireanu(2019)]{Gavrilescu2019Feedforward}
M.~Gavrilescu and N.~Vizireanu.
\newblock Feedforward neural network-based architecture for predicting emotions
  from speech.
\newblock \emph{Data}, 4\penalty0 (3):\penalty0 1--23, July 2019.
\newblock \doi{10.3390/data4030101}.
\newblock URL \url{https://doi.org/10.3390/data4030101}.

\bibitem[Ghandour et~al.(2020)Ghandour, Neji, {El-Rifaie}, and {Al
  Barakeh}]{Ghandour2020Driver}
R.~Ghandour, B.~Neji, A.~{El-Rifaie}, and Z.~{Al Barakeh}.
\newblock Driver distraction and stress detection systems: A review.
\newblock \emph{International Journal of Engineering and Applied Sciences
  (IJEAS)}, 7\penalty0 (4), April 2020.
\newblock \doi{10.31873/IJEAS.7.04.10}.
\newblock URL \url{https://doi.org/10.31873/IJEAS.7.04.10}.

\bibitem[Godthelp et~al.(1984)Godthelp, Milgram, and Blaauw]{Godthelp1984TLC}
H.~Godthelp, P.~Milgram, and G.~Blaauw.
\newblock The development of a time-related measure to describe driving
  strategy.
\newblock \emph{Human Factors}, 26\penalty0 (3):\penalty0 257--268, June 1984.
\newblock \doi{10.1177/001872088402600302}.

\bibitem[Gon{\c{c}}alves and Bengler(2015)]{Goncalves2015Driver}
J.~Gon{\c{c}}alves and K.~Bengler.
\newblock Driver state monitoring systems--transferable knowledge manual
  driving to {HAD}.
\newblock \emph{Procedia Manufacturing}, 3:\penalty0 3011--3016, 2015.
\newblock \doi{10.1016/j.promfg.2015.07.845}.
\newblock URL \url{https://doi.org/10.1016/j.promfg.2015.07.845}.

\bibitem[Gouverneur et~al.(2017)Gouverneur, {Jaworek-Korjakowska}, K{\"o}ping,
  Shirahama, Kleczek, and Grzegorzek]{Gouverneur2017Classification}
P.~Gouverneur, J.~{Jaworek-Korjakowska}, L.~K{\"o}ping, K.~Shirahama,
  P.~Kleczek, and M.~Grzegorzek.
\newblock Classification of physiological data for emotion recognition.
\newblock In \emph{International Conference on Artificial Intelligence and Soft
  Computing (ICAISC)}, volume 10245 of \emph{Lecture Notes in Computer
  Science}, pages 619--627. Springer, May 2017.
\newblock \doi{10.1007/978-3-319-59063-9_55}.
\newblock URL \url{https://doi.org/10.1007/978-3-319-59063-9_55}.

\bibitem[Gunn et~al.(2018)Gunn, Mackus, Griffin, Munaf{\`o}, and
  Adams]{Gunn2018ASystematic}
C.~Gunn, M.~Mackus, C.~Griffin, M.~Munaf{\`o}, and S.~Adams.
\newblock A systematic review of the next-day effects of heavy alcohol
  consumption on cognitive performance.
\newblock \emph{Addiction}, 113\penalty0 (12):\penalty0 2182--2193, August
  2018.
\newblock \doi{10.1111/add.14404}.
\newblock URL \url{https://doi.org/10.1111/add.14404}.

\bibitem[Guti{\'e}rrez et~al.(2021)Guti{\'e}rrez, Rodr{\'i}guez, and
  Martin]{Gutierrez2021Comprehensive}
J.~Guti{\'e}rrez, V.~Rodr{\'i}guez, and S.~Martin.
\newblock Comprehensive review of vision-based fall detection systems.
\newblock \emph{Sensors}, 21\penalty0 (3):\penalty0 1--50, February 2021.
\newblock \doi{10.3390/s21030947}.
\newblock URL \url{https://doi.org/10.3390/s21030947}.

\bibitem[Hao et~al.(2007)Hao, Wang, Yang, Wang, Guo, and Zhang]{Hao2007Effect}
X.~Hao, Z.~Wang, F.~Yang, Y.~Wang, Y.~Guo, and K.~Zhang.
\newblock The effect of traffic on situation awareness and mental workload:
  simulator-based study.
\newblock In \emph{International Conference on Engineering Psychology and
  Cognitive Ergonomics (EPCE)}, pages 288--296. Springer, 2007.
\newblock \doi{10.1007/978-3-540-73331-7_31}.
\newblock URL \url{https://doi.org/10.1007/978-3-540-73331-7_31}.

\bibitem[Harbluk et~al.(2007)Harbluk, Noy, Trbovich, and
  Eizenman]{Harbluk2007AnOnRoad}
J.~Harbluk, Y.~Noy, P.~Trbovich, and M.~Eizenman.
\newblock An on-road assessment of cognitive distraction: impacts on drivers'
  visual behavior and braking performance.
\newblock \emph{Accident Analysis \& Prevention}, 39\penalty0 (2):\penalty0
  372--379, March 2007.
\newblock \doi{10.1016/j.aap.2006.08.013}.
\newblock URL \url{https://doi.org/10.1016/j.aap.2006.08.013}.

\bibitem[Hargutt and Kruger(2000)]{Hargutt2001Eyelid}
V.~Hargutt and H.~Kruger.
\newblock Eyelid movements and their predictive value for fatigue stages.
\newblock In \emph{International Conference on Traffic and Transport Psychology
  (ICTTP)}, Bern, Switzerland, September 2000.

\bibitem[Harkous and Artail(2019)]{Harkous2019ATwoStage}
H.~Harkous and H.~Artail.
\newblock A two-stage machine learning method for highly-accurate drunk driving
  detection.
\newblock In \emph{International Conference on Wireless and Mobile Computing,
  Networking and Communications (WiMob)}, pages 1--6, Barcelona, Spain, October
  2019. IEEE.
\newblock \doi{10.1109/WiMOB.2019.8923366}.
\newblock URL \url{https://doi.org/10.1109/WiMOB.2019.8923366}.

\bibitem[Harkous et~al.(2018)Harkous, Bardawil, Artail, and
  Daher]{Harkous2018Application}
H.~Harkous, C.~Bardawil, H.~Artail, and N.~Daher.
\newblock Application of hidden {Markov} model on car sensors for detecting
  drunk drivers.
\newblock In \emph{IEEE International Multidisciplinary Conference on
  Engineering Technology (IMCET)}, pages 1--6, Beirut, Lebanon, November 2018.
  IEEE.
\newblock \doi{10.1109/IMCET.2018.8603030}.
\newblock URL \url{https://doi.org/10.1109/IMCET.2018.8603030}.

\bibitem[Hart and Staveland(1988)]{Hart1988Development}
S.~Hart and L.~Staveland.
\newblock Development of {NASA-TLX} ({T}ask {L}oad {I}ndex): results of
  empirical and theoretical research.
\newblock \emph{Advances in Psychology}, 52:\penalty0 139--183, 1988.
\newblock \doi{10.1016/s0166-4115(08)62386-9}.
\newblock URL \url{http://doi.org/10.1016/s0166-4115(08)62386-9}.

\bibitem[Healey and Picard(2005)]{Healey2005Detecting}
J.~Healey and R.~Picard.
\newblock Detecting stress during real-world driving tasks using physiological
  sensors.
\newblock \emph{IEEE Transactions on Intelligent Transportation Systems},
  6\penalty0 (2):\penalty0 156--166, June 2005.
\newblock \doi{10.1109/TITS.2005.848368}.
\newblock URL \url{https://doi.org/10.1109/TITS.2005.848368}.

\bibitem[Hecht et~al.(2018)Hecht, Feldh{\"u}tter, Radlmayr, Nakano, Miki,
  Henle, and Bengler]{Hecht2018AReview}
T.~Hecht, A.~Feldh{\"u}tter, J.~Radlmayr, Y.~Nakano, Y.~Miki, C.~Henle, and
  K.~Bengler.
\newblock A review of driver state monitoring systems in the context of
  automated driving.
\newblock In \emph{Congress of the International Ergonomics Association (IEA)},
  pages 398--408, Florence, Italy, August 2018. Springer.
\newblock \doi{10.1007/978-3-319-96074-6_43}.
\newblock URL \url{https://doi.org/10.1007/978-3-319-96074-6_43}.

\bibitem[Hermosilla et~al.(2018)Hermosilla, Verdugo, Farias, Vera, Pizarro, and
  Machuca]{Hermosilla2018Face}
G.~Hermosilla, J.~Verdugo, G.~Farias, E.~Vera, F.~Pizarro, and M.~Machuca.
\newblock Face recognition and drunk classification using infrared face images.
\newblock \emph{Journal of Sensors}, 2018:\penalty0 1--8, January 2018.
\newblock \doi{10.1155/2018/5813514}.
\newblock URL \url{https://doi.org/10.1155/2018/5813514}.

\bibitem[Hoddes et~al.(1973)Hoddes, Zarcone, Smythe, Phillips, and
  Dement]{Hoddes1973SSS}
E.~Hoddes, V.~Zarcone, H.~Smythe, R.~Phillips, and W.~Dement.
\newblock Quantification of sleepiness: A new approach.
\newblock \emph{Psychophysiology}, 10\penalty0 (4):\penalty0 431--436, July
  1973.
\newblock \doi{10.1111/j.1469-8986.1973.tb00801.x}.

\bibitem[Hu et~al.(2018)Hu, Zhu, Gao, and Zheng]{Hu2018Analysis}
H.~Hu, Z.~Zhu, Z.~Gao, and R.~Zheng.
\newblock Analysis on biosignal characteristics to evaluate road rage of
  younger drivers: A driving simulator study.
\newblock In \emph{IEEE Intelligent Vehicles Symposium}, volume~IV, pages
  156--161, Changshu, China, June 2018. IEEE.
\newblock \doi{10.1109/IVS.2018.8500444}.
\newblock URL \url{https://doi.org/10.1109/IVS.2018.8500444}.

\bibitem[Hu et~al.(2013)Hu, Xie, and Li]{Hu2013Negative}
T.-Y. Hu, X.~Xie, and J.~Li.
\newblock Negative or positive? the effect of emotion and mood on risky
  driving.
\newblock \emph{Transportation Research Part F: Traffic Psychology and
  Behaviour}, 16:\penalty0 29--40, January 2013.
\newblock \doi{10.1016/j.trf.2012.08.009}.
\newblock URL \url{https://doi.org/10.1016/j.trf.2012.08.009}.

\bibitem[Hultman et~al.(2021)Hultman, Johansson, Lindqvist, and
  Ahlstr{\"o}m]{Hultman2021Driver}
M.~Hultman, I.~Johansson, F.~Lindqvist, and C.~Ahlstr{\"o}m.
\newblock Driver sleepiness detection with deep neural networks using
  electrophysiological data.
\newblock \emph{Physiological Measurement}, 42\penalty0 (3), 2021.
\newblock \doi{10.1088/1361-6579/abe91e}.
\newblock URL \url{https://doi.org/10.1088/1361-6579/abe91e}.

\bibitem[Irwin et~al.(2017)Irwin, Iudakhina, Desbrow, and
  McCartney]{Irwin2017Effects}
C.~Irwin, E.~Iudakhina, B.~Desbrow, and D.~McCartney.
\newblock Effects of acute alcohol consumption on measures of simulated
  driving: A systematic review and meta-analysis.
\newblock \emph{Accident Analysis \& Prevention}, 102:\penalty0 248--266, May
  2017.
\newblock \doi{10.1016/j.aap.2017.03.001}.
\newblock URL \url{https://doi.org/10.1016/j.aap.2017.03.001}.

\bibitem[Izumi et~al.(2017)Izumi, Matsunaga, Nakamura, Kawaguchi, and
  Yoshimoto]{Izumicontact}
S.~Izumi, D.~Matsunaga, R.~Nakamura, H.~Kawaguchi, and M.~Yoshimoto.
\newblock A contact-less heart rate sensor system for driver health monitoring,
  2017.

\bibitem[{Jacob\'e de Naurois} et~al.(2019){Jacob\'e de Naurois}, Bourdin,
  Stratulat, Diaz, and Vercher]{JacobedeNaurois2019Detection}
C.~{Jacob\'e de Naurois}, C.~Bourdin, A.~Stratulat, E.~Diaz, and J.-L. Vercher.
\newblock Detection and prediction of driver drowsiness using artificial neural
  network models.
\newblock \emph{Accident Analysis \& Prevention}, 126:\penalty0 95--104, May
  2019.
\newblock \doi{10.1016/j.aap.2017.11.038}.
\newblock URL \url{https://doi.org/10.1016/j.aap.2017.11.038}.

\bibitem[Jeong and Ko(2018)]{Jeong2018Driver}
M.~Jeong and B.~Ko.
\newblock Driver's facial expression recognition in real-time for safe driving.
\newblock \emph{Sensors}, 18\penalty0 (12):\penalty0 1--17, 2018.
\newblock \doi{10.3390/s18124270}.
\newblock URL \url{https://doi.org/10.3390/s18124270}.

\bibitem[Johns(2000)]{Johns2000ASleep}
M.~Johns.
\newblock A sleep physiologist's view of the drowsy driver.
\newblock \emph{Transportation Research Part F: Traffic Psychology and
  Behaviour}, 3\penalty0 (4):\penalty0 241--249, December 2000.
\newblock \doi{10.1016/S1369-8478(01)00008-0}.
\newblock URL \url{https://doi.org/10.1016/S1369-8478(01)00008-0}.

\bibitem[Johns(2001)]{Johns2001Assessing}
M.~Johns.
\newblock Assessing the drowsiness of drivers, 2001.
\newblock Unpublished report commissioned by VicRoads.

\bibitem[Johns et~al.(2005)Johns, Tucker, and Chapman]{Johns2005Monitoring}
M.~Johns, A.~Tucker, and R.~Chapman.
\newblock Monitoring the drowsiness of drivers: A new method based on the
  velocity of eyelid movements.
\newblock In \emph{World Congress on Intelligent Transport Systems}, pages
  1--16, San Francisco, California, USA, November 2005.

\bibitem[Johns et~al.(2014)Johns, Sibi, and Ju]{Johns2014Effect}
M.~Johns, S.~Sibi, and W.~Ju.
\newblock Effect of cognitive load in autonomous vehicles on driver performance
  during transfer of control.
\newblock In \emph{Adjunct Proceedings of the 6th International Conference on
  Automotive User Interfaces and Interactive Vehicular Applications},
  AutomotiveUI '14, pages 1--4, Seattle, Washington, USA, September 2014.
  Association for Computing Machinery.
\newblock \doi{10.1145/2667239.2667296}.
\newblock URL \url{https://doi.org/10.1145/2667239.2667296}.

\bibitem[Joye et~al.(2020)Joye, Rocher, D{\'e}glon, Sidib{\'e}, Favrat,
  Augsburger, and Thomas]{Joye2020Driving}
T.~Joye, K.~Rocher, J.~D{\'e}glon, J.~Sidib{\'e}, B.~Favrat, M.~Augsburger, and
  A.~Thomas.
\newblock Driving under the influence of drugs: A single parallel
  monitoring-based quantification approach on whole blood.
\newblock \emph{Frontiers in Chemistry}, 8:\penalty0 1--10, August 2020.
\newblock \doi{10.3389/fchem.2020.00626}.
\newblock URL \url{https://doi.org/10.3389/fchem.2020.00626}.

\bibitem[Kahneman et~al.(1969)Kahneman, Tursky, Shapiro, and
  Crider]{Kahneman1969Pupillary}
D.~Kahneman, B.~Tursky, D.~Shapiro, and A.~Crider.
\newblock Pupillary, heart rate, and skin resistance changes during a mental
  task.
\newblock \emph{Journal of Experimental Psychology}, 79\penalty0 (1):\penalty0
  164--167, January 1969.
\newblock \doi{10.1037/h0026952}.
\newblock URL \url{https://doi.org/10.1037/h0026952}.

\bibitem[Kajiwara(2014)]{Kajiwara2014Evaluation}
S.~Kajiwara.
\newblock Evaluation of driver's mental workload by facial temperature and
  electrodermal activity under simulated driving conditions.
\newblock \emph{International Journal of Automative Technology}, 15\penalty0
  (1):\penalty0 65--70, 2014.
\newblock \doi{10.1007/s12239-014-0007-9}.
\newblock URL \url{https://doi.org/10.1007/s12239-014-0007-9}.

\bibitem[Kang(2013)]{Kang2013Various}
H.~Kang.
\newblock Various approaches for driver and driving behavior monitoring: A
  review.
\newblock In \emph{IEEE International Conference on Computer Vision Workshops
  (ICCV Workshops)}, pages 616--623, Sydney, NSW, Australia, December 2013.
\newblock \doi{10.1109/ICCVW.2013.85}.
\newblock URL \url{https://doi.org/10.1109/ICCVW.2013.85}.

\bibitem[Kaplan et~al.(2015)Kaplan, Guvensan, Yavuz, and
  Karalurt]{Kaplan2015Driver}
S.~Kaplan, M.~Guvensan, A.~Yavuz, and Y.~Karalurt.
\newblock Driver behavior analysis for safe driving: A survey.
\newblock \emph{IEEE Transactions on Intelligent Transportation Systems},
  16\penalty0 (6):\penalty0 3017--3032, 2015.
\newblock \doi{10.1109/TITS.2015.2462084}.
\newblock URL \url{https://dx.doi.org/10.1109/TITS.2015.2462084}.

\bibitem[Kass et~al.(2007)Kass, Cole, and Stanny]{Kass2007Effects}
S.~Kass, K.~Cole, and C.~Stanny.
\newblock Effects of distraction and experience on situation awareness and
  simulated driving.
\newblock \emph{Transportation Research Part F: Traffic Psychology and
  Behaviour}, 10\penalty0 (4):\penalty0 321--329, July 2007.
\newblock \doi{10.1016/j.trf.2006.12.002}.
\newblock URL \url{https://doi.org/10.1016/j.trf.2006.12.002}.

\bibitem[Kaye et~al.(2018)Kaye, Lewis, and Freeman]{Kaye2018Comparison}
S.-A. Kaye, I.~Lewis, and J.~Freeman.
\newblock Comparison of self-report and objective measures of driving behavior
  and road safety: A systematic review.
\newblock \emph{Journal of Safety Research}, 65:\penalty0 141--151, June 2018.
\newblock \doi{10.1016/j.jsr.2018.02.012}.
\newblock URL \url{https://doi.org/10.1016/j.jsr.2018.02.012}.

\bibitem[Khan and Lee(2019)]{Khan2019AComprehensive}
M.~Khan and S.~Lee.
\newblock A comprehensive survey of driving monitoring and assistance systems.
\newblock \emph{Sensors}, 19\penalty0 (11):\penalty0 1--32, June 2019.
\newblock \doi{10.3390/s19112574}.
\newblock URL \url{https://dx.doi.org/10.3390/s19112574}.

\bibitem[Kiashari et~al.(2020)Kiashari, Nahvi, Bakhoda, Homayounfard, and
  Tashakori]{Kiashari2020Evaluation}
S.~Kiashari, A.~Nahvi, H.~Bakhoda, A.~Homayounfard, and M.~Tashakori.
\newblock Evaluation of driver drowsiness using respiration analysis by thermal
  imaging on a driving simulator.
\newblock \emph{Multimedia Tools and Applications}, 79:\penalty0 17793--17815,
  July 2020.
\newblock \doi{10.1007/s11042-020-08696-x}.
\newblock URL \url{https://doi.org/10.1007/s11042-020-08696-x}.

\bibitem[Kim et~al.(2013)Kim, Jeong, Jung, Park, and Jung]{Kim2013Highly}
J.~Kim, C.~Jeong, M.~Jung, J.~Park, and D.~Jung.
\newblock Highly reliable driving workload analysis using driver
  electroencephalogram ({EEG}) activities during driving.
\newblock \emph{International Journal of Automative Technology}, 14\penalty0
  (6):\penalty0 965--970, 2013.
\newblock \doi{10.1007/s12239-013-0106-z}.
\newblock URL \url{https://doi.org/10.1007/s12239-013-0106-z}.

\bibitem[Kircher et~al.(2002)Kircher, Uddman, and Sandin]{Kircher2002Vehicle}
A.~Kircher, M.~Uddman, and J.~Sandin.
\newblock Vehicle control and drowsiness.
\newblock Technical report, VTI, Link\"oping, Sweden, May 2002.

\bibitem[Kircher and Ahlstr{\"o}m(2016)]{Kircher2016Minimum}
K.~Kircher and C.~Ahlstr{\"o}m.
\newblock Minimum required attention: a human-centered approach to driver
  inattention.
\newblock \emph{Human Factors}, 59\penalty0 (3):\penalty0 471--484, October
  2016.
\newblock \doi{10.1177/0018720816672756}.
\newblock URL \url{https://doi.org/10.1177/0018720816672756}.

\bibitem[Kojima et~al.(2009)Kojima, Maeda, Ogura, Fujita, Murata, Kamei, Tsuji,
  Kaneko, and Yoshizumi]{Kojima2009Noninvasive}
S.~Kojima, S.~Maeda, Y.~Ogura, E.~Fujita, K.~Murata, T.~Kamei, T.~Tsuji,
  S.~Kaneko, and M.~Yoshizumi.
\newblock Noninvasive biological sensor system for detection of drunk driving.
\newblock In \emph{International Conference on Information Technology and
  Applications in Biomedicine ({ITAB})}, pages 1--4, Larnaka, Cyprus, November
  2009. IEEE.
\newblock \doi{10.1109/ITAB.2009.5394324}.
\newblock URL \url{https://doi.org/10.1109/ITAB.2009.5394324}.

\bibitem[Kosch et~al.(2018)Kosch, Hassib, Buschek, and Schmidt]{Kosch2018Look}
T.~Kosch, M.~Hassib, D.~Buschek, and A.~Schmidt.
\newblock Look into my eyes: using pupil dilation to estimate mental workload
  for task complexity adaptation.
\newblock In \emph{Extended Abstracts of the CHI Conference on Human Factors in
  Computing Systems}, pages 1--6, Montr{\'e}al, Canada, April 2018. ACM.
\newblock \doi{10.1145/3170427.3188643}.
\newblock URL \url{https://doi.org/10.1145/3170427.3188643}.

\bibitem[Koukiou and Anastassopoulos(2018)]{Koukiou2017Local}
G.~Koukiou and V.~Anastassopoulos.
\newblock Local difference patterns for drunk person identification.
\newblock \emph{Multimedia Tools and Applications}, 77:\penalty0 9293--9305,
  April 2018.
\newblock \doi{10.1007/s11042-017-4892-6}.
\newblock URL \url{https://doi.org/10.1007/s11042-017-4892-6}.

\bibitem[Kumari and Kumar(2017)]{Kumari2017ASurvey}
B.~Kumari and P.~Kumar.
\newblock A survey on drowsy driver detection system.
\newblock In \emph{International Conference on Big Data Analytics and
  Computational Intelligence ({ICBDAC})}, pages 272--279, Chirala, India, March
  2017. IEEE.
\newblock \doi{10.1109/ICBDACI.2017.8070847}.
\newblock URL \url{https://doi.org/10.1109/ICBDACI.2017.8070847}.

\bibitem[Lal and Craig(2001)]{Lal2001ACritical}
S.~Lal and A.~Craig.
\newblock A critical review of the psychophysiology of driver fatigue.
\newblock \emph{Biological Psychology}, 55\penalty0 (3):\penalty0 173--194,
  February 2001.
\newblock \doi{10.1016/S0301-0511(00)00085-5}.
\newblock URL \url{https://doi.org/10.1016/S0301-0511(00)00085-5}.

\bibitem[Laouz et~al.(2020)Laouz, Ayad, and Terrissa]{Laouz2020Literature}
H.~Laouz, S.~Ayad, and L.~Terrissa.
\newblock Literature review on driver\'s drowsiness and fatigue detection.
\newblock In \emph{International Conference on Intelligent Systems and Computer
  Vision (ISCV)}, pages 1--7, Fez, Morocco, June 2020. IEEE.
\newblock \doi{10.1109/ISCV49265.2020.9204306}.
\newblock URL \url{https://doi.org/10.1109/ISCV49265.2020.9204306}.

\bibitem[Le et~al.(2020)Le, Suzuki, and Aoki]{Le2020Evaluating}
A.~Le, T.~Suzuki, and H.~Aoki.
\newblock Evaluating driver cognitive distraction by eye tracking: From
  simulator to driving.
\newblock \emph{Transportation Research Interdisciplinary Perspectives},
  4:\penalty0 1--7, March 2020.
\newblock \doi{10.1016/j.trip.2019.100087}.
\newblock URL \url{https://doi.org/10.1016/j.trip.2019.100087}.

\bibitem[Le et~al.(2016)Le, Zhu, Zheng, Luu, and Savvides]{Le2016Robust}
T.~Le, C.~Zhu, Y.~Zheng, K.~Luu, and M.~Savvides.
\newblock Robust hand detection in vehicles.
\newblock In \emph{IEEE International Conference on Pattern Recognition
  (ICPR)}, pages 573--578, Cancun, Mexico, December 2016.
\newblock \doi{10.1109/ICPR.2016.7899695}.
\newblock URL \url{https://doi.org/10.1109/ICPR.2016.7899695}.

\bibitem[Le et~al.(2017)Le, Quach, Zhu, Duong, Luu, and Savvides]{Le2017Robust}
T.~Le, K.~Quach, C.~Zhu, C.~Duong, K.~Luu, and M.~Savvides.
\newblock Robust hand detection and classification in vehicles and in the wild.
\newblock In \emph{IEEE International Conference on Computer Vision and Pattern
  Recognition Workshops (CVPRW)}, pages 39--46, Honolulu, Hawaii, USA, 2017.
  IEEE.
\newblock \doi{10.1109/CVPRW.2017.159}.
\newblock URL \url{https://doi.org/10.1109/CVPRW.2017.159}.

\bibitem[Leicht et~al.(2015)Leicht, Skobel, Mathissen, Leonhardt, Weyer,
  Wartzek, Reith, M{\"o}hler, and Teichmann]{Leicht2015Capacitive}
L.~Leicht, E.~Skobel, M.~Mathissen, S.~Leonhardt, S.~Weyer, T.~Wartzek,
  S.~Reith, W.~M{\"o}hler, and D.~Teichmann.
\newblock Capacitive {ECG} recording and beat-to-beat interval estimation after
  major cardiac event.
\newblock In \emph{Annual International Conference of the IEEE Engineering in
  Medicine and Biology Society (EMBC)}, pages 7614--7617, Milan, Italy, August
  2015. IEEE.
\newblock \doi{10.1109/EMBC.2015.7320155}.
\newblock URL \url{https://doi.org/10.1109/EMBC.2015.7320155}.

\bibitem[Leonhardt et~al.(2018)Leonhardt, Leicht, and
  Teichmann]{Leonhardt2018Unobtrusive}
S.~Leonhardt, L.~Leicht, and D.~Teichmann.
\newblock Unobtrusive vital sign monitoring in automotive environments - a
  review.
\newblock \emph{Sensors}, 18\penalty0 (9):\penalty0 1--38, September 2018.
\newblock \doi{10.3390/s18093080}.
\newblock URL \url{https://doi.org/10.3390/s18093080}.

\bibitem[Li et~al.(2017{\natexlab{a}})Li, Sun, Xu, and Chen]{Li2017Multimodal}
H.~Li, J.~Sun, Z.~Xu, and L.~Chen.
\newblock Multimodal {2D+3D} facial expression recognition with deep fusion
  convolutional neural network.
\newblock \emph{IEEE Transactions on Multimedia}, 19\penalty0 (12):\penalty0
  2816--2831, December 2017{\natexlab{a}}.
\newblock \doi{10.1109/TMM.2017.2713408}.
\newblock URL \url{https://doi.org/10.1109/TMM.2017.2713408}.

\bibitem[Li et~al.(2008)Li, Liu, and Luo]{Li2008Design}
R.~Li, C.~Liu, and F.~Luo.
\newblock A design for automotive {CAN} bus monitoring system.
\newblock In \emph{IEEE Vehicle Power and Propulsion Conference}, pages 1--5,
  Harbin, China, September 2008. IEEE.
\newblock \doi{10.1109/VPPC.2008.4677544}.
\newblock URL \url{https://doi.org/10.1109/VPPC.2008.4677544}.

\bibitem[Li et~al.(2015)Li, Jin, and Zhao]{Li2015Drunk}
Z.~Li, X.~Jin, and X.~Zhao.
\newblock Drunk driving detection based on classification of multivariate time
  series.
\newblock \emph{Journal of Safety Research}, 54:\penalty0 61--67, September
  2015.
\newblock \doi{10.1016/j.jsr.2015.06.007}.
\newblock URL \url{https://doi.org/10.1016/j.jsr.2015.06.007}.

\bibitem[Li et~al.(2017{\natexlab{b}})Li, Bao, Kolmanovsky, and
  Yin]{Li2017Visual}
Z.~Li, S.~Bao, I.~Kolmanovsky, and X.~Yin.
\newblock Visual-manual distraction detection using driving performance
  indicators with naturalistic driving data.
\newblock \emph{IEEE Transactions on Intelligent Transportation Systems},
  19\penalty0 (8):\penalty0 2528--2535, August 2017{\natexlab{b}}.
\newblock \doi{10.1109/TITS.2017.2754467}.
\newblock URL \url{https://doi.org/10.1109/TITS.2017.2754467}.

\bibitem[Liang and Lee(2010)]{Liang2010Combining}
Y.~Liang and J.~Lee.
\newblock Combining cognitive and visual distraction: less than the sum of its
  parts.
\newblock \emph{Accident Analysis \& Prevention}, 42\penalty0 (3):\penalty0
  881--890, May 2010.
\newblock \doi{10.1016/j.aap.2009.05.001}.
\newblock URL \url{http://doi.org/10.1016/j.aap.2009.05.001}.

\bibitem[Liang et~al.(2007)Liang, Reyes, and Lee]{Liang2007Real}
Y.~Liang, M.~Reyes, and J.~Lee.
\newblock Real-time detection of driver cognitive distraction using support
  vector machines.
\newblock \emph{IEEE Transactions on Intelligent Transportation Systems},
  8\penalty0 (2):\penalty0 340--350, June 2007.
\newblock \doi{10.1109/TITS.2007.895298}.
\newblock URL \url{https://doi.org/10.1109/TITS.2007.895298}.

\bibitem[Liang et~al.(2019)Liang, Horrey, Howard, Lee, Anderson, Shreeve,
  {O'Brien}, and Czeisler]{Liang2019Prediction}
Y.~Liang, W.~Horrey, M.~Howard, M.~Lee, C.~Anderson, M.~Shreeve, C.~{O'Brien},
  and C.~Czeisler.
\newblock Prediction of drowsiness events in night shift workers during morning
  driving.
\newblock \emph{Accident Analysis \& Prevention}, 126:\penalty0 105--114, May
  2019.
\newblock \doi{10.1016/j.aap.2017.11.004}.
\newblock URL
  \url{https://www.sciencedirect.com/science/article/pii/S0001457517303913}.

\bibitem[Liao et~al.(2016)Liao, Li, Wang, Wang, Li, and
  Cheng]{Liao2016Detection}
Y.~Liao, S.~Li, W.~Wang, Y.~Wang, G.~Li, and B.~Cheng.
\newblock Detection of driver cognitive distraction: A comparison study of
  stop-controlled intersection and speed-limited highway.
\newblock \emph{IEEE Transactions on Intelligent Transportation Systems},
  17\penalty0 (6):\penalty0 1628--1637, June 2016.
\newblock \doi{10.1109/TITS.2015.2506602}.
\newblock URL \url{https://doi.org/10.1109/TITS.2015.2506602}.

\bibitem[Linardatos et~al.(2021)Linardatos, Papastefanopoulos, and
  Kotsiantis]{Linardatos2021Explainable}
P.~Linardatos, V.~Papastefanopoulos, and S.~Kotsiantis.
\newblock Explainable {AI}: A review of machine learning interpretability
  methods.
\newblock \emph{Entropy}, 23\penalty0 (1):\penalty0 1--45, 2021.
\newblock \doi{10.3390/e23010018}.
\newblock URL \url{https://doi.org/10.3390/e23010018}.

\bibitem[Lisper et~al.(1986)Lisper, Laurell, and {van
  Loon}]{Lisper1986Relation}
H.-O. Lisper, H.~Laurell, and J.~{van Loon}.
\newblock Relation between time to falling asleep behind the wheel on a closed
  track and changes in subsidiary reaction time during prolonged driving on a
  motorway.
\newblock \emph{Ergonomics}, 29\penalty0 (3):\penalty0 445--453, 1986.
\newblock \doi{10.1080/00140138608968278}.
\newblock URL \url{https://doi.org/10.1080/00140138608968278}.

\bibitem[Liu et~al.(2009)Liu, Hosking, and Lenn{\'e}]{Liu2009Predicting}
C.~Liu, S.~Hosking, and M.~Lenn{\'e}.
\newblock Predicting driver drowsiness using vehicle measures: recent insights
  and future challenges.
\newblock \emph{Journal of Safety Research}, 40\penalty0 (4):\penalty0
  239--245, August 2009.
\newblock \doi{10.1016/j.jsr.2009.04.005}.
\newblock URL \url{https://doi.org/10.1016/j.jsr.2009.04.005}.

\bibitem[Liu et~al.(2019)Liu, Li, Lv, and Xu]{Liu2019AReview}
F.~Liu, X.~Li, T.~Lv, and F.~Xu.
\newblock A review of driver fatigue detection: progress and prospect.
\newblock In \emph{IEEE International Conference on Consumer Electronics
  (ICCE)}, pages 1--6, Las Vegas, Nevada, USA, January 2019. IEEE.
\newblock \doi{10.1109/ICCE.2019.8662098}.
\newblock URL \url{https://doi.org/10.1109/ICCE.2019.8662098}.

\bibitem[Lowenstein et~al.(1963)Lowenstein, Feinberg, and
  Loewenfeld]{Lowenstein1963Pupillary}
O.~Lowenstein, R.~Feinberg, and I.~Loewenfeld.
\newblock Pupillary movements during acute and chronic fatigue: A new test for
  the objective evaluation of tiredness.
\newblock \emph{Investigative Ophthalmology \& Visual Science}, 2\penalty0
  (2):\penalty0 138--157, April 1963.

\bibitem[Lu et~al.(2021)Lu, Wei, and Li]{Lu2021TheEvolution}
S.~Lu, F.~Wei, and G.~Li.
\newblock The evolution of the concept of stress and the framework of the
  stress system.
\newblock \emph{Cell Stress}, 5\penalty0 (6):\penalty0 76--85, April 2021.
\newblock \doi{10.15698/cst2021.06.250}.
\newblock URL \url{https://doi.org/10.15698/cst2021.06.250}.

\bibitem[Marillier and Verstraete(2019)]{Marillier2019Driving}
M.~Marillier and A.~Verstraete.
\newblock Driving under the influence of drugs.
\newblock \emph{WIREs Forensic Science}, 1\penalty0 (3):\penalty0 1--24,
  January 2019.
\newblock \doi{10.1002/wfs2.1326}.
\newblock URL \url{https://doi.org/10.1002/wfs2.1326}.

\bibitem[{Marina Martinez} et~al.(2018){Marina Martinez}, Heucke, Wang, Gao,
  and Cao]{Martinez2017Driving}
C.~{Marina Martinez}, M.~Heucke, F.~Wang, B.~Gao, and D.~Cao.
\newblock Driving style recognition for intelligent vehicle control and
  advanced driver assistance: A survey.
\newblock \emph{IEEE Transactions on Intelligent Transportation Systems},
  19\penalty0 (3):\penalty0 666--676, 2018.
\newblock \doi{10.1109/TITS.2017.2706978}.
\newblock URL \url{https://dx.doi.org/10.1109/TITS.2017.2706978}.

\bibitem[Marquart and {de Winter}(2015)]{Marquat2015Workload}
G.~Marquart and J.~{de Winter}.
\newblock Workload assessment for mental arithmetic tasks using the task-evoked
  pupillary response.
\newblock \emph{PeerJ Computer Science}, 1:\penalty0 1--20, August 2015.
\newblock ISSN 2376-5992.
\newblock \doi{10.7717/peerj-cs.16}.
\newblock URL \url{https://doi.org/10.7717/peerj-cs.16}.

\bibitem[Marquart et~al.(2015)Marquart, Cabrall, and
  de~Winter]{Marquat2015Review}
G.~Marquart, C.~Cabrall, and J.~de~Winter.
\newblock Review of eye-related measures of drivers' mental workload.
\newblock \emph{Procedia Manufacturing}, 3:\penalty0 2854--2861, 2015.
\newblock ISSN 2351-9789.
\newblock \doi{/10.1016/j.promfg.2015.07.783}.
\newblock URL \url{https://doi.org/10.1016/j.promfg.2015.07.783}.

\bibitem[Martin et~al.(2013)Martin, Solbeck, Mayers, Langille, Buczek, and
  Pelletier]{Martin2013AReview}
T.~Martin, P.~Solbeck, D.~Mayers, R.~Langille, Y.~Buczek, and M.~Pelletier.
\newblock A review of alcohol-impaired driving: the role of blood alcohol
  concentration and complexity of the driving task.
\newblock \emph{Journal of Forensic Sciences}, 58\penalty0 (5):\penalty0
  1238--1250, September 2013.
\newblock \doi{10.1111/1556-4029.12227}.
\newblock URL \url{https://doi.org/10.1111/1556-4029.12227}.

\bibitem[Mashko(2015)]{Mashko2015Review}
A.~Mashko.
\newblock Review of approaches to the problem of driver fatigue and drowsiness.
\newblock In \emph{Smart Cities Symposium Prague (SCSP)}, pages 1--5, Prague,
  Czech Republic, June 2015. IEEE.
\newblock \doi{10.1109/SCSP.2015.7181569}.
\newblock URL \url{https://doi.org/10.1109/SCSP.2015.7181569}.

\bibitem[Mashru and Gandhi(2018)]{Mashru2018Detection}
D.~Mashru and V.~Gandhi.
\newblock Detection of a drowsy state of the driver on road using wearable
  sensors: A survey.
\newblock In \emph{International Conference on Inventive Communication and
  Computational Technologies (ICICCT)}, pages 691--695, Coimbatore, India,
  April 2018. IEEE.
\newblock \doi{10.1109/ICICCT.2018.8473245}.
\newblock URL \url{https://doi.org/10.1109/ICICCT.2018.8473245}.

\bibitem[Masood et~al.(2020)Masood, Rai, Aggarwal, Doja, and
  Ahmad]{Masood2018Detecting}
S.~Masood, A.~Rai, A.~Aggarwal, M.~Doja, and M.~Ahmad.
\newblock Detecting distraction of drivers using convolutional neural network.
\newblock \emph{Pattern Recognition Letters}, 139:\penalty0 79--85, November
  2020.
\newblock \doi{10.1016/j.patrec.2017.12.023}.
\newblock URL \url{https://doi.org/10.1016/j.patrec.2017.12.023}.

\bibitem[Massoz(2019)]{Massoz2019NonInvasive}
Q.~Massoz.
\newblock \emph{Non-invasive, automatic, and real-time characterization of
  drowsiness based on eye closure dynamics}.
\newblock PhD thesis, University of Li{\`e}ge, Belgium, April 2019.
\newblock URL \url{https://orbi.uliege.be/handle/2268/233355}.

\bibitem[Massoz et~al.(2018)Massoz, Verly, and {Van
  Droogenbroeck}]{Massoz2018MultiTimescale}
Q.~Massoz, J.~Verly, and M.~{Van Droogenbroeck}.
\newblock Multi-timescale drowsiness characterization based on a video of a
  driver's face.
\newblock \emph{Sensors}, 18\penalty0 (9):\penalty0 1--17, August 2018.
\newblock ISSN 1424-8220.
\newblock \doi{10.3390/s18092801}.
\newblock URL \url{http://doi.org/10.3390/s18092801}.

\bibitem[May and Baldwin(2008)]{May2008Driver}
J.~May and C.~Baldwin.
\newblock Driver fatigue: the importance of identifying causal factors of
  fatigue when considering detection and countermeasure technologies.
\newblock \emph{Transportation Research Part F: Traffic Psychology and
  Behaviour}, 12\penalty0 (3):\penalty0 218--224, May 2008.
\newblock \doi{10.1016/j.trf.2008.11.005}.
\newblock URL \url{https://doi.org/10.1016/j.trf.2008.11.005}.

\bibitem[May et~al.(1990)May, Kennedy, Williams, Dunlap, and
  Brannan]{May1990Eye}
J.~May, R.~Kennedy, M.~Williams, W.~Dunlap, and J.~Brannan.
\newblock Eye movement indices of mental workload.
\newblock \emph{Acta Psychologica}, 75\penalty0 (1):\penalty0 75--89, October
  1990.
\newblock \doi{10.1016/0001-6918(90)90067-P}.
\newblock URL \url{https://doi.org/10.1016/0001-6918(90)90067-P}.

\bibitem[Melnicuk et~al.(2016)Melnicuk, Birrell, Crundall, and
  Jennings]{Melnicuk2016Towards}
V.~Melnicuk, S.~Birrell, E.~Crundall, and P.~Jennings.
\newblock Towards hybrid driver state monitoring: review, future perspectives
  and the role of consumer electronics.
\newblock In \emph{IEEE Intelligent Vehicles Symposium}, volume~IV, pages
  1392--1397, Gothenburg, Sweden, June 2016. IEEE.
\newblock \doi{10.1109/IVS.2016.7535572}.
\newblock URL \url{https://doi.org/10.1109/IVS.2016.7535572}.

\bibitem[Melnicuk et~al.(2017)Melnicuk, Birrell, Crundall, and
  Jennings]{Melnicuk2017Employing}
V.~Melnicuk, S.~Birrell, E.~Crundall, and P.~Jennings.
\newblock Employing consumer electronic devices in physiological and emotional
  evaluation of common driving activities.
\newblock In \emph{IEEE Intelligent Vehicles Symposium}, volume~IV, pages
  1529--1534, Los Angeles, California, USA, June 2017.
\newblock \doi{10.1109/IVS.2017.7995926}.
\newblock URL \url{https://doi.org/10.1109/IVS.2017.7995926}.

\bibitem[Menon et~al.(2019)Menon, Swathi, Anit, Nair, and
  Sarath]{Menon2019Driver}
S.~Menon, J.~Swathi, S.~Anit, A.~Nair, and S.~Sarath.
\newblock Driver face recognition and sober drunk classification using thermal
  images.
\newblock In \emph{International Conference on Communication and Signal
  Processing (ICCSP)}, pages 400--404, Chennai, India, April 2019. IEEE.
\newblock \doi{10.1109/ICCSP.2019.8697908}.
\newblock URL \url{https://doi.org/10.1109/ICCSP.2019.8697908}.

\bibitem[Mets et~al.(2011)Mets, Kuipers, {de Senerpont Domis}, Leenders,
  Olivier, and Verster]{Mets2011Effects}
M.~Mets, E.~Kuipers, L.~{de Senerpont Domis}, M.~Leenders, B.~Olivier, and
  J.~Verster.
\newblock Effects of alcohol on highway driving in the {STISIM} driving
  simulator.
\newblock \emph{Human Psychopharmacology: Clinical and Experimental},
  26\penalty0 (6):\penalty0 434--439, August 2011.
\newblock \doi{10.1002/hup.1226}.
\newblock URL \url{https://doi.org/10.1002/hup.1226}.

\bibitem[Michael et~al.(2012)Michael, Passmann, and
  Becker]{Michael2012Electrodermal}
L.~Michael, S.~Passmann, and R.~Becker.
\newblock Electrodermal lability as an indicator for subjective sleepiness
  during total sleep deprivation.
\newblock \emph{Journal of Sleep Research}, 21\penalty0 (4):\penalty0 470--478,
  August 2012.
\newblock \doi{10.1111/j.1365-2869.2011.00984.x}.
\newblock URL
  \url{https://onlinelibrary.wiley.com/doi/abs/10.1111/j.1365-2869.2011.00984.x}.

\bibitem[Mittal et~al.(2016)Mittal, Kumar, Dhamija, and Kaur]{Mittal2016Head}
A.~Mittal, K.~Kumar, S.~Dhamija, and M.~Kaur.
\newblock Head movement-based driver drowsiness detection: A review of
  state-of-art techniques.
\newblock In \emph{IEEE International Conference on Engineering and Technology
  (ICETECH)}, pages 903--908, Coimbatore, India, March 2016.
\newblock \doi{10.1109/ICETECH.2016.7569378}.
\newblock URL \url{https://doi.org/10.1109/ICETECH.2016.7569378}.

\bibitem[Monk(1989)]{Monk1989VAS}
T.~Monk.
\newblock A visual analogue scale technique to measure global vigor and affect.
\newblock \emph{Psychiatry Research}, 27\penalty0 (1):\penalty0 89--99, January
  1989.
\newblock ISSN 0165-1781.
\newblock \doi{10.1016/0165-1781(89)90013-9}.
\newblock URL
  \url{http://www.sciencedirect.com/science/article/pii/0165178189900139}.

\bibitem[Mukherjee and Robertson(2015)]{Mukherjee2015Deep}
S.~Mukherjee and N.~Robertson.
\newblock Deep head pose: gaze-direction estimation in multimodal video.
\newblock \emph{IEEE Transactions on Multimedia}, 17\penalty0 (11):\penalty0
  2094--2107, November 2015.
\newblock \doi{10.1109/TMM.2015.2482819}.
\newblock URL \url{https://doi.org/10.1109/TMM.2015.2482819}.

\bibitem[Murata et~al.(2011)Murata, Fujita, Kojima, Maeda, Ogura, Kamei, Tsuji,
  Kaneko, Yoshizumi, and Suzuki]{Murata2011Noninvasive}
K.~Murata, E.~Fujita, S.~Kojima, S.~Maeda, Y.~Ogura, T.~Kamei, T.~Tsuji,
  S.~Kaneko, M.~Yoshizumi, and N.~Suzuki.
\newblock Noninvasive biological sensor system for detection of drunk driving.
\newblock \emph{IEEE Transactions on Information Technology in Biomedicine},
  15\penalty0 (1):\penalty0 19--25, January 2011.
\newblock \doi{10.1109/TITB.2010.2091646}.
\newblock URL \url{https://doi.org/10.1109/TITB.2010.2091646}.

\bibitem[Murthy et~al.(2004)Murthy, Pavlidis, and
  Tsiamyrtzis]{Murthy2004Touchless}
R.~Murthy, I.~Pavlidis, and P.~Tsiamyrtzis.
\newblock Touchless monitoring of breathing function.
\newblock In \emph{Annual International Conference of the IEEE Engineering in
  Medicine and Biology Society}, pages 1196--1199, San Francisco, California,
  USA, 2004.
\newblock \doi{10.1109/IEMBS.2004.1403382}.
\newblock URL \url{https://doi.org/10.1109/IEMBS.2004.1403382}.

\bibitem[Murugan et~al.(2019)Murugan, Selvaraj, and
  Sahayadhas]{Murugan2019Analysis}
S.~Murugan, J.~Selvaraj, and A.~Sahayadhas.
\newblock Analysis of different measures to detect driver states: A review.
\newblock In \emph{IEEE International Conference on System, Computation,
  Automation and Networking (ICSCAN)}, pages 1--6, Pondicherry, India, March
  2019.
\newblock \doi{10.1109/ICSCAN.2019.8878844}.
\newblock URL \url{https://doi.org/10.1109/ICSCAN.2019.8878844}.

\bibitem[Musabini and Chetitah(2020)]{Musabini2020Heatmap}
A.~Musabini and M.~Chetitah.
\newblock Heatmap-based method for estimating drivers' cognitive distraction.
\newblock \emph{CoRR}, abs/2005.14136, May 2020.
\newblock URL \url{https://arxiv.org/abs/2005.14136}.

\bibitem[Nair et~al.(2016)Nair, Ebrahimkutty, Priyanka, Sreeja, and
  Gopu]{Nair2016ASurvey}
I.~Nair, N.~Ebrahimkutty, B.~Priyanka, M.~Sreeja, and D.~Gopu.
\newblock A survey on driver fatigue-drowsiness detection system.
\newblock \emph{International Journal of Engineering and Computer Science},
  5\penalty0 (11):\penalty0 19237--19240, November 2016.
\newblock \doi{10.18535/ijecs/v5i11.92}.
\newblock URL \url{http://www.ijecs.in/index.php/ijecs/article/view/3093}.

\bibitem[Naqvi et~al.(2018)Naqvi, Arsalan, Batchuluun, Yoon, and
  Park]{Naqvi2018DeepLearning}
R.~Naqvi, M.~Arsalan, G.~Batchuluun, H.~Yoon, and K.~Park.
\newblock Deep learning-based gaze detection system for automobile drivers
  using a {NIR} camera sensor.
\newblock \emph{Sensors}, 18\penalty0 (2):\penalty0 1--34, February 2018.
\newblock \doi{10.3390/s18020456}.
\newblock URL \url{https://doi.org/10.3390/s18020456}.

\bibitem[{National Center for Statistics and Analysis}(2020)]{NCSA2020verview}
{National Center for Statistics and Analysis}.
\newblock Overview of motor vehicle crashes in 2019.
\newblock Technical report, National Highway Traffic Safety Administration,
  Washington, District of Columbia, USA, December 2020.
\newblock Traffic Safety Facts Research Note. Report No. DOT HS 813 060.

\bibitem[Ngxande et~al.(2017)Ngxande, Tapamo, and Burke]{Ngxande2017Driver}
M.~Ngxande, J.~Tapamo, and M.~Burke.
\newblock Driver drowsiness detection using behavioral measures and machine
  learning techniques: A review of state-of-art techniques.
\newblock In \emph{Pattern Recognition Association of South Africa and Robotics
  and Mechatronics (PRASA-RobMech)}, pages 156--161, Bloemfontein, South
  Africa, November 2017. IEEE.
\newblock \doi{10.1109/RoboMech.2017.8261140}.
\newblock URL \url{https://doi.org/10.1109/RoboMech.2017.8261140}.

\bibitem[NHTSA(1998)]{NHTSA1998TheVisual}
NHTSA.
\newblock The visual detection of {DWI} motorists.
\newblock Technical Report DOT HS 808 677, National Highway Traffic Safety
  Administration, 1998.

\bibitem[NHTSA(2010)]{NHTSA2010Overview}
NHTSA.
\newblock Overview of the {N}ational {H}ighway {T}raffic {S}afety
  {A}dministration's driver distraction program.
\newblock Technical report, National Highway Traffic Safety Administration,
  Washington, District of Columbia, USA, April 2010.
\newblock DOT HS 811 299.

\bibitem[Nishiyama et~al.(2007)Nishiyama, Tanida, Kusumi, and
  Hirata]{Nishiyama2007ThePupil}
J.~Nishiyama, K.~Tanida, M.~Kusumi, and Y.~Hirata.
\newblock The pupil as a possible premonitor of drowsiness.
\newblock In \emph{Annual International Conference of the IEEE Engineering in
  Medicine and Biology Society (EMBC)}, pages 1586--1589, Lyon, France, August
  2007. IEEE.
\newblock \doi{10.1109/IEMBS.2007.4352608}.

\bibitem[N\v{e}mcov\'{a} et~al.(2021)N\v{e}mcov\'{a}, Svozilov\'{a},
  Bucsuh\'{a}zy, Smi\v{s}ek, M\'{e}zl, Hesko, Bel\'{a}k, Bilik, Maxera, Seitl,
  Dominik, Semela, \v{S}ucha, and Kol\'{a}\v{r}]{Nemcova2021Multimodal}
A.~N\v{e}mcov\'{a}, V.~Svozilov\'{a}, K.~Bucsuh\'{a}zy, R.~Smi\v{s}ek,
  M.~M\'{e}zl, B.~Hesko, M.~Bel\'{a}k, M.~Bilik, P.~Maxera, M.~Seitl,
  T.~Dominik, M.~Semela, M.~\v{S}ucha, and R.~Kol\'{a}\v{r}.
\newblock Multimodal features for detection of driver stress and fatigue:
  review.
\newblock \emph{IEEE Transactions on Intelligent Transportation Systems},
  22\penalty0 (6):\penalty0 3214--3233, June 2021.
\newblock \doi{10.1109/TITS.2020.2977762}.
\newblock URL \url{https://doi.org/10.1109/TITS.2020.2977762}.

\bibitem[{O'Donnel} and Eggemeier(1986)]{Donnel1986Workload}
R.~{O'Donnel} and F.~Eggemeier.
\newblock \emph{Workload assessment methodology}, chapter~42, pages 1--49.
\newblock Wiley, 1986.

\bibitem[Ollander et~al.(2016)Ollander, Godin, Campagne, and
  Charbonnier]{Ollander2016Comparison}
S.~Ollander, C.~Godin, A.~Campagne, and S.~Charbonnier.
\newblock A comparison of wearable and stationary sensors for stress detection.
\newblock In \emph{IEEE International Conference on Systems, Man, and
  Cybernetics (SMC)}, pages 4362--4366, Budapest, Hungary, October 2016. IEEE.
\newblock \doi{10.1109/SMC.2016.7844917}.
\newblock URL \url{https://doi.org/10.1109/SMC.2016.7844917}.

\bibitem[{Oscar-Berman} et~al.(1997){Oscar-Berman}, Shagrin, Evert, and
  Epstein]{Berman1997Impairments}
M.~{Oscar-Berman}, B.~Shagrin, D.~Evert, and C.~Epstein.
\newblock Impairments of brain and behavior: the neurological effects of
  alcohol.
\newblock \emph{Alcohol Health and Research World}, 21\penalty0 (1):\penalty0
  65--75, 1997.
\newblock URL \url{https://pubmed.ncbi.nlm.nih.gov/15706764/}.

\bibitem[Oviedo-Trespalacios et~al.(2016)Oviedo-Trespalacios, Haque, King, and
  Washington]{Oviedo-Trespalacios2016Understanding}
O.~Oviedo-Trespalacios, M.~Haque, M.~King, and S.~Washington.
\newblock Understanding the impacts of mobile phone distraction on driving
  performance: A systematic review.
\newblock \emph{Transportation Research Part C: Emerging Technologies},
  72:\penalty0 360--380, November 2016.
\newblock \doi{10.1016/j.trc.2016.10.006}.
\newblock URL \url{https://doi.org/10.1016/j.trc.2016.10.006}.

\bibitem[PAHO(2018)]{PAHO2018Drinking}
PAHO.
\newblock Drinking and driving.
\newblock Technical Report PAHO/NMH/18-011, Pan American Health Organization,
  2018.

\bibitem[Palasek et~al.(2019)Palasek, Lavie, and
  Palmer]{Palasek2019Attentional}
P.~Palasek, N.~Lavie, and L.~Palmer.
\newblock Attentional demand estimation with attentive driving models.
\newblock In \emph{British Machine Vision Conference (BMVC)}, pages 1--13,
  Cardiff, Wales, September 2019.

\bibitem[Papantoniou et~al.(2017)Papantoniou, Papadimitriou, and
  Yannis]{Papantoniou2017Review}
P.~Papantoniou, E.~Papadimitriou, and G.~Yannis.
\newblock Review of driving performance parameters critical for distracted
  driving research.
\newblock \emph{Transportation Research Procedia}, 25:\penalty0 1796--1805,
  2017.
\newblock \doi{10.1016/j.trpro.2017.05.148}.
\newblock URL \url{https://doi.org/10.1016/j.trpro.2017.05.148}.

\bibitem[Partala and Surakka(2003)]{Partala2003Pupil}
T.~Partala and V.~Surakka.
\newblock Pupil size variation as an indication of affective processing.
\newblock \emph{International Journal of Human-Computer Studies}, 59\penalty0
  (1-2):\penalty0 185--198, July 2003.
\newblock \doi{10.1016/S1071-5819(03)00017-X}.
\newblock URL \url{https://doi.org/10.1016/S1071-5819(03)00017-X}.

\bibitem[Paxion et~al.(2014)Paxion, Galy, and Berthelon]{Paxion2014Mental}
J.~Paxion, E.~Galy, and C.~Berthelon.
\newblock Mental workload and driving.
\newblock \emph{Frontiers in Psychology}, 5:\penalty0 1344, December 2014.
\newblock \doi{10.3389/fpsyg.2014.01344}.
\newblock URL \url{https://doi.org/10.3389/fpsyg.2014.01344}.

\bibitem[Pecher et~al.(2010)Pecher, Lemercier, and
  Cellier]{Pecher2010Influence}
C.~Pecher, C.~Lemercier, and J.-M. Cellier.
\newblock The influence of emotions on driving behavior.
\newblock In Dwight Hennessy, editor, \emph{Traffic Psychology: An
  International Perspective}, chapter~9, pages 1--27. Nova Science Publishers,
  December 2010.
\newblock ISBN 978-1-61668-846-2.

\bibitem[Peck et~al.(2008)Peck, Gebers, Voas, and
  Romano]{Peck2008TheRelationship}
R.~Peck, M.~Gebers, R.~Voas, and E.~Romano.
\newblock The relationship between blood alcohol concentration ({BAC}), age,
  and crash risk.
\newblock \emph{Journal of Safety Research}, 39\penalty0 (3):\penalty0
  311--319, 2008.
\newblock \doi{10.1016/j.jsr.2008.02.030}.
\newblock URL \url{https://doi.org/10.1016/j.jsr.2008.02.030}.

\bibitem[Persson et~al.(2021)Persson, Jonasson, Fredriksson, Wiklund, and
  Ahlstr{\"o}m]{Persson2021Heart}
A.~Persson, H.~Jonasson, I.~Fredriksson, U.~Wiklund, and C.~Ahlstr{\"o}m.
\newblock Heart rate variability for classification of alert versus sleep
  deprived drivers in real road driving conditions.
\newblock \emph{IEEE Transactions on Intelligent Transportation Systems},
  22\penalty0 (6):\penalty0 3316--3325, 2021.
\newblock \doi{10.1109/TITS.2020.2981941}.
\newblock URL \url{https://doi.org/10.1109/TITS.2020.2981941}.

\bibitem[Pfleging et~al.(2016)Pfleging, Fekety, Schmidt, and
  Kun]{Pfleging2016AModel}
B.~Pfleging, D.~Fekety, A.~Schmidt, and A.~Kun.
\newblock A model relating pupil diameter to mental workload and lighting
  conditions.
\newblock In \emph{CHI Conference on Human Factors in Computing Systems}, CHI
  '16, pages 5776--5788, Santa Clara, California, USA, 2016. ACM.
\newblock ISBN 978-1-4503-3362-7.
\newblock \doi{10.1145/2858036.2858117}.
\newblock URL \url{http://doi.acm.org/10.1145/2858036.2858117}.

\bibitem[Pratama et~al.(2017)Pratama, Ardiyanto, and Adji]{Pratama2017AReview}
B.~Pratama, I.~Ardiyanto, and T.~Adji.
\newblock A review on driver drowsiness based on image, bio-signal, and driver
  behavior.
\newblock In \emph{International Conference on Science and Technology -
  Computer (ICST)}, pages 70--75, Yogyakarta, Indonesia, July 2017.
\newblock \doi{10.1109/ICSTC.2017.8011855}.
\newblock URL \url{https://doi.org/10.1109/ICSTC.2017.8011855}.

\bibitem[Ragot et~al.(2017)Ragot, Martin, Em, Pallamin, and
  Diverrez]{Ragot2017Emotion}
M.~Ragot, N.~Martin, S.~Em, N.~Pallamin, and J.-M. Diverrez.
\newblock Emotion recognition using physiological signals: laboratory vs.
  wearable sensors.
\newblock In \emph{International Conference on Applied Human Factors and
  Ergonomics}, volume 608 of \emph{Advances in Intelligent Systems and
  Computing}, pages 15--22. Springer, 2017.
\newblock \doi{10.1007/978-3-319-60639-2_2}.
\newblock URL \url{https://doi.org/10.1007/978-3-319-60639-2_2}.

\bibitem[Ramzan et~al.(2019)Ramzan, Khan, Awan, Ismail, Ilyas, and
  Mahmood]{Ramzan2019ASurvey}
M.~Ramzan, H.~Khan, S.~Awan, A.~Ismail, M.~Ilyas, and A.~Mahmood.
\newblock A survey on state-of-the-art drowsiness detection techniques.
\newblock \emph{IEEE Access}, 7:\penalty0 61904--61919, 2019.
\newblock \doi{10.1109/ACCESS.2019.2914373}.
\newblock URL \url{https://dx.doi.org/10.1109/ACCESS.2019.2914373}.

\bibitem[Ranney(2008)]{Ranney2008Driver}
T.~Ranney.
\newblock Driver distraction: A review of the current state-of-knowledge.
\newblock Technical report, National Highway Traffic Safety Administration,
  April 2008.

\bibitem[Ranney et~al.(2000)Ranney, Mazzae, Garrott, and
  Goodman]{Ranney2000NHTSA}
T.~Ranney, E.~Mazzae, R.~Garrott, and M.~Goodman.
\newblock {NHTSA} driver distraction research: Past, present, and future.
\newblock Technical report, SAE, July 2000.

\bibitem[Ray et~al.(2017)Ray, Das, Kundu, Ghosh, and Rana]{Ray2017Prevention}
A.~Ray, A.~Das, A.~Kundu, A.~Ghosh, and T.~Rana.
\newblock Prevention of driving under influence using microcontroller.
\newblock In \emph{International Conference on Electronics, Materials
  Engineering and Nano-Technology (IEMENTech)}, pages 1--2, Kolkata, India,
  April 2017. IEEE.
\newblock \doi{10.1109/IEMENTECH.2017.8077023}.
\newblock URL \url{https://doi.org/10.1109/IEMENTECH.2017.8077023}.

\bibitem[Regan et~al.(2008)Regan, Lee, and Young]{Regan2008Driver}
M.~Regan, J.~Lee, and K.~Young.
\newblock \emph{Driver distraction: theory, effects, and mitigation}.
\newblock CRC Press, 2008.

\bibitem[Regan et~al.(2011)Regan, Hallett, and Gordon]{Regan2011Driver}
M.~Regan, C.~Hallett, and C.~Gordon.
\newblock Driver distraction and driver inattention: definition, relationship
  and taxonomy.
\newblock \emph{Accident Analysis \& Prevention}, 43\penalty0 (5):\penalty0
  1771--1781, September 2011.
\newblock \doi{10.1016/j.aap.2011.04.008}.
\newblock URL \url{https://doi.org/10.1016/j.aap.2011.04.008}.

\bibitem[Reimer et~al.(2009)Reimer, Mehler, Coughlin, Godfrey, and
  Tan]{Reimer2009Road}
B.~Reimer, B.~Mehler, J.~Coughlin, K.~Godfrey, and C.~Tan.
\newblock An on-road assessment of the impact of cognitive workload on
  physiological arousal in young adult drivers.
\newblock In \emph{International Conference on Automotive User Interfaces and
  Interactive Vehicular Applications}, AutomotiveUI '09, pages 115--118, Essen,
  Germany, September 2009. ACM.
\newblock \doi{10.1145/1620509.1620531}.
\newblock URL \url{https://doi.org/10.1145/1620509.1620531}.

\bibitem[{Rosero-Montalvo} et~al.(2021){Rosero-Montalvo}, {L{\'o}pez-Batista},
  and {Peluffo-Ord{\'o}{\~n}ez}]{Rosero-Montalvo2021Hybrid}
P.~{Rosero-Montalvo}, V.~{L{\'o}pez-Batista}, and D.~{Peluffo-Ord{\'o}{\~n}ez}.
\newblock Hybrid embedded-systems-based approach to in-driver drunk status
  detection using image processing and sensor networks.
\newblock \emph{IEEE Sensors Journal}, 21\penalty0 (14):\penalty0 15729--15740,
  July 2021.
\newblock \doi{10.1109/JSEN.2020.3038143}.
\newblock URL \url{https://doi.org/10.1109/JSEN.2020.3038143}.

\bibitem[Russell(1994)]{Russell1994IsThere}
J.~Russell.
\newblock Is there universal recognition of emotion from facial expression? {A}
  review of the cross-cultural studies.
\newblock \emph{Psychological Bulletin}, 115\penalty0 (1):\penalty0 102,
  January 1994.
\newblock \doi{10.1037/0033-2909.115.1.102}.
\newblock URL \url{https://doi.org/10.1037/0033-2909.115.1.102}.

\bibitem[{SAE International}(2021)]{Sae2021Taxonomy}
{SAE International}.
\newblock Taxonomy and definitions for terms related to driving automation
  systems for on-road motor vehicles.
\newblock Technical Report SAE Standard J3016\_202104, Society of Automobile
  Engineers, Warrendale, PA, USA, April 2021.
\newblock URL \url{https://doi.org/10.4271/J3016_202104}.

\bibitem[Sahayadhas et~al.(2012)Sahayadhas, Sundaraj, and
  Murugappan]{Sahayadhas2012Detecting}
A.~Sahayadhas, K.~Sundaraj, and M.~Murugappan.
\newblock Detecting driver drowsiness based on sensors: A review.
\newblock \emph{Sensors}, 12\penalty0 (12):\penalty0 16937--16953, December
  2012.
\newblock \doi{10.3390/s121216937}.
\newblock URL \url{https://doi.org/10.3390/s121216937}.

\bibitem[Sakairi(2012)]{Sakairi2012Water}
M.~Sakairi.
\newblock Water-cluster-detecting breath sensor and applications in cars for
  detecting drunk or drowsy driving.
\newblock \emph{IEEE Sensors Journal}, 12\penalty0 (5):\penalty0 1078--1083,
  May 2012.
\newblock \doi{10.1109/JSEN.2011.2163816}.
\newblock URL \url{https://doi.org/10.1109/JSEN.2011.2163816}.

\bibitem[Sanders and McCormick(1998)]{Sanders1998Human}
M.~Sanders and E.~McCormick.
\newblock \emph{Human Factors in Engineering and Design}, volume~25.
\newblock Mcgraw-Hill Book Company, 1998.
\newblock \doi{10.1108/ir.1998.25.2.153.2}.
\newblock URL \url{https://doi.org/10.1108/ir.1998.25.2.153.2}.

\bibitem[Sanghvi(2018)]{Sanghvi2018Drunk}
K.~Sanghvi.
\newblock Drunk driving detection.
\newblock \emph{Computer Science and Information Technology}, 6\penalty0
  (2):\penalty0 24--30, 2018.
\newblock \doi{10.13189/csit.2018.060202}.
\newblock URL \url{https://doi.org/10.13189/csit.2018.060202}.

\bibitem[Saponara et~al.(2019)Saponara, Greco, and Gini]{Saponara2019Radar}
S.~Saponara, M.~Greco, and F.~Gini.
\newblock Radar-on-chip/in-package in autonomous driving vehicles and
  intelligent transport systems: opportunities and challenges.
\newblock \emph{IEEE Signal Processing Magazine}, 36\penalty0 (5):\penalty0
  71--84, September 2019.
\newblock \doi{10.1109/MSP.2019.2909074}.
\newblock URL \url{https://doi.org/10.1109/MSP.2019.2909074}.

\bibitem[Schaap et~al.(2013)Schaap, {Van der Horst}, {van Arem}, and
  Brookhuis]{Schaap2013Relationship}
T.~Schaap, A.~{Van der Horst}, B.~{van Arem}, and K.~Brookhuis.
\newblock \emph{The relationship between driver distraction and mental
  workload}, volume~1, pages 63--80.
\newblock CRC Press, 2013.
\newblock ISBN 1409425851.

\bibitem[Schires et~al.(2018)Schires, Georgiou, and Lande]{Schires2018Vital}
E.~Schires, P.~Georgiou, and T.~Lande.
\newblock Vital sign monitoring through the back using an {UWB} impulse radar
  with body coupled antennas.
\newblock \emph{IEEE Transactions on Biomedical Circuits and Systems},
  12\penalty0 (2):\penalty0 292--302, April 2018.
\newblock \doi{10.1109/TBCAS.2018.2799322}.
\newblock URL \url{https://doi.org/10.1109/TBCAS.2018.2799322}.

\bibitem[Schleicher et~al.(2008)Schleicher, Galley, Briest, and
  Galley]{Schleicher2008Blinks}
R.~Schleicher, N.~Galley, S.~Briest, and L.~Galley.
\newblock Blinks and saccades as indicators of fatigue in sleepiness warnings:
  looking tired?
\newblock \emph{Ergonomics}, 51\penalty0 (7):\penalty0 982--1010, June 2008.
\newblock \doi{10.1080/00140130701817062}.
\newblock URL \url{https://doi.org/10.1080/00140130701817062}.

\bibitem[Schr{\"o}ger et~al.(2000)Schr{\"o}ger, Giard, and
  Wolff]{Schroger2000Auditory}
E.~Schr{\"o}ger, M.-H. Giard, and C.~Wolff.
\newblock Auditory distraction: event-related potential and behavioral indices.
\newblock \emph{Clinical Neurophysiology}, 111\penalty0 (8):\penalty0
  1450--1460, August 2000.
\newblock \doi{10.1016/s1388-2457(00)00337-0}.
\newblock URL \url{https://doi.org/10.1016/s1388-2457(00)00337-0}.

\bibitem[Scott-Parker(2017)]{Scott-Parker2017Emotions}
B.~Scott-Parker.
\newblock Emotions, behaviour, and the adolescent driver: A literature review.
\newblock \emph{Transportation Research Part F: Traffic Psychology and
  Behaviour}, 50:\penalty0 1--37, October 2017.
\newblock \doi{10.1016/j.trf.2017.06.019}.
\newblock URL \url{https://doi.org/10.1016/j.trf.2017.06.019}.

\bibitem[Seth(2020)]{Seth2020ASurvey}
I.~Seth.
\newblock A survey on driver behavior detection techniques.
\newblock \emph{International Journal of Scientific Research in Science and
  Technology}, 7\penalty0 (3):\penalty0 401--404, June 2020.
\newblock \doi{10.32628/IJSRST207384}.
\newblock URL \url{https://doi.org/10.32628/IJSRST207384}.

\bibitem[Sevil et~al.(2017)Sevil, Hajizadeh, Samadi, Feng, Lazaro, Frantz, Yu,
  Brandt, Maloney, and Cinar]{Sevil2017Social}
M.~Sevil, I.~Hajizadeh, S.~Samadi, J.~Feng, C.~Lazaro, N.~Frantz, X.~Yu,
  R.~Brandt, Z.~Maloney, and A.~Cinar.
\newblock Social and competition stress detection with wristband physiological
  signals.
\newblock In \emph{IEEE International Conference on Wearable and Implantable
  Body Sensor Networks (BSN)}, pages 39--42, Eindhoven, The Netherlands, May
  2017.
\newblock \doi{10.1109/BSN.2017.7936002}.
\newblock URL \url{https://doi.org/10.1109/BSN.2017.7936002}.

\bibitem[Shameen et~al.(2018)Shameen, Yusoff, Saad, Malik, and
  Muzammel]{Shameen2018Electroencephalography}
Z.~Shameen, M.~Yusoff, M.~Saad, A.~Malik, and M.~Muzammel.
\newblock Electroencephalography ({EEG}) based drowsiness detection for
  drivers: A review.
\newblock \emph{ARPN Journal of Engineering and Applied Sciences}, 13\penalty0
  (4):\penalty0 1458--1464, February 2018.
\newblock ISSN 1819-6608.

\bibitem[Shen et~al.(2006)Shen, Barbera, and Shapiro]{Shen2006Distinguishing}
J.~Shen, J.~Barbera, and C.~Shapiro.
\newblock Distinguishing sleepiness and fatigue: focus on definition and
  measurement.
\newblock \emph{Sleep Medecine Reviews}, 10\penalty0 (1):\penalty0 63--76,
  February 2006.
\newblock \doi{10.1016/j.smrv.2005.05.004}.
\newblock URL \url{https://doi.org/10.1016/j.smrv.2005.05.004}.

\bibitem[Shi et~al.(2007)Shi, Ruiz, Taib, Choi, and Chen]{Shi2007Galvanic}
Y.~Shi, N.~Ruiz, R.~Taib, E.~Choi, and F.~Chen.
\newblock Galvanic skin response ({GSR}) as an index of cognitive load.
\newblock In \emph{CHI Extended Abstracts on Human Factors in Computing
  Systems}, CHI EA '07, pages 2651--2656, San Jose, California, USA, April-May
  2007. ACM.
\newblock \doi{10.1145/1240866.1241057}.
\newblock URL \url{https://doi.org/10.1145/1240866.1241057}.

\bibitem[Shirazi and Rad(2014)]{Shirazi2014Detection}
M.~Shirazi and A.~Rad.
\newblock Detection of intoxicated drivers using online system identification
  of steering behavior.
\newblock \emph{IEEE Transactions on Intelligent Transportation Systems},
  15\penalty0 (4):\penalty0 1738--1747, August 2014.
\newblock \doi{10.1109/TITS.2014.2307891}.
\newblock URL \url{https://doi.org/10.1109/TITS.2014.2307891}.

\bibitem[Sigari et~al.(2014)Sigari, Pourshahabi, Soryani, and
  Fathy]{Sigari2014AReview}
M.~Sigari, M.~Pourshahabi, M.~Soryani, and M.~Fathy.
\newblock A review on driver face monitoring systems for fatigue and
  distraction detection.
\newblock \emph{International Journal of Advanced Science and Technology},
  64:\penalty0 73--100, 2014.
\newblock \doi{10.14257/ijast.2014.64.07}.
\newblock URL \url{https://doi.org/10.14257/ijast.2014.64.07}.

\bibitem[Sikander and Anwar(2019)]{Sikander2019Driver}
G.~Sikander and S.~Anwar.
\newblock Driver fatigue detection systems: A review.
\newblock \emph{IEEE Transactions on Intelligent Transportation Systems},
  20\penalty0 (6):\penalty0 2339--2352, June 2019.
\newblock \doi{10.1109/TITS.2018.2868499}.
\newblock URL \url{https://doi.org/10.1109/TITS.2018.2868499}.

\bibitem[Silva et~al.(2012)Silva, Louren{\c{c}}o, and Fred]{Silva2012InVehicle}
H.~Silva, A.~Louren{\c{c}}o, and A.~Fred.
\newblock In-vehicle driver recognition based on hand {ECG} signals.
\newblock In \emph{ACM International Conference on Intelligent User Interfaces
  (IUI)}, pages 25--28, Lisbon, Portugal, February 2012.
\newblock \doi{10.1145/2166966.2166971}.
\newblock URL \url{https://doi.org/10.1145/2166966.2166971}.

\bibitem[Singh and Kathuria(2021)]{Singh2021Analyzing}
H.~Singh and A.~Kathuria.
\newblock Analyzing driver behavior under naturalistic driving conditions: A
  review.
\newblock \emph{Accident Analysis \& Prevention}, 150:\penalty0 1--21, February
  2021.
\newblock \doi{10.1016/j.aap.2020.105908}.
\newblock URL \url{https://doi.org/10.1016/j.aap.2020.105908}.

\bibitem[Singh(2018)]{Singh2018Critical}
S.~Singh.
\newblock Critical reasons for crashes investigated in the {N}ational {M}otor
  {V}ehicle {C}rash {C}ausation {S}urvey.
\newblock Technical report, National Highway Traffic Safety Administration,
  Washington, District of Columbia, USA, March 2018.
\newblock URL
  \url{https://crashstats.nhtsa.dot.gov/Api/Public/ViewPublication/812506}.

\bibitem[Sodnik et~al.(2008)Sodnik, Dicke, Toma{\v{z}}i{\v{c}}, and
  Billinghurst]{Sodnik2008User}
J.~Sodnik, C.~Dicke, S.~Toma{\v{z}}i{\v{c}}, and M.~Billinghurst.
\newblock A user study of auditory versus visual interfaces for use while
  driving.
\newblock \emph{International Journal of Human-Computer Studies}, 66\penalty0
  (5):\penalty0 318--332, May 2008.
\newblock \doi{10.1016/j.ijhcs.2007.11.001}.
\newblock URL \url{https://doi.org/10.1016/j.ijhcs.2007.11.001}.

\bibitem[Son et~al.(2018)Son, Suzuki, and Aoki]{Son2018Evaluation}
L.~Son, T.~Suzuki, and H.~Aoki.
\newblock Evaluation of cognitive distraction in a real vehicle based on the
  reflex eye movement.
\newblock \emph{International Journal of Automotive Engineering}, 9\penalty0
  (1):\penalty0 1--8, February 2018.
\newblock \doi{10.20485/jsaeijae.9.1_1}.
\newblock URL \url{https://doi.org/10.20485/jsaeijae.9.1_1}.

\bibitem[Sonnleitner et~al.(2014)Sonnleitner, Treder, Simon, Willmann, Ewald,
  Buchner, and Schrauf]{Sonnleitner2014EEG}
A.~Sonnleitner, M.~Treder, M.~Simon, S.~Willmann, A.~Ewald, A.~Buchner, and
  M.~Schrauf.
\newblock {EEG} alpha spindles and prolonged brake reaction times during
  auditory distraction in an on-road driving study.
\newblock \emph{Accident Analysis \& Prevention}, 62:\penalty0 110--118,
  January 2014.
\newblock \doi{10.1016/j.aap.2013.08.026}.
\newblock URL \url{https://doi.org/10.1016/j.aap.2013.08.026}.

\bibitem[Strayer and Drews(2007)]{Strayer2007Cell}
D.~Strayer and F.~Drews.
\newblock Cell-phone -- induced driver distraction.
\newblock \emph{Current Directions in Psychological Science}, 16\penalty0
  (3):\penalty0 128--131, 2007.
\newblock \doi{10.1111/j.1467-8721.2007.00489.x}.
\newblock URL \url{https://doi.org/10.1111/j.1467-8721.2007.00489.x}.

\bibitem[Strayer et~al.(2013)Strayer, Cooper, Turrill, Coleman, Medeiros-Ward,
  and Biondi]{Strayer2013Measuring}
D.~Strayer, J.~Cooper, J.~Turrill, J.~Coleman, N.~Medeiros-Ward, and F.~Biondi.
\newblock Measuring cognitive distraction in the automobile.
\newblock Technical report, AAA, Foundation for Traffic Safety, Washington,
  District of Columbia, USA, 2013.

\bibitem[Strayer et~al.(2015)Strayer, Turrill, Cooper, Coleman, Medeiros-Ward,
  and Biondi]{Strayer2015Assessing}
D.~Strayer, J.~Turrill, J.~Cooper, J.~Coleman, N.~Medeiros-Ward, and F.~Biondi.
\newblock Assessing cognitive distraction in the automobile.
\newblock \emph{Human Factors}, 57\penalty0 (8):\penalty0 1300--1324, December
  2015.
\newblock \doi{10.1177/0018720815575149}.
\newblock URL \url{https://doi.org/10.1177/0018720815575149}.

\bibitem[Subbaiah et~al.(2019)Subbaiah, Reddy, and Rao]{Subbaiah2019Driver}
D.~Subbaiah, P.~Reddy, and K.~Rao.
\newblock Driver drowsiness detection methods: A comprehensive survey.
\newblock \emph{International Journal of Research in Advent Technology},
  7\penalty0 (3):\penalty0 992--997, March 2019.
\newblock \doi{10.32622/ijrat.73201918}.
\newblock URL \url{https://doi.org/10.32622/ijrat.73201918}.

\bibitem[Tantisatirapong et~al.(2010)Tantisatirapong, Senavongse, and
  Phothisonothai]{Tantisatirapong2010Fractal}
S.~Tantisatirapong, W.~Senavongse, and M.~Phothisonothai.
\newblock Fractal dimension based electroencephalogram analysis of drowsiness
  patterns.
\newblock In \emph{International Conference on Electrical
  Engineering/Electronics, Computer, Telecommunications and Information
  Technology (ECTI-CON)}, pages 497--500, Chiang Mai, Thailand, May 2010. IEEE.
\newblock URL \url{https://ieeexplore.ieee.org/document/5491439}.

\bibitem[Teyeb et~al.(2015)Teyeb, Jemai, Zaied, and Amar]{Teyeb2015Vigilance}
I.~Teyeb, O.~Jemai, M.~Zaied, and C.~Amar.
\newblock Vigilance measurement system through analysis of visual and emotional
  driver's signs using wavelet networks.
\newblock In \emph{International Conference on Intelligent Systems Design and
  Applications (ISDA)}, pages 140--147, Marrakech, Morocco, December 2015.
  IEEE.
\newblock \doi{10.1109/ISDA.2015.7489215}.
\newblock URL \url{https://doi.org/10.1109/ISDA.2015.7489215}.

\bibitem[Teyeb et~al.(2016)Teyeb, Jemai, Zaied, and Amar]{Teyeb2016Towards}
I.~Teyeb, O.~Jemai, M.~Zaied, and C.~Amar.
\newblock Towards a smart car seat design for drowsiness detection based on
  pressure distribution of the driver's body.
\newblock In \emph{International Conference on Software Engineering Advances
  (ICSEA)}, pages 217--222, Rome, Italy, August 2016.
\newblock ISBN 978-1-61208-498-5.
\newblock URL
  \url{http://www.thinkmind.org/index.php?view=article&articleid=icsea_2016_9_30_10258}.

\bibitem[Thiffault and Bergeron(2003)]{Thiffault2003Monotony}
P.~Thiffault and J.~Bergeron.
\newblock Monotony of road environment and driver fatigue: A simulator study.
\newblock \emph{Accident Analysis \& Prevention}, 35\penalty0 (3):\penalty0
  381--391, May 2003.
\newblock \doi{10.1016/S0001-4575(02)00014-3}.
\newblock URL \url{https://doi.org/10.1016/S0001-4575(02)00014-3}.

\bibitem[Tijerina(2000)]{Tijerina2000Issues}
L.~Tijerina.
\newblock Issues in the evaluation of driver distraction associated with
  in-vehicle information and telecommunications systems.
\newblock \emph{Transportation Research Inc}, 12:\penalty0 54--67, 2000.

\bibitem[Tu et~al.(2016)Tu, Wei, Hu, Sheng, Nicanfar, Hu, Ngai, and
  Leung]{Tu2016ASurvey}
W.~Tu, L.~Wei, W.~Hu, Z.~Sheng, H.~Nicanfar, X.~Hu, E.~Ngai, and V.~Leung.
\newblock A survey on mobile sensing based mood-fatigue detection for drivers.
\newblock In \emph{Smart City $360^0$}, volume 166 of \emph{Lecture Notes of
  the Institute for Computer Sciences, Social Informatics and
  Telecommunications Engineering}, pages 3--15. Springer, 2016.
\newblock \doi{10.1007/978-3-319-33681-7_1}.
\newblock URL \url{https://doi.org/10.1007/978-3-319-33681-7_1}.

\bibitem[Ukwuoma and Bo(2019)]{Ukwuoma2019Deep}
C.~Ukwuoma and C.~Bo.
\newblock Deep learning review on drivers drowsiness detection.
\newblock In \emph{Technology Innovation Management and Engineering Science
  International Conference (TIMES-iCON)}, pages 1--5, Bangkok, Thailand,
  December 2019. IEEE.
\newblock \doi{10.1109/TIMES-iCON47539.2019.9024642}.
\newblock URL \url{https://doi.org/10.1109/TIMES-iCON47539.2019.9024642}.

\bibitem[Verster et~al.(2014)Verster, Bervoets, {de Klerk}, Vreman, Olivier,
  Roth, and Brookhuis]{Verster2014Effects}
J.~Verster, A.~Bervoets, S.~{de Klerk}, R.~Vreman, B.~Olivier, T.~Roth, and
  K.~Brookhuis.
\newblock Effects of alcohol hangover on simulated highway driving performance.
\newblock \emph{Psychopharmacology}, 231:\penalty0 2999--3008, August 2014.
\newblock \doi{10.1007/s00213-014-3474-9}.
\newblock URL \url{https://doi.org/10.1007/s00213-014-3474-9}.

\bibitem[Verwey and Zaidel(2000)]{Verwey2000Predicting}
W.~Verwey and D.~Zaidel.
\newblock Predicting drowsiness accidents from personal attributes, eye blinks
  and ongoing driving behaviour.
\newblock \emph{Personality and Individual Differences}, 28\penalty0
  (1):\penalty0 123--142, January 2000.
\newblock ISSN 0191-8869.
\newblock \doi{10.1016/S0191-8869(99)00089-6}.
\newblock URL
  \url{http://www.sciencedirect.com/science/article/pii/S0191886999000896}.

\bibitem[Vicente et~al.(2015)Vicente, Huang, Xiong, {De la Torre}, Zhang, and
  Levi]{Vicente2015Driver}
F.~Vicente, Z.~Huang, X.~Xiong, F.~{De la Torre}, W.~Zhang, and D.~Levi.
\newblock Driver gaze tracking and eyes off the road detection system.
\newblock \emph{IEEE Transactions on Intelligent Transportation Systems},
  16\penalty0 (4):\penalty0 2014--2027, August 2015.
\newblock \doi{10.1109/TITS.2015.2396031}.
\newblock URL \url{http://doi.org/10.1109/TITS.2015.2396031}.

\bibitem[Vicente et~al.(2016)Vicente, Laguna, Bartra, and
  Bail{\'o}n]{Vicente2016Drowsiness}
J.~Vicente, P.~Laguna, A.~Bartra, and R.~Bail{\'o}n.
\newblock Drowsiness detection using heart rate variability.
\newblock \emph{Medical {\&} Biological Engineering {\&} Computing},
  54\penalty0 (6):\penalty0 927--937, June 2016.
\newblock ISSN 1741-0444.
\newblock \doi{10.1007/s11517-015-1448-7}.
\newblock URL \url{https://doi.org/10.1007/s11517-015-1448-7}.

\bibitem[Vilaca et~al.(2017)Vilaca, Cunha, and Ferreira]{Vilaca2017Systematic}
A.~Vilaca, P.~Cunha, and A.~Ferreira.
\newblock Systematic literature review on driving behavior.
\newblock In \emph{International Conference on Intelligent Transportation
  Systems (ITSC)}, pages 1--8, Yokohama, Japan, October 2017. IEEE.
\newblock \doi{10.1109/ITSC.2017.8317786}.
\newblock URL \url{https://doi.org/10.1109/ITSC.2017.8317786}.

\bibitem[Vincent et~al.(2006)Vincent, Gribonval, and
  Fevotte]{Vincent2006Performance}
E.~Vincent, R.~Gribonval, and C.~Fevotte.
\newblock Performance measurement in blind audio source separation.
\newblock \emph{IEEE Transactions on Audio, Speech and Language Processing},
  14\penalty0 (4):\penalty0 1462--1469, June 2006.
\newblock \doi{10.1109/TSA.2005.858005}.
\newblock URL \url{https://doi.org/10.1109/TSA.2005.858005}.

\bibitem[Vismaya and Saritha(2020)]{Vismaya2020AReview}
U.~Vismaya and E.~Saritha.
\newblock A review on driver distraction detection methods.
\newblock In \emph{International Conference on Communication and Signal
  Processing (ICCSP)}, pages 483--487, Chennai, India, July 2020. IEEE.
\newblock \doi{10.1109/ICCSP48568.2020.9182316}.
\newblock URL \url{https://doi.org/10.1109/ICCSP48568.2020.9182316}.

\bibitem[Wan et~al.(2017)Wan, Wu, Lin, and Ma]{Wan2017Road}
P.~Wan, C.~Wu, Y.~Lin, and X.~Ma.
\newblock On-road experimental study on driving anger identification model
  based on physiological features by {ROC} curve analysis.
\newblock \emph{IET Intelligent Transport Systems}, 11\penalty0 (5):\penalty0
  290--298, May 2017.
\newblock \doi{10.1049/iet-its.2016.0127}.
\newblock URL \url{https://doi.org/10.1049/iet-its.2016.0127}.

\bibitem[Wang et~al.(2006)Wang, Yang, Ren, and Zheng]{Wang2006Driver}
Q.~Wang, J.~Yang, M.~Ren, and Y.~Zheng.
\newblock Driver fatigue detection: A survey.
\newblock In \emph{World Congress on Intelligent Control and Automation},
  volume~2, pages 8587--8591, Dalian, China, June 2006.
\newblock \doi{10.1109/WCICA.2006.1713656}.
\newblock URL \url{https://doi.org/10.1109/WCICA.2006.1713656}.

\bibitem[Welch et~al.(2019)Welch, Harnett, and Lee]{Welch2019AReview}
K.~Welch, C.~Harnett, and Y.-C. Lee.
\newblock A review on measuring affect with practical sensors to monitor driver
  behavior.
\newblock \emph{Safety}, 5\penalty0 (4):\penalty0 1--18, October 2019.
\newblock \doi{10.3390/safety5040072}.
\newblock URL \url{https://doi.org/10.3390/safety5040072}.

\bibitem[WHO(2018)]{WHO2018Global}
WHO.
\newblock Global status report on road safety 2018: Summary.
\newblock Technical Report WHO/NMH/NVI/18.20, World Health Organization, 2018.

\bibitem[Wickens et~al.(2015)Wickens, Hollands, Banbury, and
  Parasuraman]{Wickens2015Engineering}
C.~Wickens, J.~Hollands, S.~Banbury, and R.~Parasuraman.
\newblock \emph{Engineering psychology and human performance}.
\newblock Psychology Press, 2015.
\newblock \doi{10.4324/9781315665177}.
\newblock URL \url{https://doi.org/10.4324/9781315665177}.

\bibitem[Wierwille and Ellsworth(1994)]{Wierwille1994Evaluation}
W.~Wierwille and L.~Ellsworth.
\newblock Evaluation of driver drowsiness by trained raters.
\newblock \emph{Accident Analysis \& Prevention}, 26\penalty0 (5):\penalty0
  571--581, October 1994.
\newblock \doi{10.1016/0001-4575(94)90019-1}.
\newblock URL
  \url{http://www.sciencedirect.com/science/article/pii/0001457594900191}.

\bibitem[Wierwille et~al.(1994)Wierwille, Ellsworth, Wreggit, Fairbanks, and
  Kirn]{Wierwille1994Research}
W.~Wierwille, L.~Ellsworth, S.~Wreggit, R.~Fairbanks, and C.~Kirn.
\newblock Research on vehicle-based driver status/performance monitoring;
  development, validation, and refinement of algorithms for detection of driver
  drowsiness.
\newblock Technical Report DOT HS 808 247, National Highway Traffic Safety
  Administration, Washington, District of Columbia, USA, December 1994.

\bibitem[Wilhelm et~al.(1998)Wilhelm, Wilhelm, L\"udtke, Streicher, and
  Adler]{Wilhelm1998Pupillographic}
B.~Wilhelm, H.~Wilhelm, H.~L\"udtke, P.~Streicher, and M.~Adler.
\newblock Pupillographic assessment of sleepiness in sleep-deprived healthy
  subjects.
\newblock \emph{Sleep}, 21\penalty0 (3):\penalty0 258--265, May 1998.

\bibitem[Wouters and Bos(2000)]{Wouters2000Traffic}
P.~Wouters and J.~Bos.
\newblock Traffic accident reduction by monitoring driver behaviour with in-car
  data recorders.
\newblock \emph{Accident Analysis \& Prevention}, 32\penalty0 (5):\penalty0
  643--650, September 2000.
\newblock \doi{10.1016/S0001-4575(99)00095-0}.
\newblock URL \url{https://doi.org/10.1016/S0001-4575(99)00095-0}.

\bibitem[Wu et~al.(2016{\natexlab{a}})Wu, Tsang, and Chi]{Wu2016AWearable}
C.~Wu, K.~Tsang, and H.~Chi.
\newblock A wearable drunk detection scheme for healthcare applications.
\newblock In \emph{IEEE International Conference on Industrial Informatics
  (INDIN)}, pages 878--881, Poitiers, France, July 2016{\natexlab{a}}. IEEE.
\newblock \doi{10.1109/INDIN.2016.7819284}.
\newblock URL \url{https://doi.org/10.1109/INDIN.2016.7819284}.

\bibitem[Wu et~al.(2016{\natexlab{b}})Wu, Tsang, Chi, and Hung]{Wu2016APrecise}
C.~Wu, K.~Tsang, H.~Chi, and F.~Hung.
\newblock A precise drunk driving detection using weighted kernel based on
  electrocardiogram.
\newblock \emph{Sensors}, 16\penalty0 (5):\penalty0 1--9, May
  2016{\natexlab{b}}.
\newblock \doi{10.3390/s16050659}.
\newblock URL \url{https://dx.doi.org/10.3390/s16050659}.

\bibitem[Wusk and Gabler(2018)]{Wusk2018NonInvasive}
G.~Wusk and H.~Gabler.
\newblock Non-invasive detection of respiration and heart rate with a vehicle
  seat sensor.
\newblock \emph{Sensors}, 18\penalty0 (5):\penalty0 1--11, May 2018.
\newblock \doi{10.3390/s18051463}.
\newblock URL \url{https://doi.org/10.3390/s18051463}.

\bibitem[Yan et~al.(2016)Yan, Teng, Smith, and Zhang]{Yan2016Driver}
S.~Yan, Y.~Teng, J.~Smith, and B.~Zhang.
\newblock Driver behavior recognition based on deep convolutional neural
  networks.
\newblock In \emph{International Conference on Natural Computation, Fuzzy
  Systems and Knowledge Discovery (ICNC-FSKD)}, pages 636--641, Changsha,
  China, August 2016. IEEE.
\newblock \doi{10.1109/FSKD.2016.7603248}.
\newblock URL \url{http://doi.org/10.1109/FSKD.2016.7603248}.

\bibitem[Yokoyama et~al.(2018)Yokoyama, Eihata, Muramatsu, and
  Fujiwara]{Yokoyama2018Prediction}
H.~Yokoyama, K.~Eihata, J.~Muramatsu, and Y.~Fujiwara.
\newblock Prediction of driver's workload from slow fluctuations of pupil
  diameter.
\newblock In \emph{International Conference on Intelligent Transportation
  Systems (ITSC)}, pages 1775--1780, Maui, HI, USA, November 2018. IEEE.
\newblock \doi{10.1109/ITSC.2018.8569279}.
\newblock URL \url{https://doi.org/10.1109/ITSC.2018.8569279}.

\bibitem[Young and Regan(2007)]{Young2007Driver}
K.~Young and M.~Regan.
\newblock \emph{Driver distraction: A review of the literature}, pages
  379--405.
\newblock Australasian College of Road Safety, 2007.

\bibitem[Young et~al.(2009)Young, Regan, and Lee]{Young2009Measuring}
K.~Young, M.~Regan, and J.~Lee.
\newblock \emph{Measuring the effects of driver distraction: direct driving
  performance methods and measures}, chapter~7, pages 85--105.
\newblock CRC Press, 2009.
\newblock ISBN 9780849374265.

\bibitem[Yusoff et~al.(2017)Yusoff, Ahmad, Guillet, Malik, Saad, and
  M{\'e}rienne]{Yusoff2017Selection}
N.~Yusoff, R.~Ahmad, C.~Guillet, A.~Malik, N.~Saad, and F.~M{\'e}rienne.
\newblock Selection of measurement method for detection of driver visual
  cognitive distraction: A review.
\newblock \emph{IEEE Access}, 5:\penalty0 22844--22854, September 2017.
\newblock \doi{10.1109/ACCESS.2017.2750743}.
\newblock URL \url{https://doi.org/10.1109/ACCESS.2017.2750743}.

\bibitem[Zablocki et~al.(2021)Zablocki, Ben-Younes, P{\'e}rez, and
  Cord]{Zablocki2021Explainability}
{\'E}.~Zablocki, H.~Ben-Younes, P.~P{\'e}rez, and M.~Cord.
\newblock Explainability of vision-based autonomous driving systems: Review and
  challenges.
\newblock \emph{CoRR}, abs/2101.05307, 2021.
\newblock URL \url{https://arxiv.org/abs/2101.05307}.

\bibitem[Zador et~al.(2000)Zador, Krawchuk, and Voas]{Zador2000Alcohol}
P.~Zador, S.~Krawchuk, and R.~Voas.
\newblock Alcohol-related relative risk of driver fatalities and driver
  involvement in fatal crashes in relation to driver age and gender: An update
  using 1996 data.
\newblock \emph{Journal of Studies on Alcohol}, 61\penalty0 (3):\penalty0
  387--395, May 2000.
\newblock \doi{10.15288/jsa.2000.61.387}.
\newblock URL \url{https://doi.org/10.15288/jsa.2000.61.387}.

\bibitem[Zapata et~al.(2021)Zapata, Matey, Montalvo, and
  {Garc{\'i}a-Ruiz}]{Zapata2021Chemical}
F.~Zapata, J.~Matey, G.~Montalvo, and C.~{Garc{\'i}a-Ruiz}.
\newblock Chemical classification of new psychoactive substances ({NPS}).
\newblock \emph{Microchemical Journal}, 163:\penalty0 1--13, April 2021.
\newblock \doi{10.1016/j.microc.2020.105877}.
\newblock URL \url{https://doi.org/10.1016/j.microc.2020.105877}.

\bibitem[Zhang et~al.(2013)Zhang, Qiu, Fu, Zhang, and Ma]{Zhang2013Review}
J.~Zhang, W.~Qiu, H.~Fu, M.~Zhang, and Q.~Ma.
\newblock Review of techniques for driver fatigue detection.
\newblock \emph{Applied Mechanics and Materials}, 433-435:\penalty0 928--931,
  October 2013.
\newblock \doi{10.4028/www.scientific.net/AMM.433-435.928}.
\newblock URL \url{https://doi.org/10.4028/www.scientific.net/AMM.433-435.928}.

\bibitem[Zhang et~al.(2014)Zhang, Zhang, Liu, Zhang, and Yang]{Zhang2014Fast}
K.~Zhang, L.~Zhang, Q.~Liu, D.~Zhang, and M.-H. Yang.
\newblock Fast visual tracking via dense spatio-temporal context learning.
\newblock In \emph{European Conference on Computer Vision (ECCV)}, volume 8693
  of \emph{Lecture Notes in Computer Science}, pages 127--141. Springer, 2014.
\newblock \doi{10.1007/978-3-319-10602-1_9}.
\newblock URL \url{https://doi.org/10.1007/978-3-319-10602-1_9}.

\bibitem[Zhang et~al.(2017)Zhang, Wu, Zhou, Wu, Ou, and Zhou]{Zhang2017Webcam}
Q.~Zhang, Q.~Wu, Y.~Zhou, X.~Wu, Y.~Ou, and H.~Zhou.
\newblock Webcam-based, non-contact, real-time measurement for the
  physiological parameters of drivers.
\newblock \emph{Measurement}, 100:\penalty0 311--321, March 2017.
\newblock \doi{10.1016/j.measurement.2017.01.007}.
\newblock URL \url{https://doi.org/10.1016/j.measurement.2017.01.007}.

\bibitem[Zhang et~al.(2018)Zhang, Zhang, Huang, and Gao]{Zhang2018Speech}
S.~Zhang, S.~Zhang, T.~Huang, and W.~Gao.
\newblock Speech emotion recognition using deep convolutional neural network
  and discriminant temporal pyramid matching.
\newblock \emph{IEEE Transactions on Multimedia}, 20\penalty0 (6):\penalty0
  1576--1590, June 2018.
\newblock \doi{10.1109/TMM.2017.2766843}.
\newblock URL \url{https://doi.org/10.1109/TMM.2017.2766843}.

\bibitem[Zhang et~al.(2016)Zhang, Zheng, Cui, Zong, Yan, and
  Yan]{Zhang2016Deep}
T.~Zhang, W.~Zheng, Z.~Cui, Y.~Zong, J.~Yan, and K.~Yan.
\newblock A deep neural network-driven feature learning method for multi-view
  facial expression recognition.
\newblock \emph{IEEE Transactions on Multimedia}, 18\penalty0 (12):\penalty0
  2528--2536, December 2016.
\newblock \doi{10.1109/TMM.2016.2598092}.
\newblock URL \url{https://doi.org/10.1109/TMM.2016.2598092}.

\bibitem[Zhao et~al.(2016)Zhao, Adib, and Katabi]{Zhao2016Emotion}
M.~Zhao, F.~Adib, and D.~Katabi.
\newblock Emotion recognition using wireless signals.
\newblock In \emph{Annual International Conference on Mobile Computing and
  Networking}, pages 95--108, New York City, New York, USA, October 2016. ACM.
\newblock \doi{10.1145/2973750.2973762}.
\newblock URL \url{https://doi.org/10.1145/2973750.2973762}.

\bibitem[Zin et~al.(2018)Zin, Rodzi, and Ibrahim]{Zin2018Vision}
Z.~Zin, A.~Rodzi, and N.~Ibrahim.
\newblock Vision based eye closeness classification for driver's distraction
  and drowsiness using {PERCLOS} and support vector machines.
\newblock In \emph{International Conference on Machine Vision (ICMV)}, volume
  11041 of \emph{Proceedings of SPIE}, Munich, Germany, March 2018. SPIE.
\newblock \doi{10.1117/12.2522949}.
\newblock URL \url{https://doi.org/10.1117/12.2522949}.

\end{thebibliography}
\end{document}